\documentclass{article}

\usepackage[final]{corl_2026} % CoRL 2026 camera-ready version. Sources live in camera_ready/; example.tex remains the arXiv/preprint version.
\usepackage{booktabs}
\usepackage{xcolor}
\usepackage{float}
\usepackage{graphicx}
\usepackage{amsmath}
\usepackage{amssymb}
\usepackage{siunitx}
\usepackage{multirow}
\usepackage{makecell}
\usepackage{caption}
\usepackage{subcaption}

\usepackage{pgfplots}
\pgfplotsset{compat=1.18}

\definecolor{janus}{RGB}{76,114,176}
\definecolor{uninavid}{RGB}{221,132,82}
\definecolor{vgavs}{RGB}{85,168,104}
\definecolor{vlmnav}{RGB}{150,150,150}
\definecolor{ours}{RGB}{180,35,35}

\usepackage[most]{tcolorbox}

\definecolor{bestcolor}{HTML}{D9F0D3}
\definecolor{secondcolor}{HTML}{FEE8C8}
\definecolor{thirdcolor}{HTML}{E0E0F6}
\newcommand{\best}[1]{\colorbox{bestcolor}{\strut\textbf{#1}}}
\newcommand{\second}[1]{\colorbox{secondcolor}{\strut\textbf{#1}}}
\newcommand{\third}[1]{\colorbox{thirdcolor}{\strut\textbf{#1}}}
\DeclareRobustCommand{\legendbox}[2]{{\setlength{\fboxsep}{1pt}\smash{\colorbox{#1}{\textbf{#2}}}}}
\definecolor{lightblue}{RGB}{30,120,220}

\title{LIME: Learning Intent-aware Camera Motion \\ from Egocentric Video}

\newcommand{\authorhref}[2]{\href{#1}{#2}}

\author{
  \normalfont
  \authorhref{https://boysun045.github.io/boysun-website/}{Boyang Sun}$^{1,*}$ \quad
  \authorhref{https://jiajieli7012.github.io/}{Jiajie Li}$^{1,*}$ \quad
  \authorhref{https://royyang0714.github.io/}{Yung-Hsu Yang}$^1$ \\
  \authorhref{https://zhangcyg.github.io/}{Chenyangguang Zhang}$^1$ \quad
  \authorhref{https://scholar.google.com/citations?user=Ig5T__8AAAAJ}{Tim Engelbracht}$^1$ \quad
  \authorhref{https://sunghwanhong.github.io/}{Sunghwan Hong}$^1$ \\
  \authorhref{https://cesarcadena.ethz.ch/}{Cesar Cadena}$^1$ \quad
  \authorhref{https://scholar.google.com/citations?user=YYH0BjEAAAAJ}{Marc Pollefeys}$^{1,2}$ \quad
  \authorhref{https://hermannblum.net/}{Hermann Blum}$^3$ \\[0.35em]
  *Equal contribution. \\
  $^1$ ETH Zurich \quad
  $^2$ Microsoft \quad
  $^3$ University of Bonn \\
  Project page: \url{https://boysun045.github.io/LIME-Page/}
}

\hypersetup{
  pdftitle={LIME: Learning Intent-aware Camera Motion from Egocentric Video},
  pdfauthor={Boyang Sun, Jiajie Li, Yung-Hsu Yang, Chenyangguang Zhang, Tim Engelbracht, Sunghwan Hong, Cesar Cadena, Marc Pollefeys, Hermann Blum}
}

\begin{document}
\maketitle

% ===============================================================================

\begin{figure}[H]
    \vspace{-5mm}
    \centering  
    \includegraphics[width=\textwidth,height=0.28\textheight,keepaspectratio]{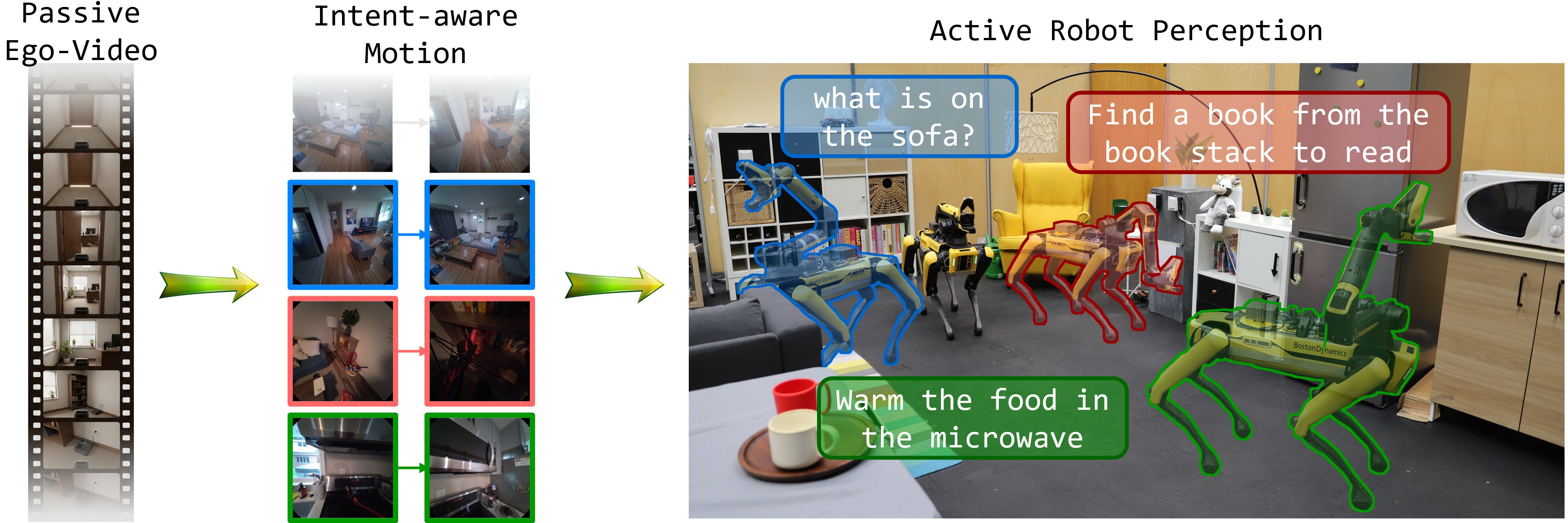}
    \vspace{-5mm}
    \caption{
        LIME learns intent-aware camera motion from passive human video and transfers it to robots: given the current view and a natural-language intent, it generates relative target camera poses that acquire intent-relevant visual evidence.
    }
    \label{fig:teaser}
    \vspace{-2mm}
\end{figure}

\begin{abstract}
Autonomous robots often need to move their camera before they can act: to inspect an object, reveal an occluded region, or obtain a view that responds to a user's intent. While vision-language navigation translates instructions to base motion and vision-language-action policies map instructions to manipulation actions, vision-language conditioned camera motion remains comparatively underexplored as a first-class action. We formulate \emph{vision-language conditioned camera motion generation}: given a current RGB observation and a free-form natural-language intent, predict a relative target camera pose for the next observation. The challenge is that useful viewpoint changes depend on latent perceptual intent, ranging from coarse spatial moves to fine-grained inspection or occlusion revealing. To model this structure, we mine multi-intent camera-motion supervision from egocentric video via hindsight labeling, pairing each relative $SE(3)$ target pose with plausible intents and observation-gain descriptions inferred from the reached view. We propose LIME, a vision-language camera-motion generator that combines an autoregressive observation-gain output with a continuous flow-matching pose head. This design lets the model jointly predict what the next view should reveal while representing multi-hypothesis target views. Across viewpoint-prediction experiments and downstream robotic tasks, we show that LIME learns intent-conditioned camera motion from passive egocentric video, turning ordinary human recordings into supervision for a reusable active-perception primitive that supports general-purpose viewpoint selection, manipulation, embodied question answering, and mid-distance navigation.
\end{abstract}


%===============================================================================

\section{Introduction}
\label{sec:intro}

Vision is often treated as an input to action: an agent observes the scene, recognizes what matters, and then decides where to move or what to manipulate \cite{siegwart2011introduction}.
In everyday behavior, however, this dependence also runs in the opposite direction: an intention often causes us to move our sensors before acting, so that the next observation contains the information we need \cite{bajcsy2018revisiting}.
We lean to see behind an occluder, step closer to inspect a small detail, look around a corner before entering, or shift viewpoint to disambiguate an object's shape.
These motions are not merely navigation or manipulation side effects; they are camera-motion actions whose purpose is to acquire intent-relevant visual evidence.
Because most modern mobile robots carry onboard cameras, learning this mapping from language intent to useful viewpoint change is a natural component of general embodied intelligence.
This motivates the problem of vision-language conditioned camera motion generation: given the current view and an intent, predict how the camera should move to obtain a more useful next observation.

Active perception has long studied sensor motion for better observations \cite{bajcsy2018revisiting,ahmed2023active,placed2023survey}, with most formulations optimizing task-specific utilities such as exploration \cite{lluvia2021active}, reconstruction \cite{li2026motion}, and object search \cite{batra2020objectnav}.
Recent language-conditioned embodied models broaden robot behavior.
In vision-language navigation, instructions specify routes, destinations, or landmarks, and camera viewpoint changes occur as a consequence of moving through the scene~\cite{zhang2024uninavid,VLN-BERT}.
In vision-language-action manipulation, visual observations condition end-effector control, while camera motion is coupled to the embodiment or execution policy~\cite{kim24openvla,zitkovich2023rt}.
These settings leave open a different interface: not ``where should the robot navigate to?'' or ``what should the arm and gripper do?'', but ``how should the camera move so the next observation better resolves a given intent?''

A central difficulty is that the desired next view depends on the intent the agent is trying to resolve, not only on the visible scene geometry.
Given the same observation, an agent may need to move differently depending on whether the intent is to inspect an object, reveal an occluded region, enter a room, or prepare for a downstream interaction.
Furthermore, for the same intent, multiple relative target poses may be valid because different viewpoints can expose different but similarly useful evidence.
Conversely, the same camera motion can support intents at different semantic granularities, from checking a visible object part to understanding the layout of a larger space.
This makes the problem one of modeling a language-conditioned distribution over target camera poses, rather than deterministic next-pose regression.
A useful model should therefore couple geometric pose prediction with an efficient representation of the visual evidence the motion is expected to reveal: the former captures where the camera should move, while the latter captures why that view is useful.

In this paper, we study this interface as vision-language conditioned camera motion generation.
Given a current RGB observation and a free-form intent, the task is to predict a distribution over relative $SE(3)$ target camera poses together with an observation-gain description of what the next view is expected to reveal.
To obtain supervision without teleoperated or purpose-collected active-perception demonstrations~\cite{xiong2025vision,kerrj2025eyerobot,yu2026egomi}, we mine egocentric video by pairing temporally separated frames: the relative camera transform provides a motion target, while a VLM, given both frames and the motion between them, labels plausible intents and observation-gain descriptions in hindsight.
We instantiate this formulation in LIME, a VLM-based model that autoregressively predicts observation gain and conditions a continuous flow-matching pose head on the resulting hidden sequence.
We evaluate this formulation through a dedicated camera-motion benchmark and downstream embodied tasks.
The results demonstrate that LIME can act as a reusable active-perception primitive across diverse embodied tasks.

In summary, our main contributions are:
\begin{itemize}
    \item We formulate \emph{intent-aware camera motion generation}, where an embodied agent predicts a relative $SE(3)$ target camera pose from the current observation and a free-form intent.
    \item We introduce LIME, a vision-language camera-motion generator trained from egocentric video frame pairs with mined intents, observation-gain descriptions, and relative camera poses.
    \item We build a $425$-example benchmark in photorealistic 3D Gaussian Splatting (3DGS) scenes spanning target approaching, exploration, and perspective shift, on which LIME outperforms navigation and active-view-selection baselines, and show the same model supports downstream manipulation, embodied question answering, and mid-distance navigation.
\end{itemize}

\section{Related Work}
\label{sec:related_work}

% Active perception has long studied how robots should move their sensors to improve observations. Classical exploration and active mapping choose next-best views to reduce map or reconstruction uncertainty \cite{yamauchi1997frontier,chaplot2021seal,yu2023frontier,schmid2020efficient,boysun2025frontiernet,yan2022mui,Kompis2021InformedScenes}. Active localization selects motions that improve state or pose estimates \cite{li2025actloc,zhang2020fisher}. Object-goal and image-goal navigation add semantic or visual targets, asking an agent to search for or reach a specified goal object or image \cite{batra2020objectnav,chang2023goat,zhang20233d,beliefmapnav,xie2025naviformer}. These methods actively reason about viewpoint, but their objectives are usually hand-designed around a specific task, such as coverage, reconstruction, localization, or target search. They do not study free-form language intent as the conditioning signal for continuous camera motion.

Active perception has long studied sensor motion for information gathering: classical exploration and active mapping reduce map or reconstruction uncertainty by designing task-specific information-gain measures~\cite{yamauchi1997frontier,chaplot2021seal,yu2023frontier,schmid2020efficient,boysun2025frontiernet,yan2022mui,Kompis2021InformedScenes}, active localization improves state or pose estimates by learning to look at feature-rich regions~\cite{li2025actloc,zhang2020fisher}, and object- or image-goal navigation searches for semantic or visual targets with accumulated scene knowledge~\cite{batra2020objectnav,chang2023goat,zhang20233d,beliefmapnav,xie2025naviformer}. These methods reason about viewpoint, but optimize predefined objectives such as coverage, reconstruction, localization, or target search, rather than free-form language intent as the conditioning signal for continuous camera motion.

Recent language-conditioned embodied models broaden the goal space of robot learning \cite{gu2022vision,zhang2024vision,kawaharazuka2025vision}, but usually place language over navigation or manipulation actions. In Vision-and-Language Navigation (VLN), natural-language instructions specify a route, destination, or landmark, while the agent acts through base motion or discrete waypoints \cite{wei2025streamvln,zeng2025janusvln,zhang2024uninavid,cheng2024navila,chu2026abot,wei2025ground}. Another line of work uses training-free pipelines that leverage LLMs or VLMs without updating the model parameters \cite{padilla2026openfrontier,long2024instructnav,goetting2024end,habibpour2025history,yin2025unigoal}. In Vision-Language-Action (VLA) manipulation, language and vision condition end-effector or whole-body actions for task execution \cite{kim24openvla,zitkovich2023rt,black2024pi_0}. A subset of works explicitly studies active viewpoint selection for better manipulation \cite{xiong2025vision,kerrj2025eyerobot,zou2026activeglasses,wang2025observer,liu2026activevla}.
These approaches typically learn viewpoint behavior jointly with task execution through imitation or reinforcement learning, which ties them to task-specific demonstrations, reward designs, or training environments and limits their generality across intents.

Existing benchmarks for active perception span a range of embodied navigation settings, largely focusing on goal reaching rather than fine-grained viewpoint adjustment. Navigation benchmarks such as ObjectNav~\cite{batra2020objectnav}, VLN-CE~\cite{krantz2020vlnce}, GOAT-Bench~\cite{khanna2024goatbench}, and HM3D-OVON~\cite{yokoyama2024hm3dovon} measure goal reaching in mesh-based simulators with discrete or low-dimensional actions, leaving fine-grained viewpoint adjustment outside the task definition. Embodied question-answering benchmarks, from EmbodiedQA~\cite{das2017embodiedquestionanswering} to OpenEQA~\cite{OpenEQA2023}, HM-EQA~\cite{ren2024explore}, and EXPRESS-Bench~\cite{EXPRESSBench}, often use synthetic or reconstructed indoor environments and navigation-style actions, but score answer correctness rather than motion quality. Closest to our setting, VG-AVS~\cite{koo2025toward} studies single-step local view selection in procedural and mesh-based scenes, I-Perceive~\cite{huang2026perceive} focuses on simulation-centered inspection pose prediction without dedicated exploration evaluation, and E3VS-Bench~\cite{sakamoto2026e3vs} uses high-fidelity 3DGS observations but centers on local inspection around a nearby or already-visible target. These benchmarks highlight the importance of viewpoint selection, while leaving room for evaluating vision-language conditioned camera motion across broader intent granularities, from exploration and target approaching to fine-grained 6-DoF perspective adjustment.
\section{Method}
\label{sec:method}

\subsection{Problem Formulation}
\label{sec:method_problem}

\begin{figure}[!t]
    \centering
    \includegraphics[width=\textwidth]{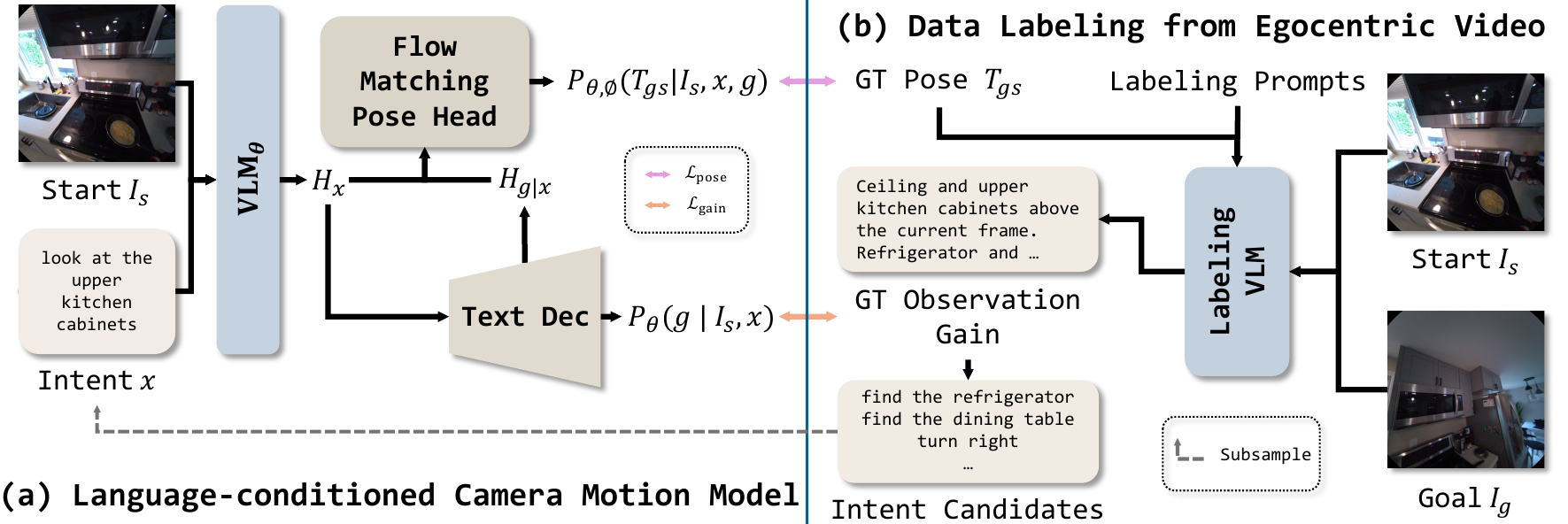}
    \vspace{-10pt}
    \caption{
        LIME overview.
        Panel (a) shows the VLM-based camera-motion generator with an autoregressive language interface and a continuous flow-matching pose head (Sec.~\ref{sec:method_model}).
        Panel (b) shows how we mine passive egocentric video into intent-conditioned camera-motion supervision with observation-gain descriptions (Sec.~\ref{sec:method_data}).
    }
    \label{fig:pipeline}
\end{figure}

We study vision-language conditioned camera motion generation: given a current RGB observation $I_s$ and a free-form language intent $x$, predict a relative camera motion $T_{gs}$ that moves the camera from the current pose $P_s$ to a goal pose $P_g$, producing a next observation $I_g$ that provides visual evidence relevant to the intent. In addition to motion, we predict an observation-gain description $g$, a natural-language summary of what the next view is expected to reveal beyond the current view.
Therefore, the model learns
\begin{equation}
% \[
p_\theta(g \mid I_s, x),
\qquad
p_{\theta,\phi}(T_{gs} \mid I_s, x, g),
% \]
\label{sec:method:eq:problem}
\end{equation}
where $T_{gs}$ is represented by 3D translation and a continuous 6D rotation parameterization.
At a high level, this asks where the camera should look next to resolve the intent.
LIME implements the query with two coupled interfaces over a shared vision-language representation: a language interface for $g$ and an action interface for $T_{gs}$. Figure~\ref{fig:pipeline} shows the overview of the proposed pipeline.

% Figure~\ref{fig:pipeline} summarizes the overall LIME pipeline.
% A data engine converts passive egocentric video into training tuples by pairing current and future views, computing their relative camera motion, and labeling plausible intents and observation-gain descriptions.
% A vision-language model then learns to map the current image and one intent to both the observation-gain description and the corresponding camera motion.
% Together, these components turn ordinary egocentric recordings into supervision for robot camera motion generation.

\subsection{Vision-Language Camera-Motion Generator}
\label{sec:method_model}

The observation-gain interface keeps part of the task in the VLM's native output space by describing the intended visual outcome of the movement.
This outcome-level language target is naturally conditioned on both image and intent, and provides supervision that encourages the hidden representation to anticipate what new evidence a future view should contain.
The continuous flow-matching head then models the conditional distribution of relative $SE(3)$ target poses from this fused representation. Formally, given $(I_s, x)$, we construct a multimodal prompt and encode the hidden sequence $H_x = \mathrm{VLM}_\theta(I_s, x)$ over visual and language tokens with the VLM backbone before generating the observation-gain description.
% with the VLM backbone,
% \[
%     H_x = \mathrm{VLM}_\theta(I_s, x),
% \]
% where $H_x$ denotes the hidden sequence over visual and language tokens before generating the observation-gain description.
The language interface uses the VLM's autoregressive decoder to generate the observation-gain description,
\begin{equation}
    p_\theta(g \mid I_s, x) = \prod_k p_\theta(g_k \mid g_{<k}, H_x).
\label{sec:method:eq:lang}
\end{equation}
%The generated observation-gain tokens are grounded in the fused image-language context and decode to a description of the new visual evidence the next view should reveal.

After autoregressive gain generation, we form the gain-conditioned hidden sequence $H_{x,g}=[H_x;H_{g|x}]$, where $H_{g|x}$ denotes the hidden states of the observation-gain tokens conditioned on the image-intent prompt.
This sequence combines three sources of information: the current visual evidence, the language intent, and the predicted visual outcome of the motion.
%We use this sequence as the condition for camera-motion prediction.
Instead of representing actions or spatial coordinates as language tokens \cite{zitkovich2023rt,kim24openvla,Qwen3-VL}, we attach a separate continuous flow-matching head, which preserves geometric supervision in $SE(3)$ and models the multimodal distribution of plausible target transforms, similar to recent works~\cite{black2024pi_0}.
This captures the multiple valid target poses that can satisfy the same $(I_s, x)$.

We parameterize the target transform $T_{gs}$ as $y=\psi(T_{gs}) \in \mathbb{R}^9$, using 3D translation and the first two columns of the rotation matrix \cite{zhou2018continuity}.
For a flow time $t \sim \mathcal{U}(0,1)$ and Gaussian noise $\epsilon \sim \mathcal{N}(0,I)$, we construct $z_t = (1-t)\epsilon + t{y}$.
We adopt the x-prediction parameterization \cite{li2025back}: instead of regressing the velocity field, the pose head $F_\phi$ takes $(H_{x,g}, z_t, t)$ and predicts the clean target $\hat{y}_\phi = F_\phi(H_{x,g}, z_t, t)$, with loss $\mathcal{L}_\mathrm{pose} = \|\hat{y}_\phi - {y}\|_2^2$.
The velocity used for numerical integration is then recovered as $v_t=(\hat{y}_\phi-z_t)/(1-t)$.
This anchors supervision to valid pose parameters, especially the orthonormalized rotation representation, rather than unconstrained velocities.

\subsection{Mining Active Camera-Motion Supervision from Passive Egocentric Video}
\label{sec:method_data}

Egocentric video provides raw camera-motion supervision: ordered frames capture viewpoint transitions, and poses provide geometry.
What it lacks is intent; available annotations usually describe coarse activities or scenes, not the fine-grained perceptual reason for moving the camera \cite{grauman2024ego,ma2024nymeria,damen2020epic,zheng2026egoscale,kareer2025egomimic}.
Yet nontrivial egocentric motions often reveal new evidence, improve existing views, or reorient toward another region, so we interpret each start--goal pair in hindsight and mine plausible intent labels using the goal view and relative motion as privileged context for labeling. For each egocentric trajectory, we first sample temporally ordered start--goal frame pairs $(I_s, I_g)$.
We retain local transitions with available RGB frames and valid camera geometry, discarding pairs with excessive displacement.
When camera poses are not provided by the dataset, they can be recovered from RGB trajectories using off-the-shelf camera-pose or reconstruction methods~\cite{depthanything3}. We then label each retained transition with a structured hindsight VLM prompt.
The labeler receives the current frame $I_s$, the goal frame $I_g$, and a compact summary of $T_{gs}$, which gives explicit motion cues, including translation direction, distance, and rotation angle, so the generated labels remain grounded in the actual camera movement.
Instead of asking for a free-form caption, the prompt asks for contrastive fields: a motion type, newly visible objects or regions, improved views of content already present in $I_s$, spatial anchors between the two views, an observation-gain description $g$, and a set of plausible intents $\mathcal{X}_{s,g}=\{x_i\}_{i=1}^{m}$.
These fields capture visual changes at multiple semantic scales while tying each label to the actual camera motion.
We unroll each labeled transition into $m$ examples $(I_s, x_i, g, T_{gs})$, one per intent.
The resulting training set contains approximately \SI{3}{M} intent-conditioned examples from RoomTour3D~\cite{han2025roomtour3d} and Nymeria~\cite{ma2024nymeria}, covering room-scale walkthroughs and body-scale egocentric interactions; full prompt details are provided in Appendix~\ref{sec:supp-dataset-curation}.

\subsection{Training and Inference}
\label{sec:method_training}

For each training tuple $(I_s, x_i, g, T_{gs})$, we apply teacher-forced next-token prediction to $g$ and clean-target flow matching to $T_{gs}$, optimizing:
\begin{equation}
    \mathcal{L}
    =
    \mathcal{L}_{\mathrm{gain}}
    +
    \lambda_{\mathrm{pose}}\mathcal{L}_{\mathrm{pose}},
    \qquad
    \mathcal{L}_{\mathrm{gain}}
    =
    -\sum_k \log p_\theta(g_k \mid g_{<k}, I_s, x_i).
\end{equation}
Together, the losses keep the backbone aligned with language generation while training the continuous head for the same intent-conditioned motion.
We instantiate Qwen3-VL-4B-Instruct \cite{Qwen3-VL}, freeze the vision encoder, and train the multimodal projector, language model, and flow-matching head; the pose loss uses detached VLM hidden states, updating only the flow head and VLM-to-pose projection while the backbone is updated by the observation-gain loss.
At inference, we autoregressively generate $\hat{g}$ and reuse cached gain-token hidden states as the flow-head condition, avoiding a second full VLM forward pass.
We train for one epoch in bf16 precision on 16 NVIDIA GH200 GPUs, taking approximately 30 hours; additional system details are provided in Appendix~\ref{sec:supp-training}.

\section{Benchmark}
\label{sec:benchmark}

\begin{table}[!t]
  \centering
  \resizebox{\textwidth}{!}{
  \begin{tabular}{lcccccccc}
    \toprule
    \multicolumn{1}{c}{\multirow{2}{*}{\textbf{Method}}}
    & \multicolumn{2}{c}{\textbf{Target-approaching}}
    & \multicolumn{2}{c}{\textbf{Exploration}}
    & \multicolumn{2}{c}{\textbf{Perspective-shift}}
    & \multicolumn{2}{c}{\textbf{All}} \\
    \cmidrule(lr){2-3}
    \cmidrule(lr){4-5}
    \cmidrule(lr){6-7}
    \cmidrule(lr){8-9}
    & \textbf{SR} & \textbf{CA-SR}
    & \textbf{SR} & \textbf{CA-SR}
    & \textbf{SR} & \textbf{CA-SR}
    & \textbf{SR} & \textbf{CA-SR} \\
    \midrule
    JanusVLN & \third{$2.0 \pm 0.0$} & \third{$1.3 \pm 0.0$} & \second{$34.5 \pm 0.6$} & \second{$27.5 \pm 0.6$} & \third{$31.3 \pm 0.6$} & \third{$31.3 \pm 0.6$} & \third{$21.9 \pm 0.2$} & \second{$19.3 \pm 0.3$} \\
    Uni-NaVid & $0.0 \pm 0.0$ & $0.0 \pm 0.0$ & $13.4 \pm 2.5$ & $11.0 \pm 1.8$ & $26.5 \pm 1.4$ & $26.2 \pm 1.3$ & $12.6 \pm 0.8$ & $11.8 \pm 0.5$ \\
    VG-AVS & \second{$8.6 \pm 0.0$} & \second{$6.6 \pm 0.0$} & \third{$30.0 \pm 0.9$} & \third{$11.7 \pm 0.9$} & \second{$36.4 \pm 0.4$} & \second{$33.1 \pm 1.0$} & \second{$24.3 \pm 0.2$} & \third{$16.5 \pm 0.0$} \\
    VLMnav & $0.0 \pm 0.0$ & $0.0 \pm 0.0$ & $4.5 \pm 0.7$ & $3.8 \pm 0.7$ & $8.4 \pm 0.6$ & $8.4 \pm 0.6$ & $4.1 \pm 0.3$ & $3.8 \pm 0.3$ \\
    \textbf{Ours} & \best{$45.8 \pm 1.6$} & \best{$31.8 \pm 2.2$} & \best{$51.4 \pm 0.6$} & \best{$32.6 \pm 1.4$} & \best{$45.8 \pm 1.1$} & \best{$39.4 \pm 1.0$} & \best{$47.7 \pm 0.1$} & \best{$34.4 \pm 0.2$} \\
    \bottomrule
  \end{tabular}
  }
  \caption{Success rate (SR, \%) and collision-aware success rate (CA-SR, \%) under the shared motion budget. CA-SR additionally requires the trajectory to stay at least $0.15$ m from occupied space. Cells report mean $\pm$ standard deviation over three runs. Colored backgrounds indicate \legendbox{bestcolor}{best}, \legendbox{secondcolor}{second}, and \legendbox{thirdcolor}{third} results.}
  \label{tab:main_results}
  \vspace{-7mm}
\end{table}

% Existing evaluation protocols do not directly measure vision-language conditioned camera motion generation.
% Navigation benchmarks entangle camera motion with long-horizon path following and goal reaching. Active perception and next-best-view benchmarks usually assume task-specific utilities, reconstruction objectives, or discrete candidate views.
% In contrast, our setting asks whether a model can predict a local 3D target camera pose from a single current observation and a language intent, such that the next view reveals intent-relevant visual evidence.
% We therefore construct a dedicated benchmark that isolates this capability while spanning different spatial scales, motion purposes, and semantic granularities. ~\todo{This part seems quite similar t o the related work section on why we need such a benchmark?}

To evaluate intent-conditioned camera motion at the granularity we study, the benchmark needs three properties: (R1) free-viewpoint photorealistic rendering over continuous $SE(3)$ poses; (R2) diverse intent coverage across spatial scales; and (R3) outcome-level success measures under a shared motion budget, so methods with different action interfaces and stopping behaviors remain comparable. 
These requirements motivate a benchmark design that combines photorealistic continuous-view rendering with intent annotations and a unified evaluation protocol.
%Sec.~\ref{sec:bench_design} instantiates (R1) and (R2) on InteriorGS with manually authored intent annotations across three intent families, and Sec.~\ref{sec:bench_setup} specifies the closed-loop protocol for (R3).

\subsection{Benchmark Design}
\label{sec:bench_design}

% We instantiate the benchmark on InteriorGS~\cite{InteriorGS2025}, which contains $1$K diverse real-world indoor scenes represented as 3DGS and supports photorealistic rendering from arbitrary camera poses and intrinsics. We sample candidate start views across the scene collection and manually construct local camera-motion tasks by moving the camera to an intent-relevant reference view and authoring a language intent. Each example contains a start observation $I_s$, a language intent $x$, a held-out goal reference observation $I_g$, and the underlying camera poses needed by the renderer. At inference time, models receive only $(I_s,x)$; $I_g$ is used only for evaluation, and success is defined by acquiring an intent-satisfying view rather than by matching the annotated goal pose exactly.

% The benchmark contains $425$ examples across three intent families. \emph{Target-approaching} asks the agent to move toward a specified visible or partially visible target, ranging from furniture-scale objects to small tabletop objects. \emph{Exploration} asks the agent to acquire evidence that is not sufficiently visible from the start view, often by following spatial cues such as doors, corridors, room boundaries, or partial glimpses. \emph{Perspective-shift} asks the agent to change viewpoint around an object or region, including revealing occluded content, inspecting spatial relations, or adjusting distance for a more informative view.

We instantiate the benchmark on InteriorGS~\cite{InteriorGS2025}, a $1$K-scene indoor 3DGS dataset that supports photorealistic rendering from arbitrary camera poses and intrinsics. We sample start views across scenes, construct benchmark examples by moving the camera to an intent-relevant reference view, author language intents, and assign per-example camera intrinsics randomly sampled within a realistic range. Each start--reference pair contains $(I_s,\mathcal{X}_{s,g},I_g)$ plus the underlying camera poses and intrinsics; at inference, models receive only $(I_s,x)$, $x \in \mathcal{X}_{s,g}$, and success is defined by acquiring an intent-satisfying view rather than matching the annotated goal pose. The benchmark contains $425$ examples across three intent families. \emph{Target-approaching}: move toward a visible or partially visible target, from furniture-scale to small tabletop objects. \emph{Exploration}: acquire evidence not sufficiently visible at start, often by following spatial cues such as doors, corridors, or room boundaries. \emph{Perspective-shift}: change viewpoint around an object or region to reveal occluded content, inspect spatial relations, or adjust distance for a more informative view. Detailed examples and the intrinsics-sampling protocol can be found in Appendix~\ref{sec:supp-benchmark-construction}.

\begin{figure*}[!t]
  \centering
  \setlength{\tabcolsep}{1.5pt}
  \renewcommand{\arraystretch}{0.92}
  \begin{tabular}{@{}cccc@{}}
    & \parbox[t]{0.315\textwidth}{\centering\scriptsize {\small\textbf{Target-approaching}}\\[1pt]{\ttfamily\frenchspacing Go to see the matches\\on the coffee table.}} & \parbox[t]{0.315\textwidth}{\centering\scriptsize {\small\textbf{Exploration}}\\[1pt]{\ttfamily\frenchspacing Go to see the painting on\\the wall in the living room.}} & \parbox[t]{0.315\textwidth}{\centering\scriptsize {\small\textbf{Perspective-shift}}\\[1pt]{\ttfamily\frenchspacing Look at the area under\\the table in front of you.}} \\[4pt]
    \rotatebox{90}{\makebox[0.137\textwidth][c]{\scriptsize JanusVLN}} & \includegraphics[width=0.315\textwidth]{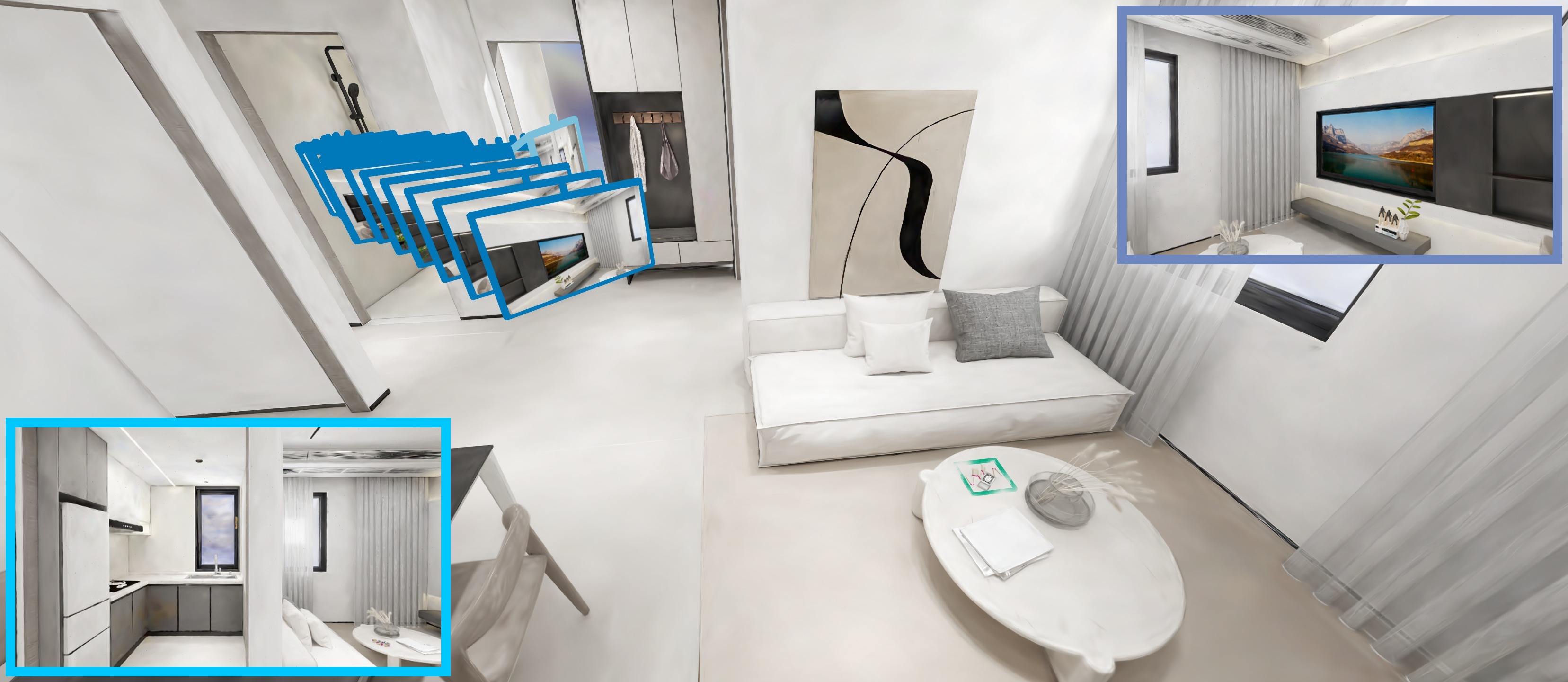} & \includegraphics[width=0.315\textwidth]{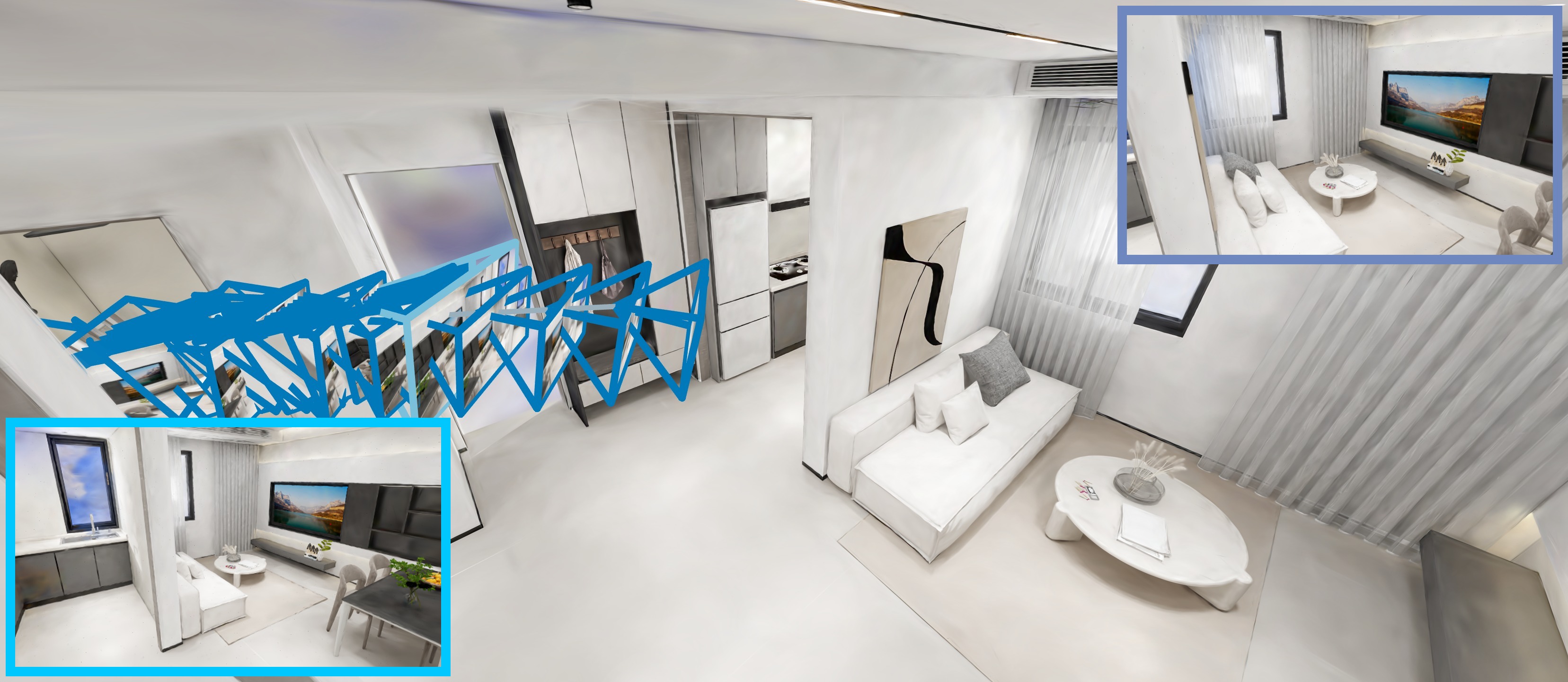} & \includegraphics[width=0.315\textwidth]{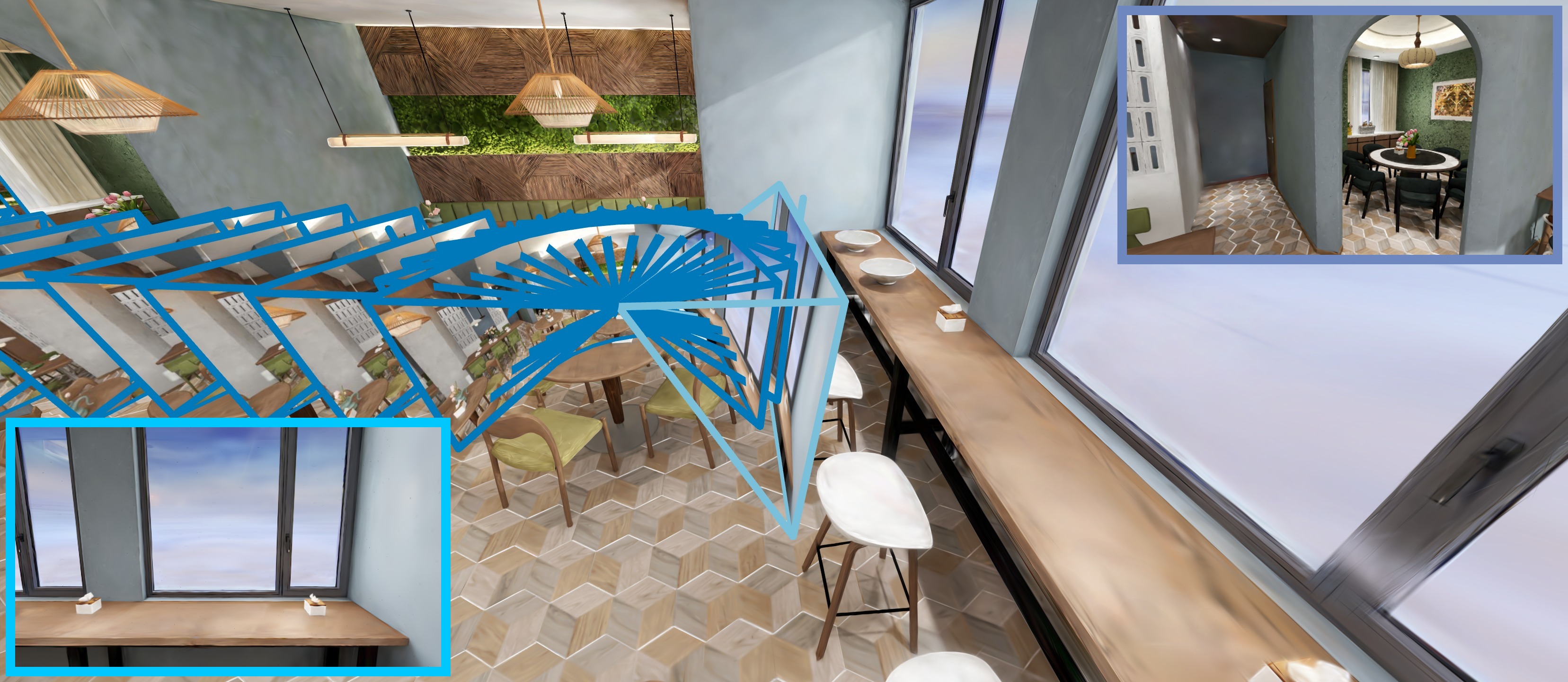} \\
    \rotatebox{90}{\makebox[0.137\textwidth][c]{\scriptsize Uni-NaVid}} & \includegraphics[width=0.315\textwidth]{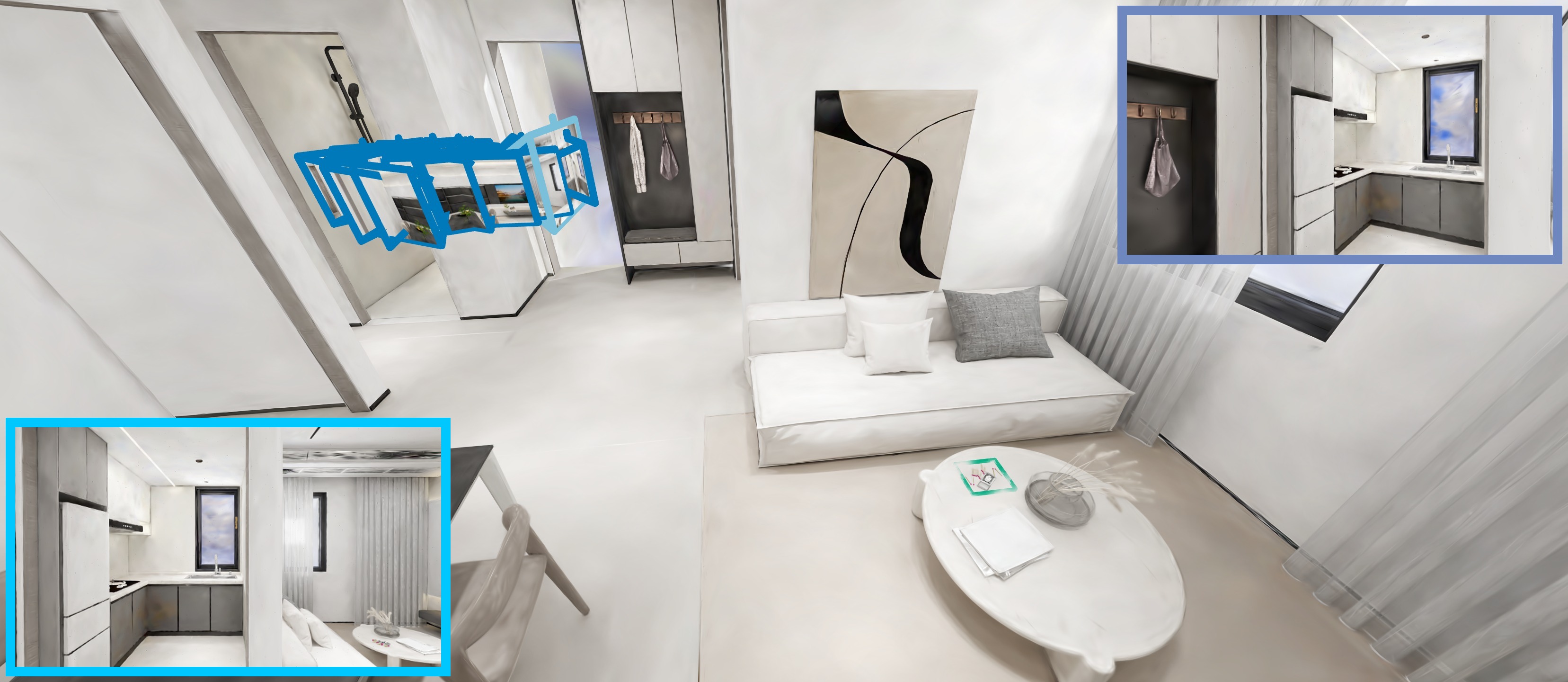} & \includegraphics[width=0.315\textwidth]{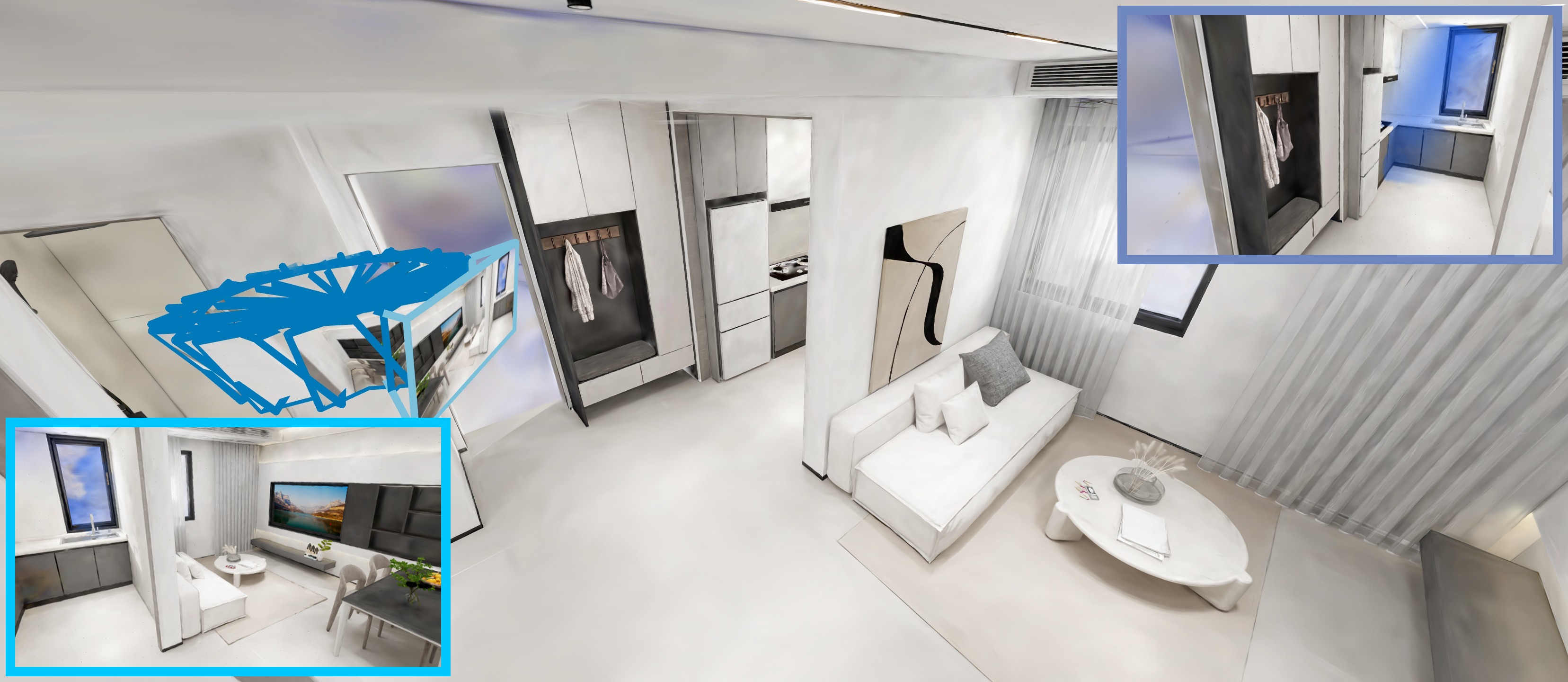} & \includegraphics[width=0.315\textwidth]{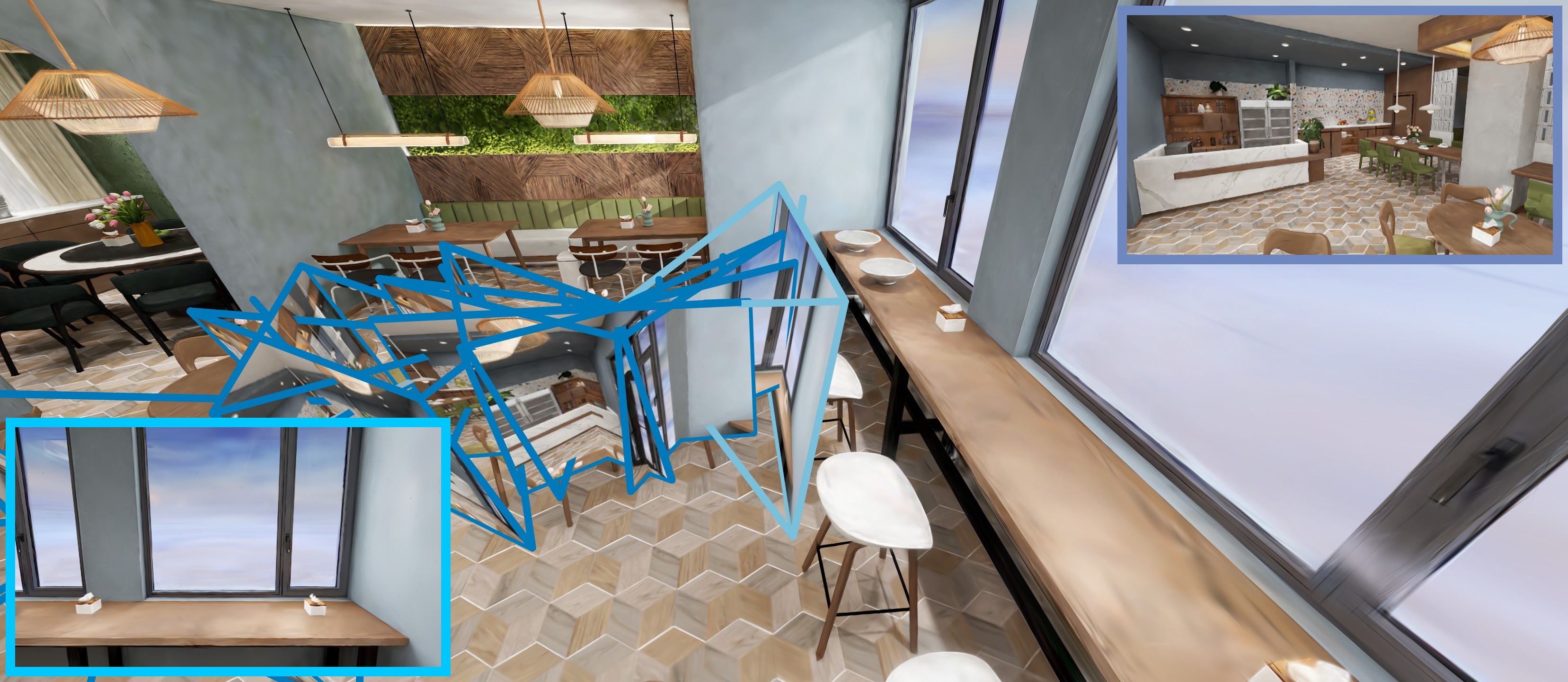} \\
    \rotatebox{90}{\makebox[0.137\textwidth][c]{\scriptsize VG-AVS}} & \includegraphics[width=0.315\textwidth]{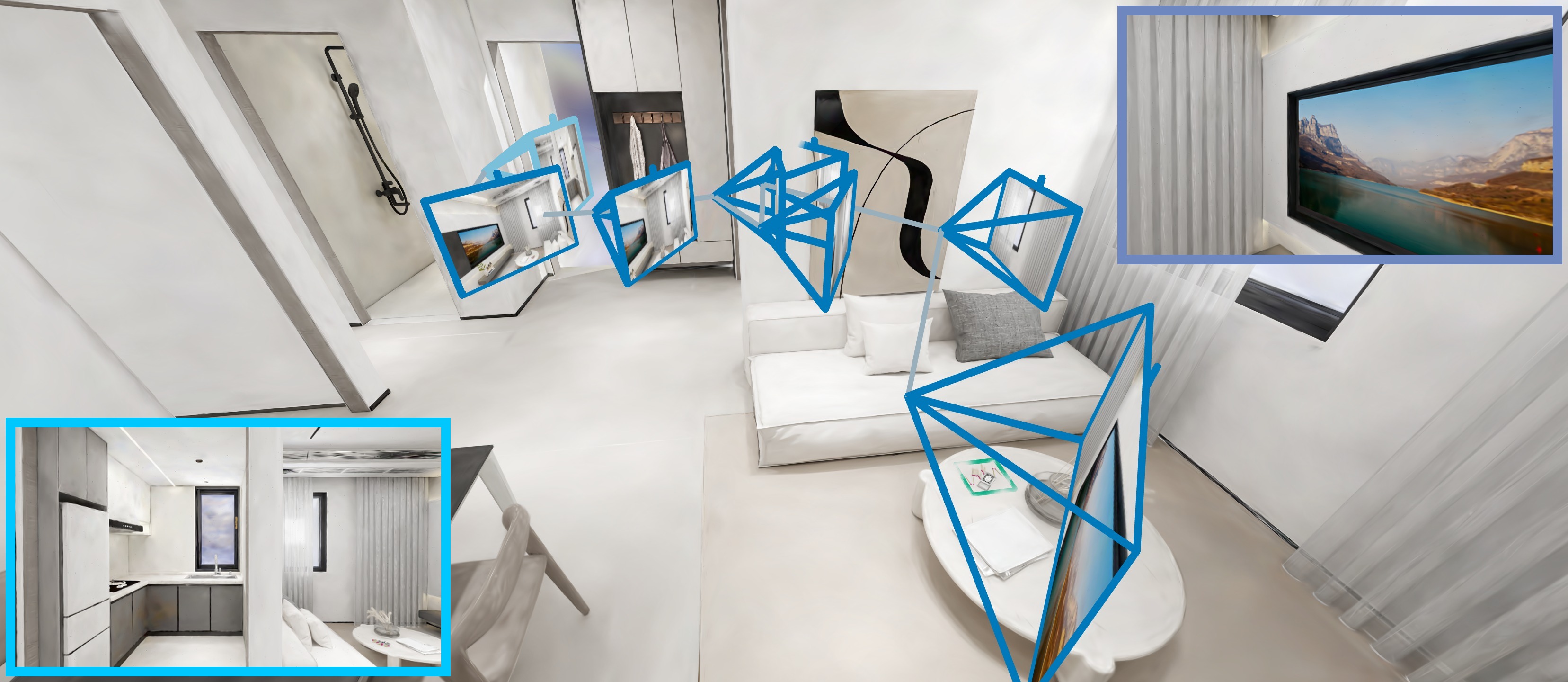} & \includegraphics[width=0.315\textwidth]{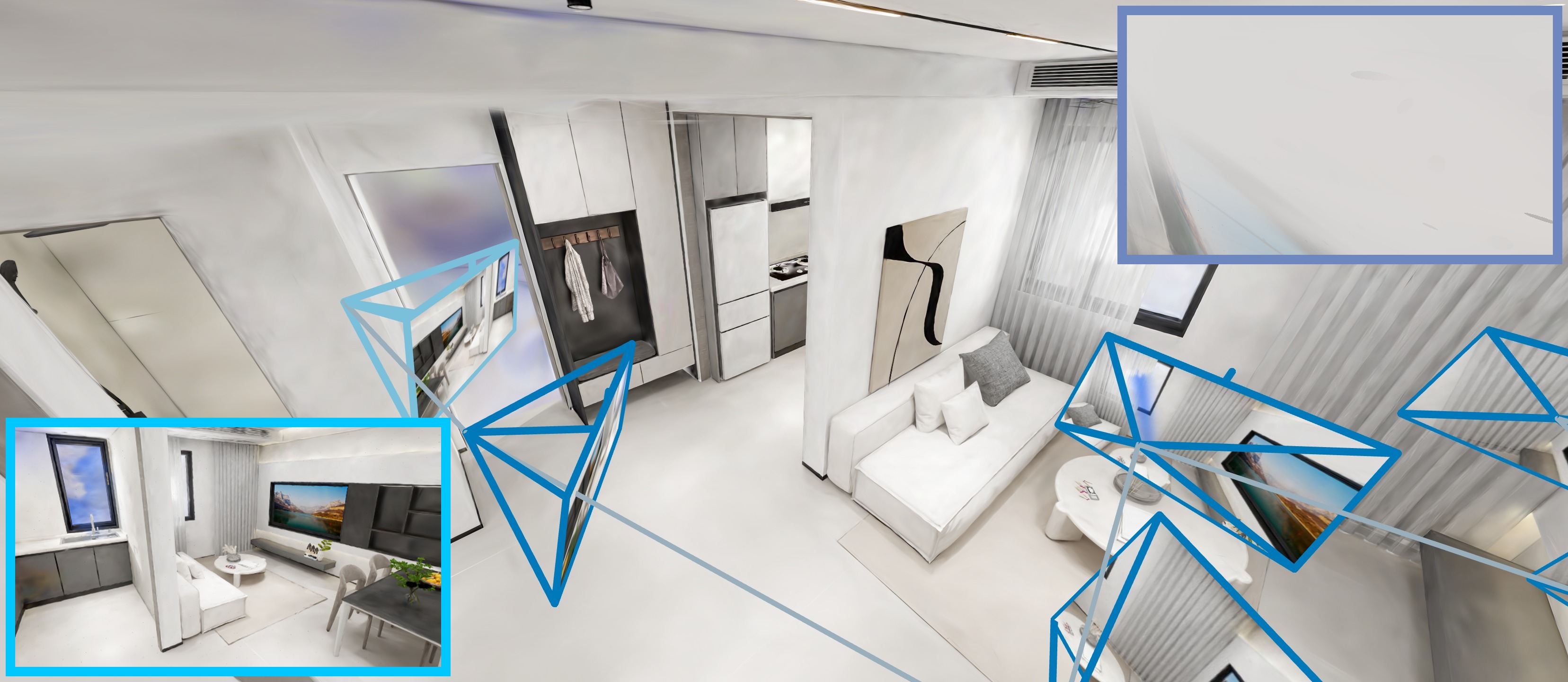} & \includegraphics[width=0.315\textwidth]{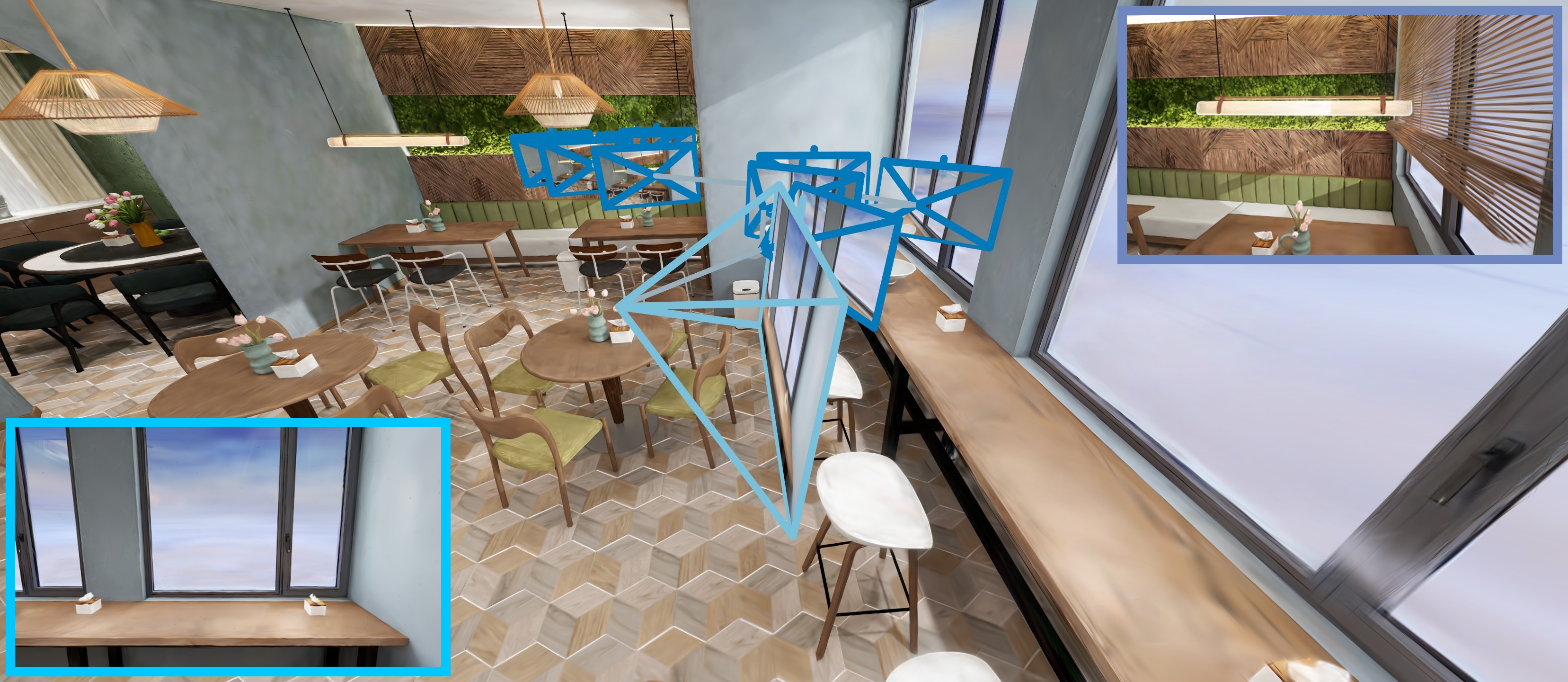} \\
    \rotatebox{90}{\makebox[0.137\textwidth][c]{\scriptsize VLMnav}} & \includegraphics[width=0.315\textwidth]{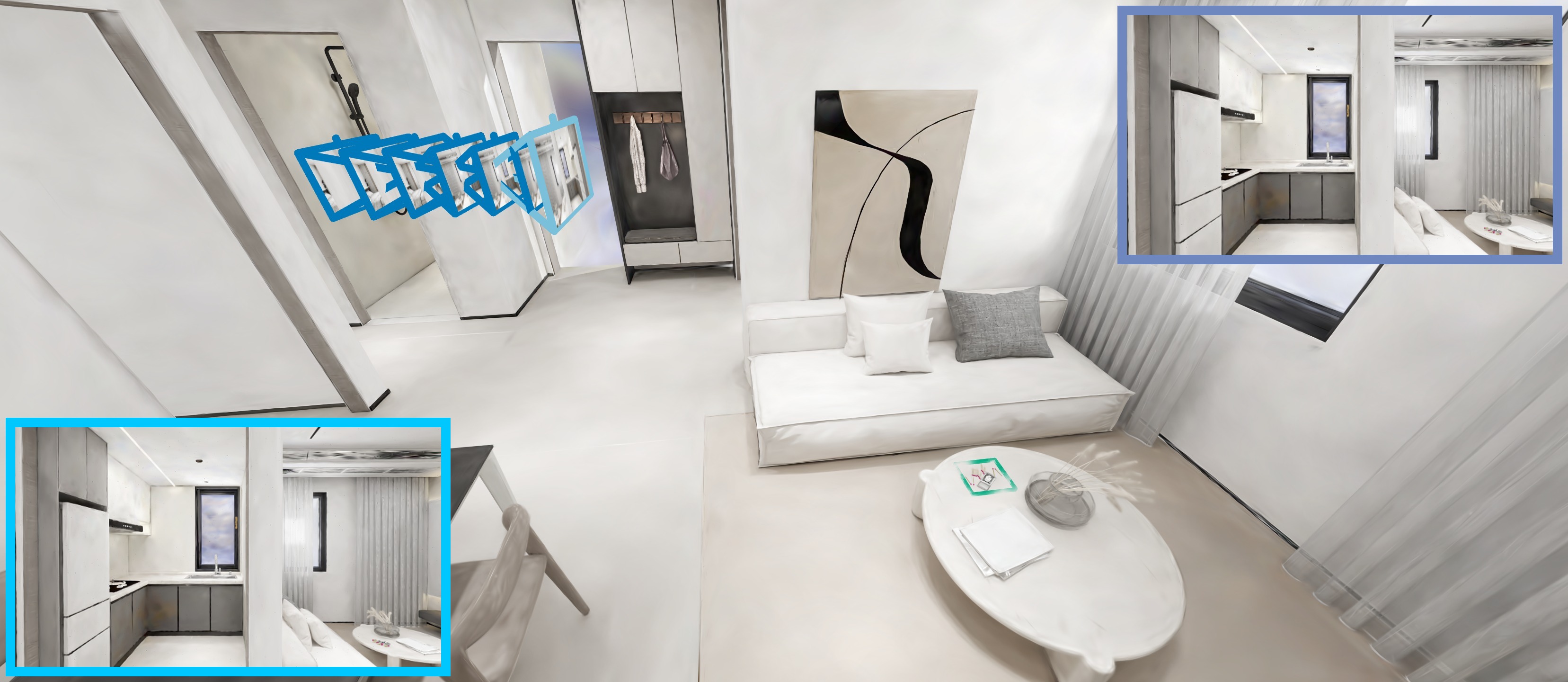} & \includegraphics[width=0.315\textwidth]{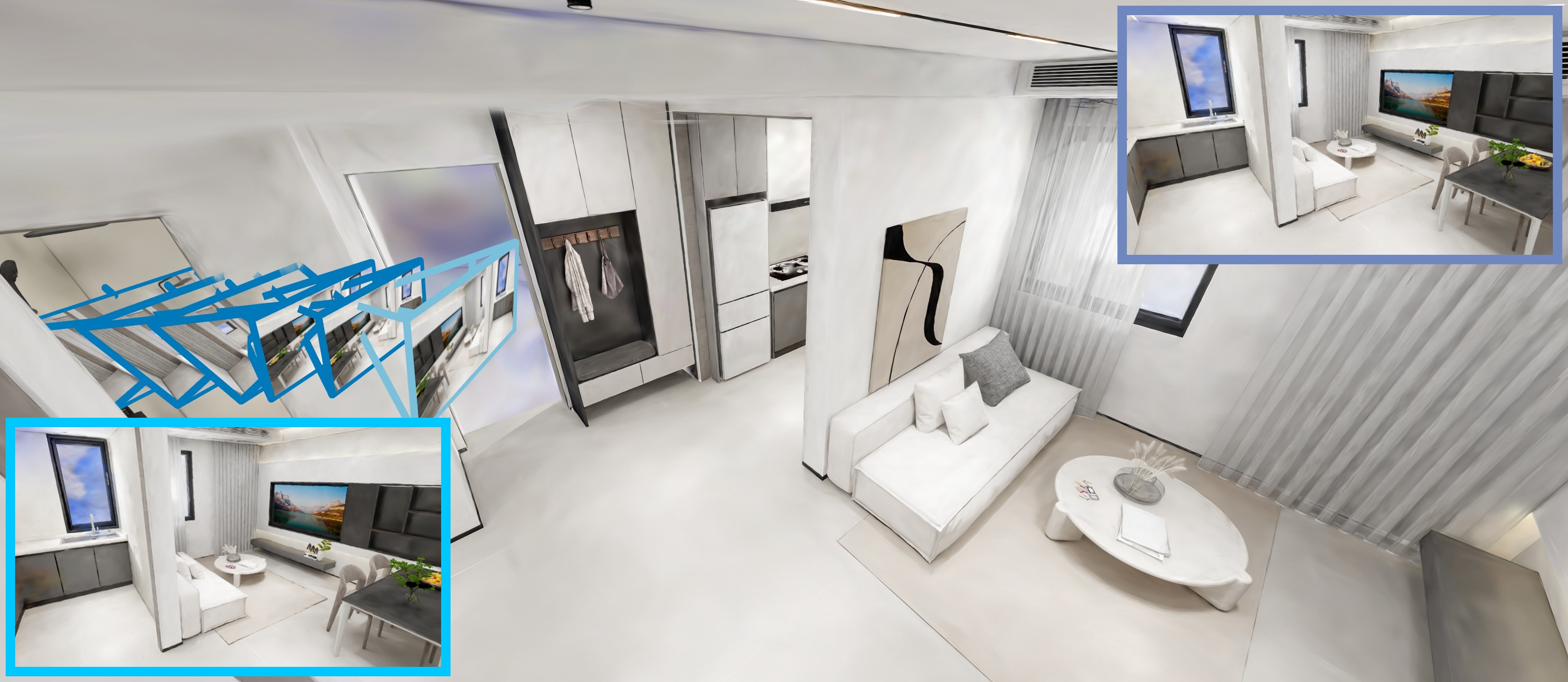} & \includegraphics[width=0.315\textwidth]{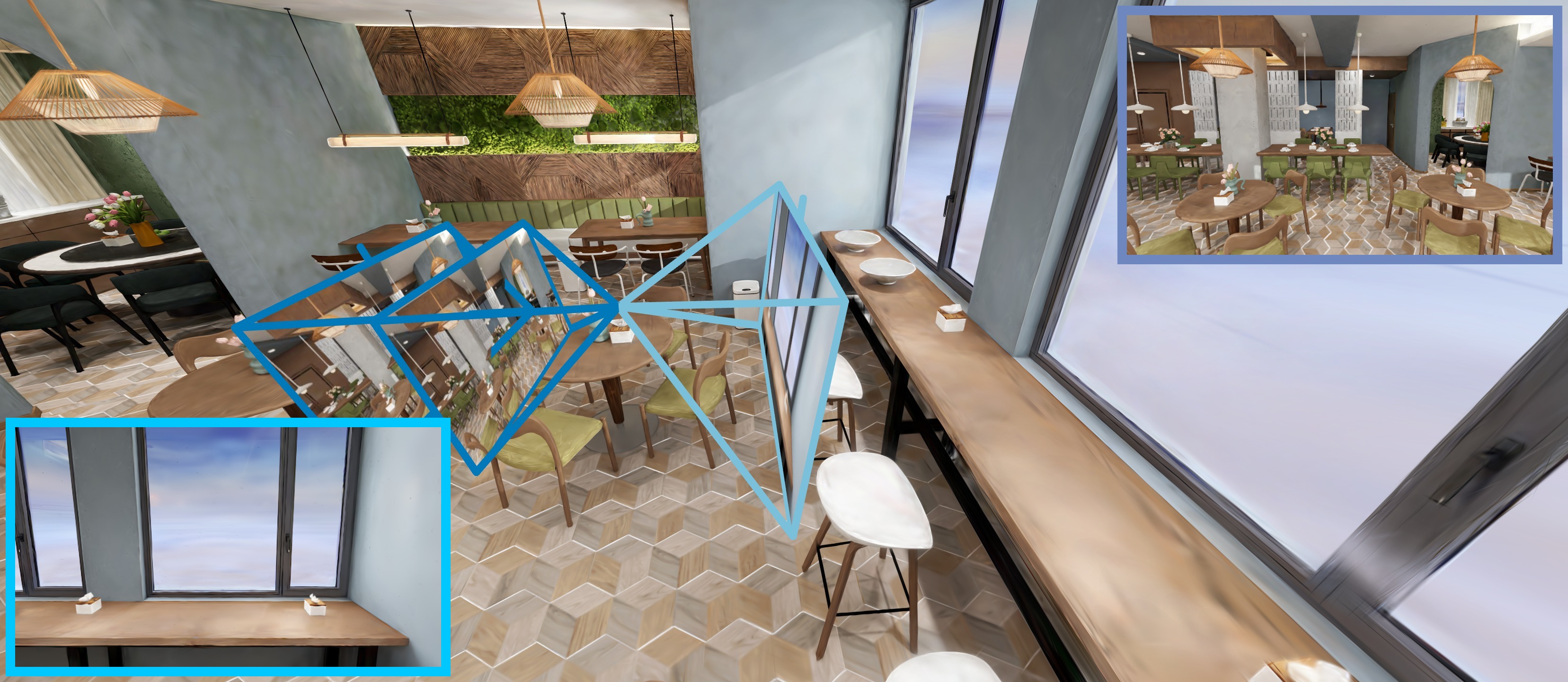} \\
    \rotatebox{90}{\makebox[0.137\textwidth][c]{\scriptsize \textbf{Ours}}} & \includegraphics[width=0.315\textwidth]{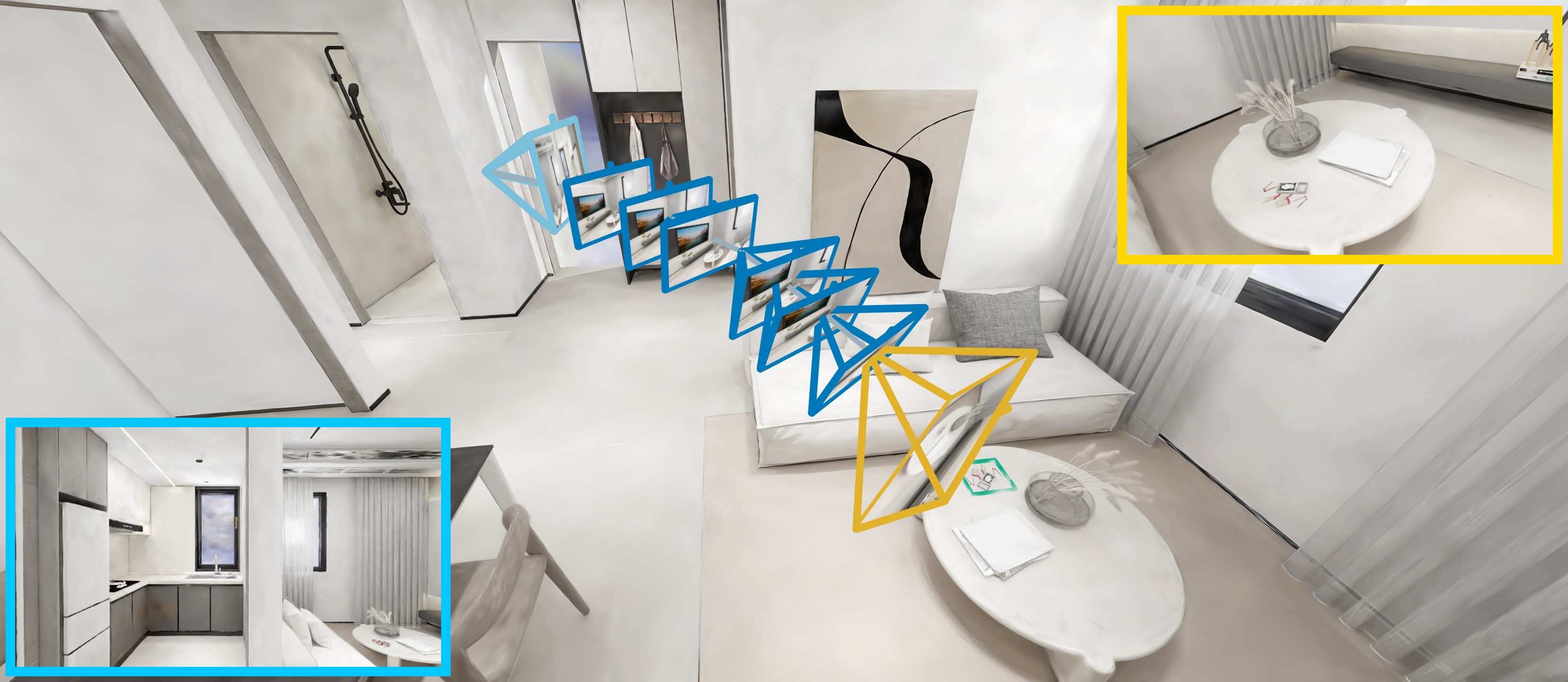} & \includegraphics[width=0.315\textwidth]{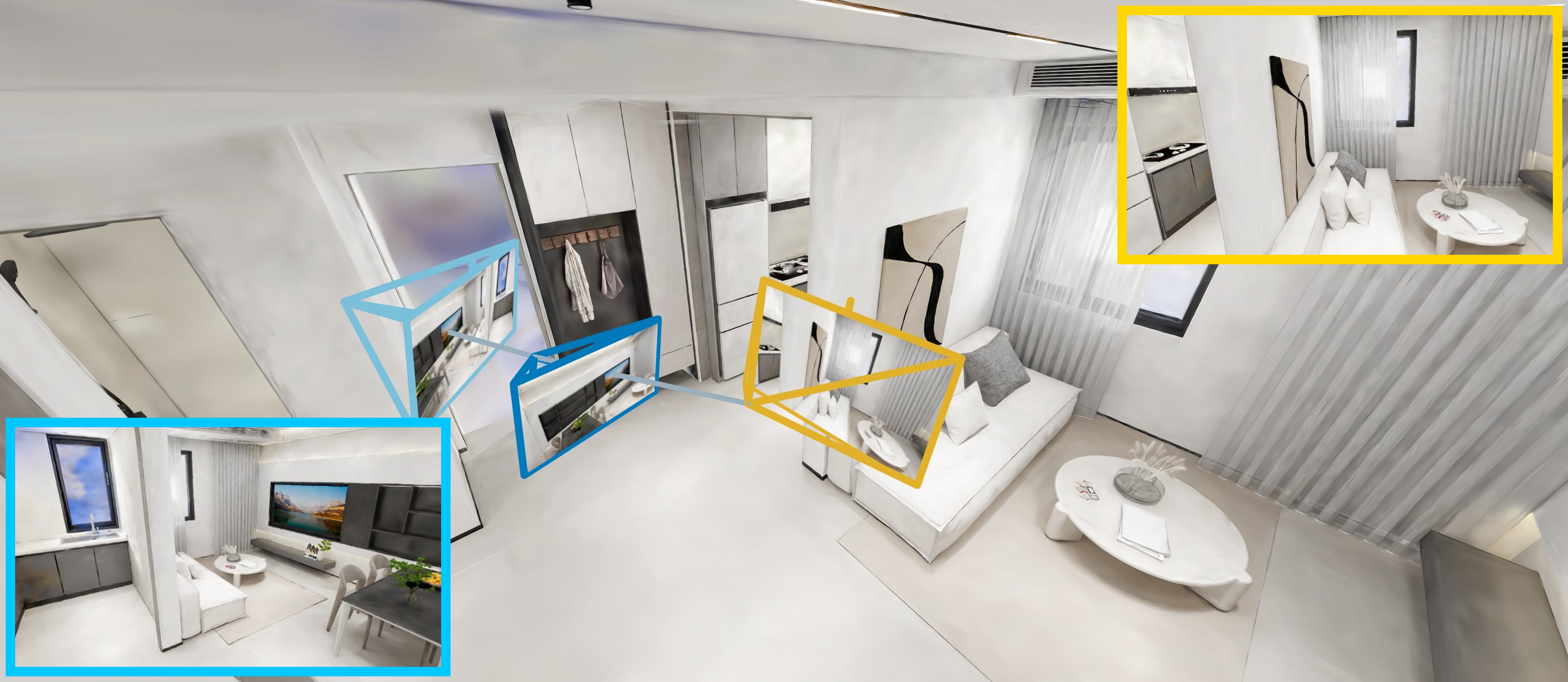} & \includegraphics[width=0.315\textwidth]{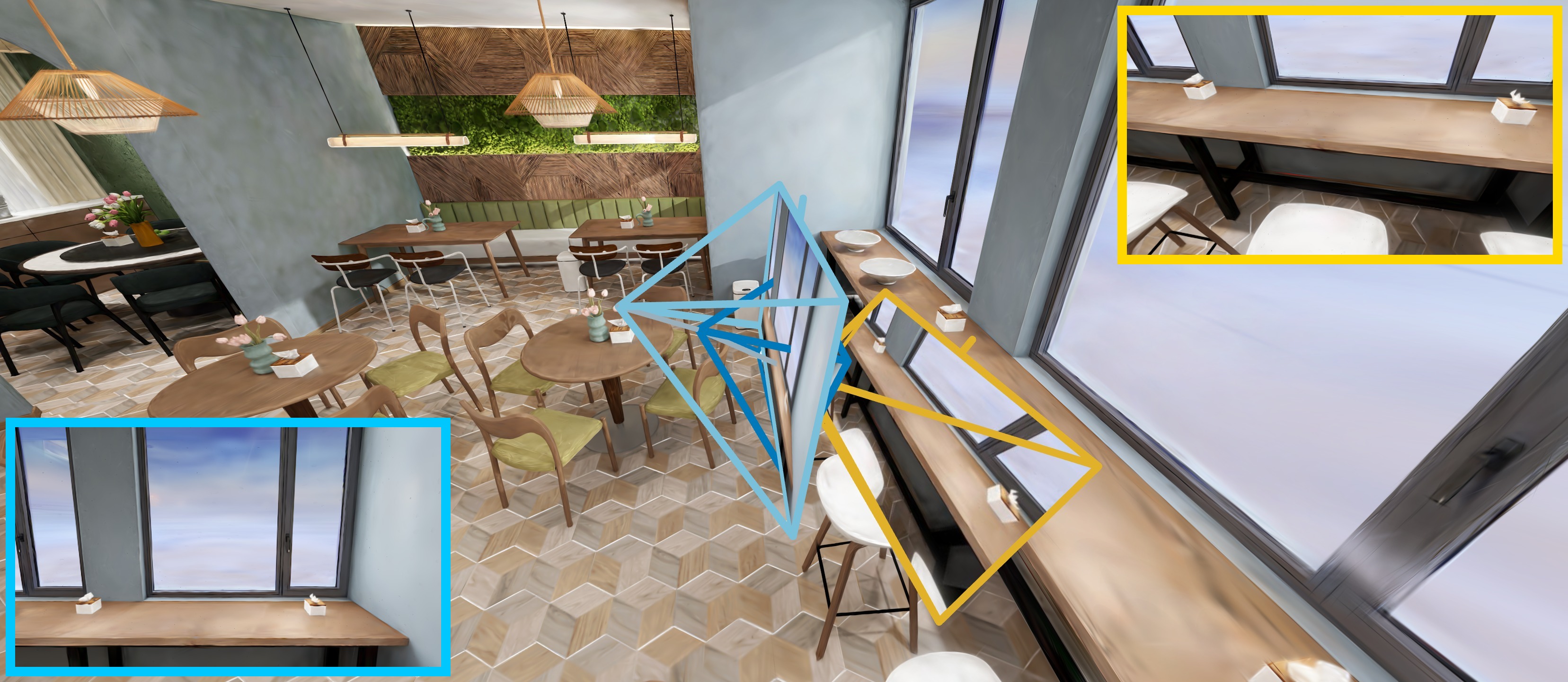} \\
  \end{tabular}
  \vspace{-2mm}
  \caption{Qualitative Comparisons. Rows compare methods and columns show intent families with their language intents. Cyan insets show the shared start view; top-right insets show the first successful frame (yellow), or the final frame when no success is reached within the motion budget.}
  \label{fig:manual_pair_qualitative}
  \vspace{-5mm}
\end{figure*}

\subsection{Evaluation Setup}
\label{sec:bench_setup}

To accommodate methods with different action interfaces and intents at different spatial granularities, we evaluate all methods with a budgeted multi-step protocol, even when the desired outcome may require only a local viewpoint adjustment. Each method outputs a camera motion from the current observation and intent, the renderer applies it and returns the next view, and the process repeats until a stop signal or until the shared motion budget is reached, equivalent to $6$~m of pure translation or $600^{\circ}$ of pure rotation. Methods without a stop signal run until the budget is reached. 
An evaluation trajectory succeeds if any rendered view before termination satisfies the intent. For Exploration and Perspective-shift, Gemini-3.1-Pro-Preview~\cite{gemini31pro} judges the candidate frame against the intent with $I_g$ as a non-exclusive reference. For Target-approaching, success requires geometric proximity to the target object's InteriorGS bounding box and Gemini-verified visual recognizability of the target. Collision-aware success rate further requires the trajectory prefix to first success to remain $0.15$~m clear of the scene point cloud. A balanced human-labeled subset audits the automatic judge, with agreement statistics reported in Sec.~\ref{sec:exp_results}. Gemini prompt templates and target-proximity thresholds can be found in Appendix~\mbox{\ref{sec:supp-judge-criteria}--\ref{sec:supp-target-criteria}}.

\section{Experiments}
\label{sec:experiment}

\subsection{Experimental Setup}
\label{sec:exp_setup}

% Previous four-question framing:
% Our experiments test four questions: (a) whether the model outperforms diverse baselines on intent-relevant 3D camera-pose prediction; (b) whether this capability generalizes across intent families and deployment settings; (c) whether it benefits downstream embodied tasks; and (d) how key component designs affect performance.
Our experiments aim to answer three questions: (a) whether LIME outperforms diverse baselines on intent-relevant 3D camera-pose prediction; (b) whether its capability generalizes across intent families and deployment settings; and (c) whether it benefits downstream embodied tasks, tested on LIBERO-Goal manipulation and a real robot. On the proposed benchmark, each method receives the same start image $I_s$ and language intent $x$ in its adapted input format, runs in the same budgeted multi-step protocol, and is evaluated by success rate under the shared motion budget. The goal image $I_g$ is withheld from the model and used only for evaluation. We report success rate (SR) and collision-aware success rate (CA-SR) per intent family and overall, following the success protocol defined in Sec.~\ref{sec:bench_setup}. Given the limited availability of public implementations under exactly matched assumptions, we compare against closest representative open-source methods spanning language-conditioned navigation, active VQA/view selection, and zero-shot VLM navigation: JanusVLN \cite{zeng2025janusvln} and Uni-NaVid \cite{zhang2024uninavid}, fine-tuned VLN models for language-instructed navigation; VG-AVS \cite{koo2025toward}, an embodied VLM that actively chooses next views for VQA; and VLMnav \cite{goetting2024end}, a zero-shot VLM-based navigation pipeline. All baselines are adapted to the same renderer, rollout budget, and success metric; further details on baseline adaptation and parameters are provided in Appendix~\ref{sec:supp-baselines}.

\subsection{Results and Discussion}
\label{sec:exp_results}

Table~\ref{tab:main_results} shows that our method achieves the highest success rate across Target-approaching, Exploration, Perspective-shift, and overall, outperforming baselines specialized for language navigation, active view selection, or VLM-based navigation.
This advantage does not stem only from LIME's richer action space: executed in the baselines' planar action frame, LIME still outperforms all baselines overall (Table~\ref{tab:baseline_mode_comparison} in Appendix~\ref{sec:supp-baselines}).
It is worth noting that the model is trained from egocentric video, receives no fine-tuning on benchmark scenes, and still performs strongly in rendered evaluation environments.
The advantage persists under the collision-aware metric.
Figure~\ref{fig:manual_pair_qualitative} illustrates these trends: our method reaches intent-relevant views with fewer evaluation steps than the fine-tuned VLN baselines and uses full 3D target-pose prediction to combine translation and rotation, such as moving closer while tilting to reveal evidence that planar or discrete-action baselines can miss.
As a sanity check, we compare Gemini-based success judgments with human judgments on a balanced subset of $90$ benchmark examples, yielding $450$ trajectory-level labels across five methods. Gemini agrees with human judgments on $91.3\%$ of labels overall, with $85.3$--$98.0\%$ agreement across intent families, and LIME also ranks first under three alternative VLM judges; full values are provided in Appendix~\mbox{\ref{sec:supp-human-judge}--\ref{sec:supp-judge-robustness}}.

\begin{figure}[!t]
  \centering
  \setlength{\tabcolsep}{13pt}
  \renewcommand{\arraystretch}{0.9}
  % One column per scene: start view on top, then four intents with five sampled target poses each.
  \newcommand{\qstart}[1]{\includegraphics[height=0.1216\textwidth]{image/quali_imgs/#1.jpg}}
  \newcommand{\qimg}[1]{\includegraphics[height=0.165\textwidth]{image/quali_imgs/#1.jpg}}
  \newcommand{\qcmd}[1]{{\scriptsize\ttfamily\frenchspacing #1}}
  \begin{tabular}{@{}ccc@{}}
    \qstart{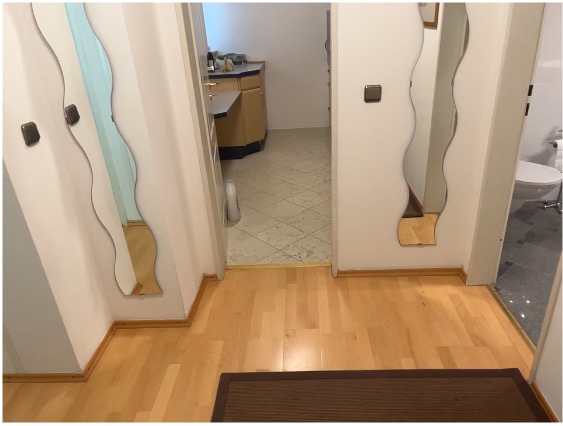} & \qstart{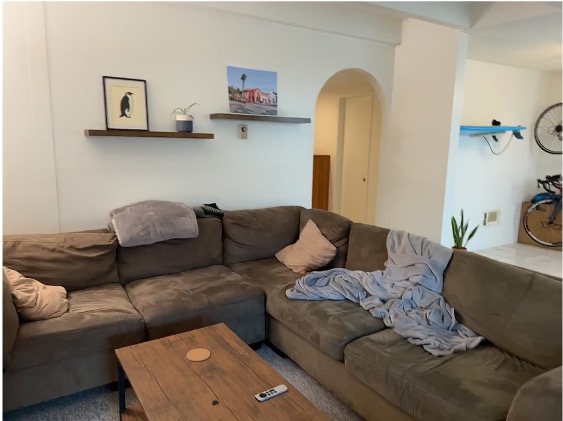} & \qstart{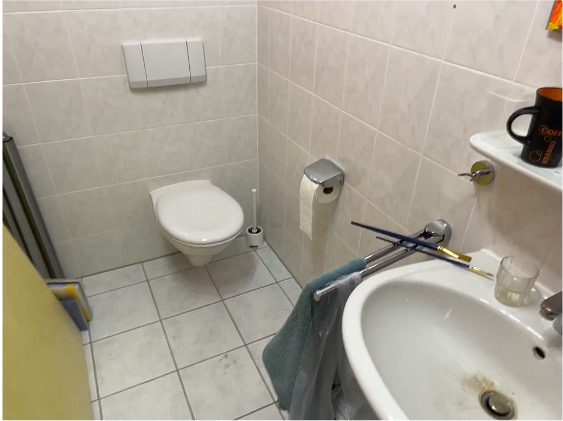} \\[2pt]
    \qcmd{Go to living room} & \qcmd{Find my bike} & \qcmd{Use the towel} \\
    \qimg{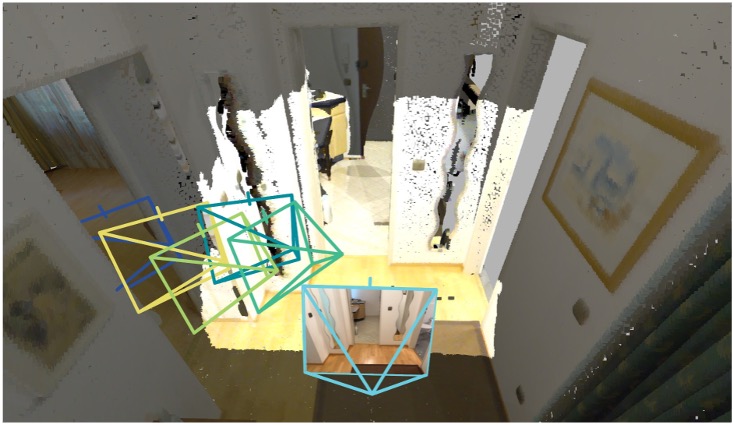} & \qimg{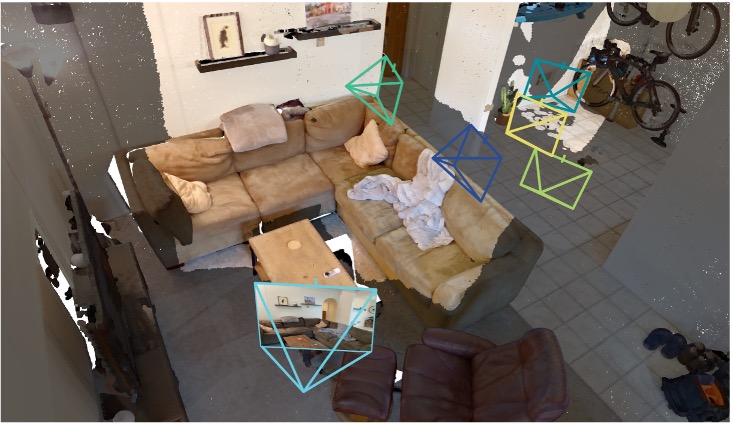} & \qimg{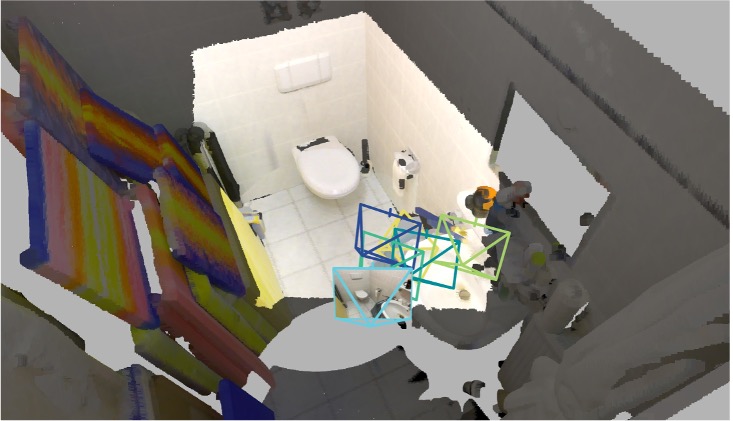} \\[2pt]
    \qcmd{Cook} & \qcmd{Watch TV} & \qcmd{Look at the mirror} \\
    \qimg{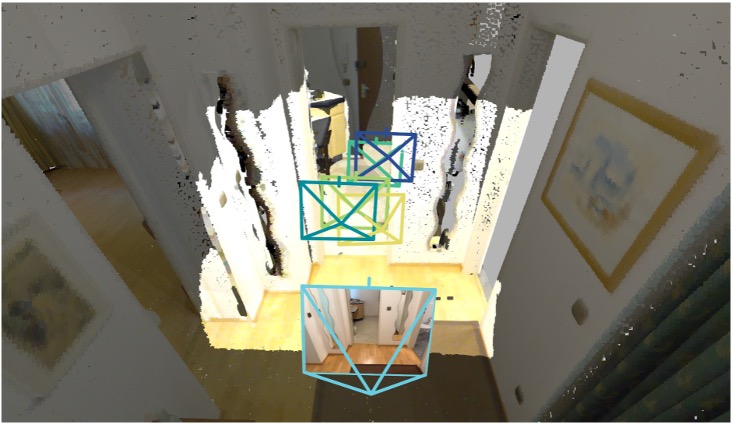} & \qimg{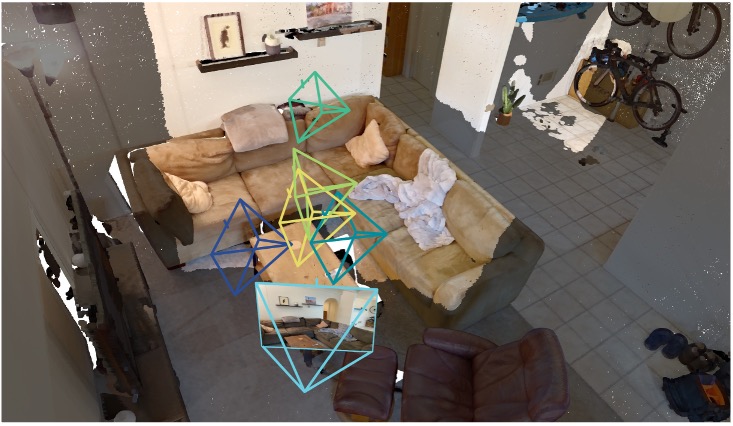} & \qimg{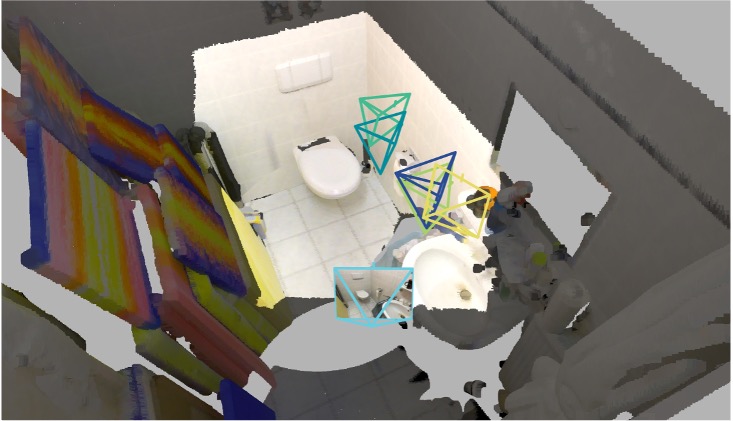} \\[2pt]
    \qcmd{Go to toilet} & \qcmd{What is on the table} & \qcmd{Find toilet} \\
    \qimg{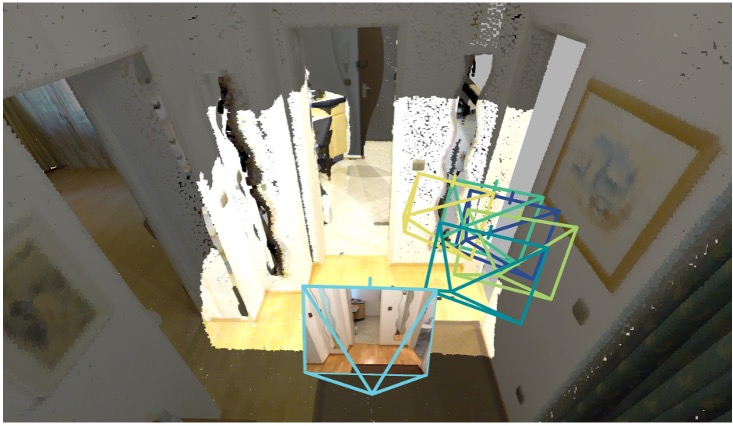} & \qimg{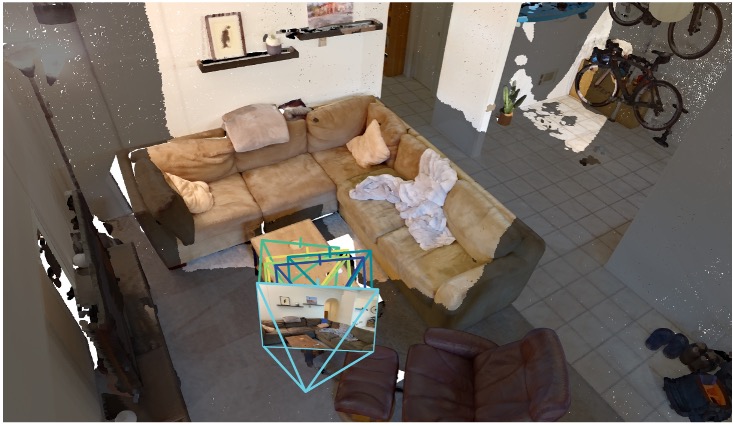} & \qimg{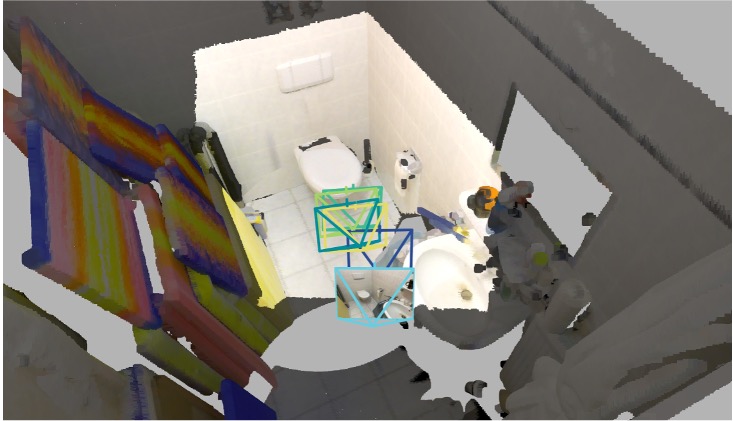} \\[2pt]
    \qcmd{Leave the room} & \qcmd{Look outside} & \qcmd{Leave bathroom} \\
    \qimg{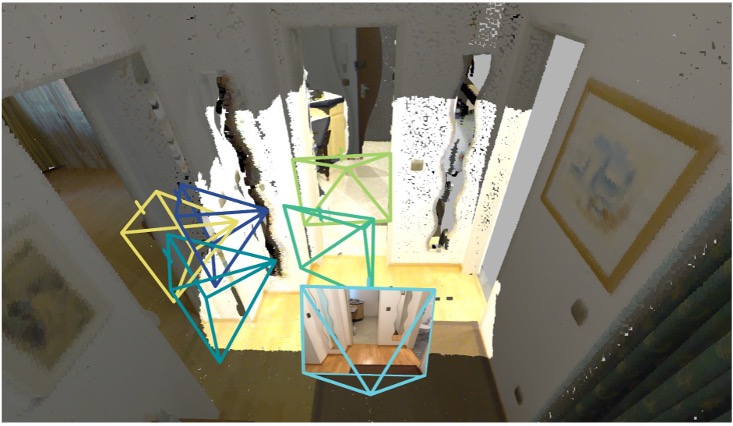} & \qimg{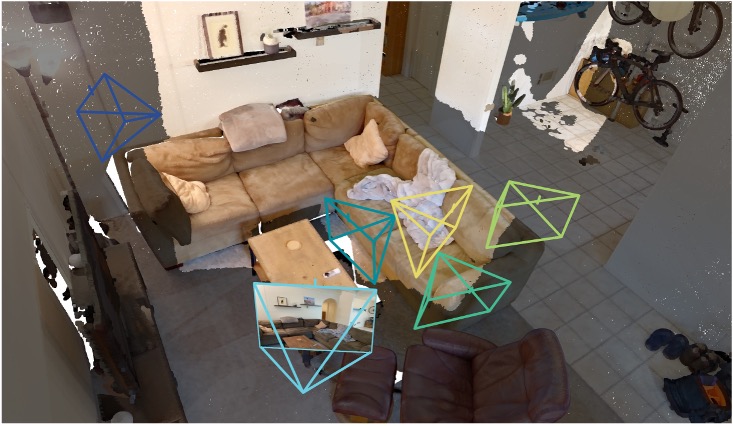} & \qimg{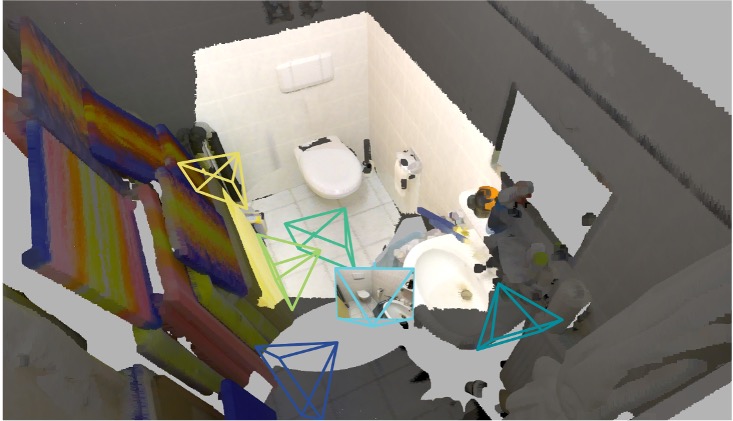} \\
  \end{tabular}
  \vspace{-2mm}
  \caption{Qualitative samples on ScanNet++ scenes. Each column fixes the same current observation (top) and varies the language intent; colored frustums show five sampled target poses from the flow-matching head, illustrating intent-conditioned motion and multimodal pose hypotheses.}
  \label{fig:scannet_qualitative}
  \vspace{-4mm}
\end{figure}

Figure~\ref{fig:scannet_qualitative} further probes generalization to ScanNet++~\cite{yeshwanth2023scannet++} scenes.
For the same current image, changing only the language intent shifts the sampled target poses toward different intent-relevant evidence, indicating that the model conditions on the intent rather than only a scene-level prior.
Our pose generator also reflects ambiguity: samples concentrate when the goal is visually supported in the current view, but spread across multiple plausible directions for ambiguous intents such as leaving a room.

We further deploy our method on a Boston Dynamics Spot with an arm, using RGB images from its hand camera. For real-world robot experiments, we use a lightweight LoRA-adapted checkpoint. The adaptation mainly aligns camera geometry: Spot's hand camera differs considerably in intrinsics and resolution from the training videos, so we fine-tune on about $1.7$K start--goal pairs recorded with Aria glasses and modified to match Spot's camera; details are provided in Appendix~\ref{sec:supp-training}. Figure~\ref{fig:real_robot_qualitative} shows language-conditioned viewpoint changes on physical scenes, such as viewing below an object or checking the region left of an oven. We also integrate the camera-motion policy with VidBot~\cite{chen2025vidbot}, a vision-language-conditioned manipulation trajectory generator that, like most manipulation policies, requires the target object to be visible before acting. When the target is initially outside the field of view, our policy first reveals the task-relevant object or region before VidBot acts, and can also verify outcomes after execution. These results suggest that the adapted camera-motion policy transfers beyond rendered benchmarks and can serve as an active perception module for downstream embodied interaction.

To quantify this effect, we evaluate the main LIME checkpoint as an active-perception front end for a vision-language-action policy on LIBERO-Goal~\cite{liu2023liberobenchmarkingknowledgetransfer}. We sample start states in which the task-relevant object or fixture is only partially visible from the wrist camera, and compare running $\pi_{0.5}$~\cite{intelligence2025pi05visionlanguageactionmodelopenworld} directly with first applying one LIME-predicted wrist-camera motion, prompted with ``Look for the \texttt{<object>}'', before running the same policy. Both pipelines use the same external- and wrist-camera inputs and the same frozen $\pi_{0.5}$ policy. Across $50$ paired initial states (five per task, one rollout per pipeline per state), this single LIME step raises overall success from $26\%$ to $74\%$ in this challenging-start setting and improves all ten tasks. Per-task results and qualitative examples (Table~\ref{tab:libero_goal_manipulation} and Fig.~\ref{fig:libero_goal_active_perception}), as well as results on embodied question answering and multi-step robot tasks, are provided in Appendix~\mbox{\ref{sec:supp-libero}--\ref{sec:supp-multistep}}.

\section{Conclusion}
\label{sec:conclusion}

\begin{figure}[!t]
  \centering
  \setlength{\tabcolsep}{0pt}
  \renewcommand{\arraystretch}{0.9}
  % One row per task: four frames from left to right, then its chained intents; \ract marks commands for the manipulation policy.
  \newcommand{\rimg}[1]{\includegraphics[width=0.2325\textwidth]{image/realrobot_imgs/#1.jpg}}
  \newcommand{\rcmd}[1]{{\scriptsize\ttfamily\frenchspacing #1}}
  \newcommand{\ract}[1]{\textcolor{lightblue}{#1}}
  \begin{tabular*}{\textwidth}{@{\extracolsep{\fill}}cccc@{}}
    \rimg{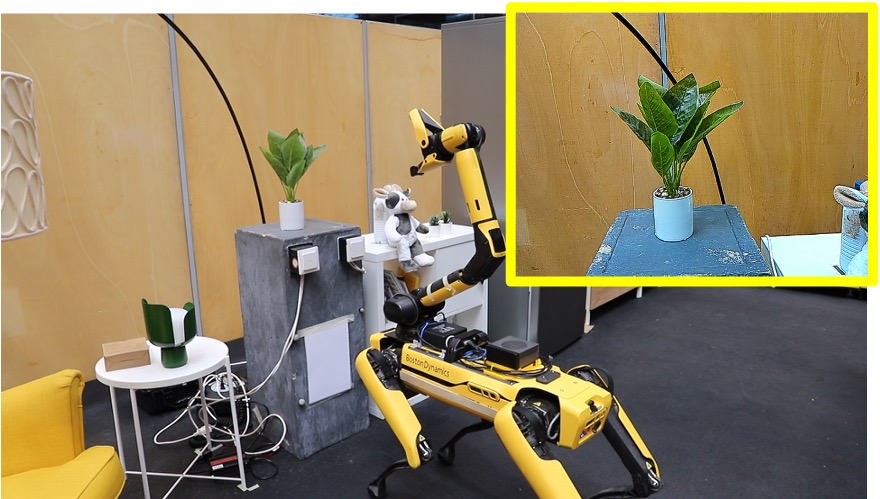} & \rimg{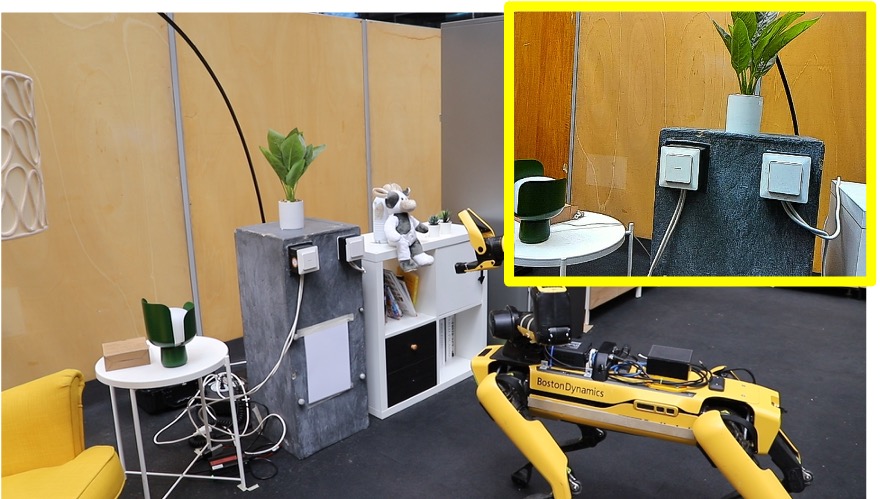} & \rimg{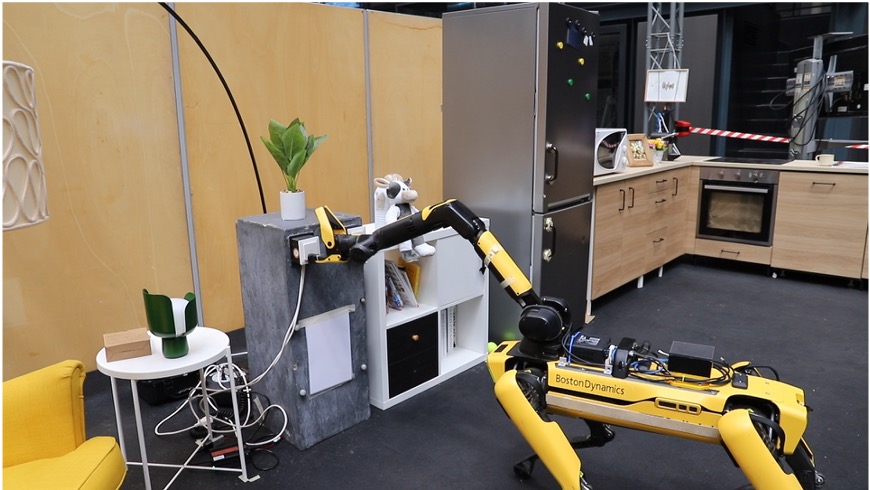} & \rimg{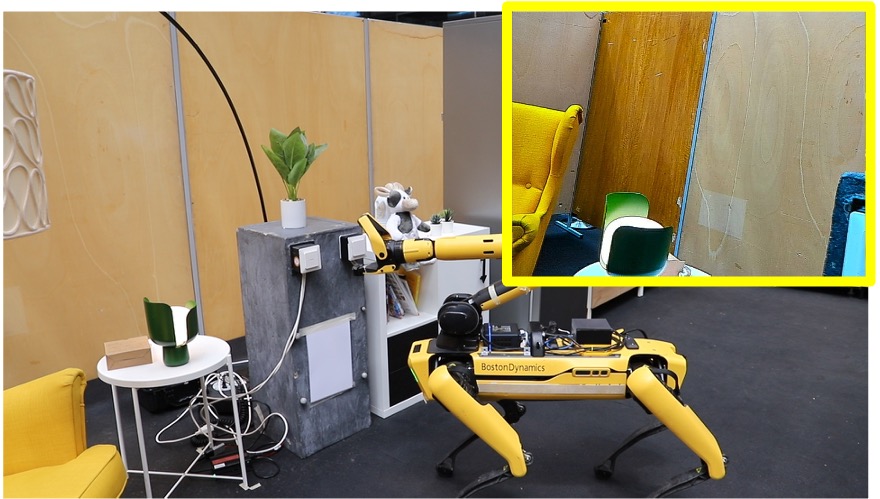} \\
    \multicolumn{4}{c}{\rcmd{Observe the area below -> \ract{Press the light switch} -> Inspect if the lamp turns on}} \\[2pt]
    \rimg{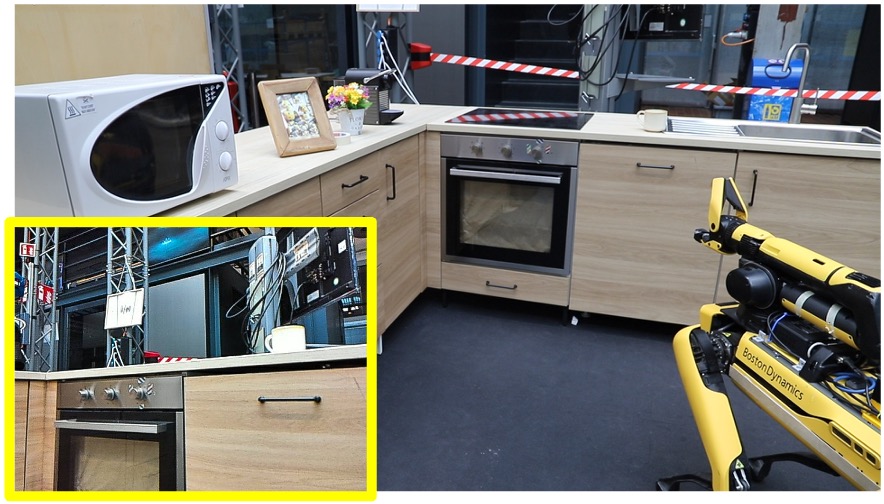} & \rimg{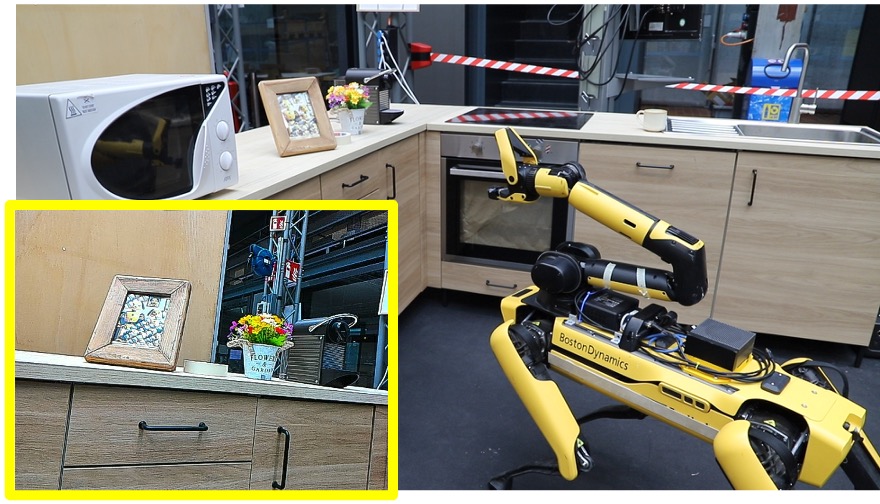} & \rimg{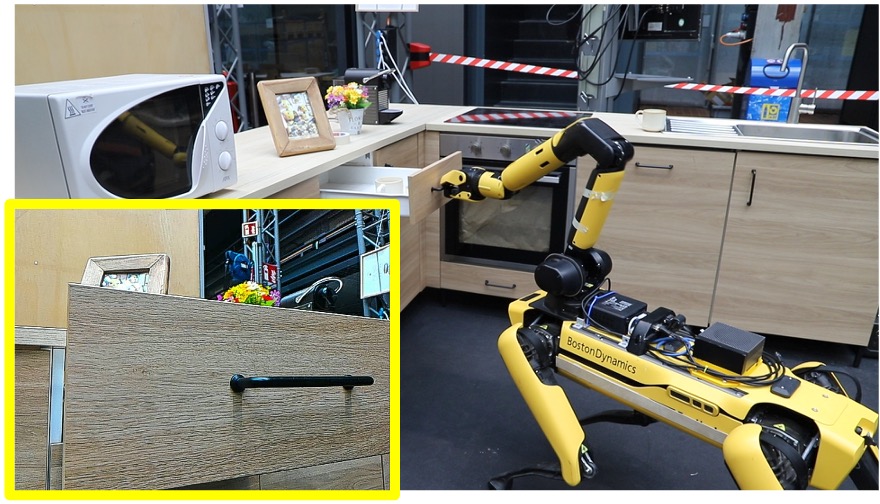} & \rimg{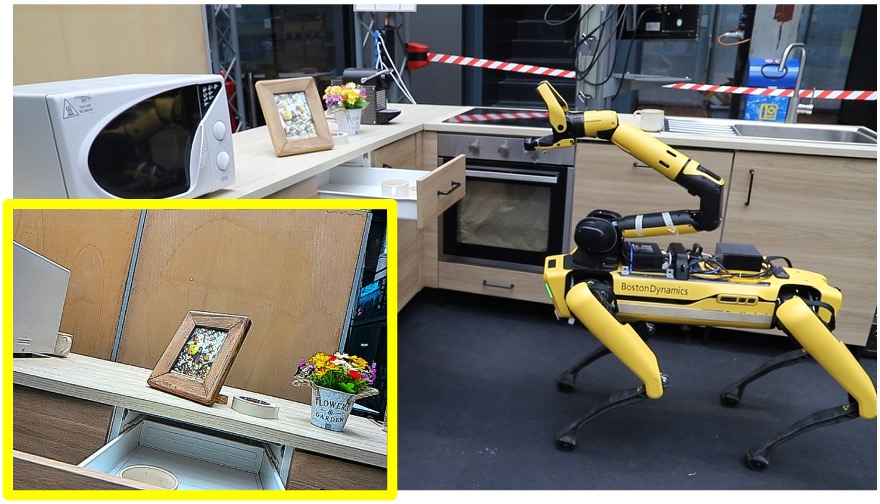} \\
    \multicolumn{4}{c}{\rcmd{What is to the left of the oven? -> \ract{Open the top drawer} -> See what is inside the open drawer}} \\[2pt]
    \rimg{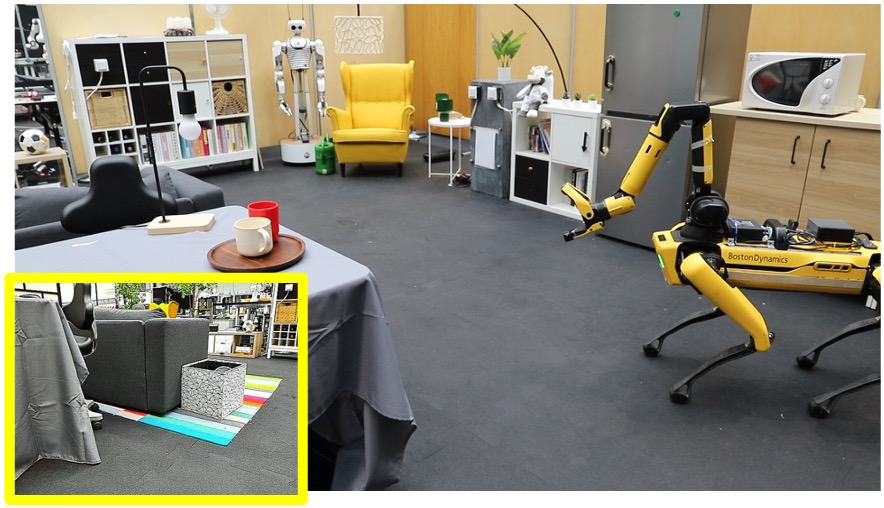} & \rimg{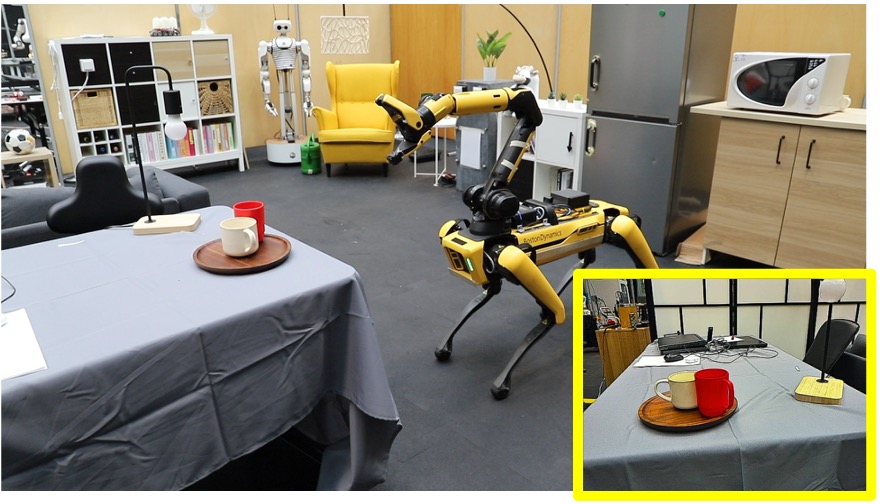} & \rimg{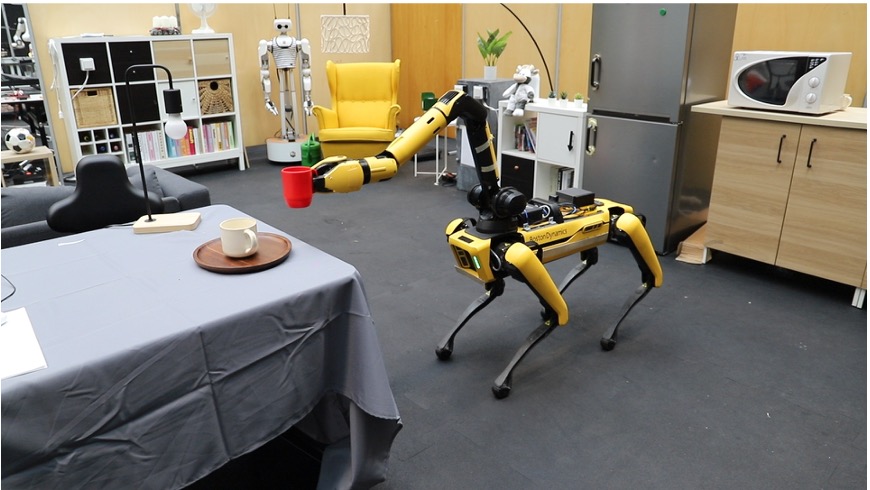} & \rimg{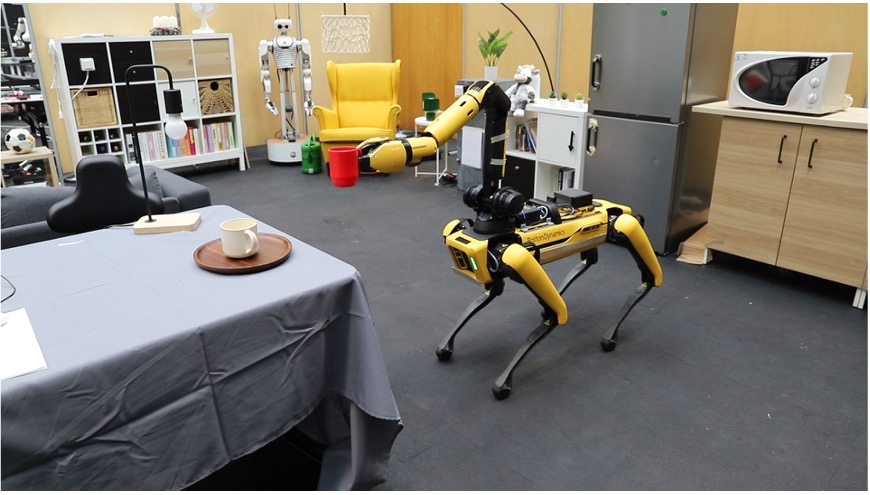} \\
    \multicolumn{4}{c}{\rcmd{What is on the table surface? -> \ract{Take the red mug}}} \\
  \end{tabular*}\par
\caption{Real-robot experiments. The learned camera-motion policy moves the robot camera to acquire visual evidence for chained perception and interaction intents. \textcolor{lightblue}{Blue text} indicates action commands passed to a separate manipulation policy.}
  \label{fig:real_robot_qualitative}
  \vspace{-4mm}
\end{figure}

We presented vision-language conditioned camera motion generation as a first-class embodied capability: given the current view and an intent, a robot should predict where to move its camera to acquire more useful visual evidence.
To study this problem, we introduced a pipeline that mines intent-conditioned camera-motion supervision from passive egocentric video via hindsight labeling, trains a VLM-based model with observation-gain language supervision and a continuous flow-matching pose head, and evaluates the resulting policy on a dedicated benchmark and downstream embodied tasks.
The results suggest that ordinary human video can provide effective supervision for intent-aware robot camera motion, enabling models to generalize across target approaching, exploration, and perspective shift, while transferring to real robot observations. More broadly, free-form intent-conditioned camera motion can serve as a reusable active-perception primitive: the same LIME interface supports viewpoint generation for manipulation, embodied question answering, mid-distance navigation, and object scanning.

\section{Limitations and Future Work}
\label{sec:limitations}
Despite its strong performance on our benchmark and downstream tasks, LIME has limitations. It predicts camera motion from a single image and keeps no memory of earlier views. It performs well in our budgeted multi-step evaluation, but longer-horizon tasks would need an observation memory. The target pose is not embodiment-aware and may fall outside the robot camera workspace. Future work can guide samples with workspace, kinematic, collision, and task-cost constraints.

% Our model currently predicts camera motion from a single current image and language intent.
% The closed-loop evaluation shows that repeated inference can support multi-step information gathering, but the policy does not yet maintain an explicit long-term memory of previously observed views, explored regions, or unresolved intent state.
% Extending the model with memory over observation history is an important direction for longer-horizon active perception and task execution.

% The predicted target pose is also not explicitly conditioned on the kinematic reachability of a particular robot platform or onboard camera mount.
% As a result, a visually useful target view may not always be reachable by the deployed embodiment.
% Future versions could guide or constrain flow-matching samples with embodiment-specific conditions, such as camera workspace, base or arm kinematics, collision constraints, and task-dependent motion costs.

\clearpage
% The acknowledgments are automatically included only in the final and preprint versions of the paper.
\acknowledgments{
This work was partially supported by DSO National Laboratories under Grant DSOCO24035. This work was also supported by the Swiss AI Initiative through a grant from the Swiss National Supercomputing Centre (CSCS) under project ID ab036 on the Alps infrastructure; the Swiss National Science Foundation under Advanced Grant 216260 (``Beyond Frozen Worlds: Capturing Functional 3D Digital Twins from the Real World''); and the European Union's Horizon Europe research and innovation programme under Grant Agreements No. 101214398 (ELLIOT) and No. 101298421 (GRAIL). Sunghwan Hong was supported by an ETH AI Center Postdoctoral Fellowship and an ETH Zurich Career Seed Award. Hermann Blum is supported by the Lamarr Institute for Machine Learning and Artificial Intelligence, and the Robotics Institute Germany.
}

%===============================================================================

% CoRL camera-ready order: main text, acknowledgments, references, then the appendix.
\bibliography{camera_ready/camera_ready}  % .bib

@inproceedings{yamauchi1997frontier,
  title={A frontier-based approach for autonomous exploration},
  author={Yamauchi, Brian},
  booktitle={Proceedings 1997 IEEE International Symposium on Computational Intelligence in Robotics and Automation CIRA'97.'Towards New Computational Principles for Robotics and Automation'},
  pages={146--151},
  year={1997},
  organization={IEEE}
}

@article{ahmed2023active,
  title={Active slam: A review on last decade},
  author={Ahmed, Muhammad Farhan and Masood, Khayyam and Fremont, Vincent and Fantoni, Isabelle},
  journal={Sensors},
  volume={23},
  number={19},
  pages={8097},
  year={2023},
  publisher={MDPI}
}

@article{li2026motion,
  title={Motion-Uncertainty-Aware Next-Best-View Planning for Moving Object Reconstruction},
  author={Li, Karen and Mantovani, Mattia and Wood, Robert J and Sabattini, Lorenzo and Gil, Stephanie},
  journal={arXiv preprint arXiv:2605.17593},
  year={2026}
}

@article{lluvia2021active,
  title={Active mapping and robot exploration: A survey},
  author={Lluvia, Iker and Lazkano, Elena and Ansuategi, Ander},
  journal={Sensors},
  volume={21},
  number={7},
  pages={2445},
  year={2021},
  publisher={MDPI}
}

@article{placed2023survey,
  title={A survey on active simultaneous localization and mapping: State of the art and new frontiers},
  author={Placed, Julio A and Strader, Jared and Carrillo, Henry and Atanasov, Nikolay and Indelman, Vadim and Carlone, Luca and Castellanos, Jos{\'e} A},
  journal={IEEE Transactions on Robotics},
  volume={39},
  number={3},
  pages={1686--1705},
  year={2023},
  publisher={IEEE}
}

@article{bajcsy2018revisiting,
  title={Revisiting active perception},
  author={Bajcsy, Ruzena and Aloimonos, Yiannis and Tsotsos, John K},
  journal={Autonomous Robots},
  volume={42},
  number={2},
  pages={177--196},
  year={2018},
  publisher={Springer}
}

@book{siegwart2011introduction,
  title={Introduction to autonomous mobile robots},
  author={Siegwart, Roland and Nourbakhsh, Illah Reza and Scaramuzza, Davide},
  year={2011},
  publisher={MIT press}
}

@article{chaplot2021seal,
  title={Seal: Self-supervised embodied active learning using exploration and 3d consistency},
  author={Chaplot, Devendra Singh and Dalal, Murtaza and Gupta, Saurabh and Malik, Jitendra and Salakhutdinov, Russ R},
  journal={Advances in neural information processing systems},
  volume={34},
  pages={13086--13098},
  year={2021}
}

@article{yu2023frontier,
  title={Frontier semantic exploration for visual target navigation},
  author={Yu, Bangguo and Kasaei, Hamidreza and Cao, Ming},
  journal={arXiv preprint arXiv:2304.05506},
  year={2023}
}

@article{schmid2020efficient,
  title={An efficient sampling-based method for online informative path planning in unknown environments},
  author={Schmid, Lukas and Pantic, Michael and Khanna, Raghav and Ott, Lionel and Siegwart, Roland and Nieto, Juan},
  journal={IEEE Robotics and Automation Letters},
  volume={5},
  number={2},
  pages={1500--1507},
  year={2020},
  publisher={IEEE}
}

@article{boysun2025frontiernet,
    author={Sun, Boyang and Chen, Hanzhi and Leutenegger, Stefan and Cadena, Cesar and Pollefeys, Marc and Blum, Hermann},
    journal={IEEE Robotics and Automation Letters}, 
    title={FrontierNet: Learning Visual Cues to Explore}, 
    year={2025},
    volume={10},
    number={7},
    pages={6576-6583},
    doi={10.1109/LRA.2025.3569122}
}

@InProceedings{li2025actloc,
  title = 	 {ActLoc: Learning to Localize on the Move via Active Viewpoint Selection},
  author =       {Li, Jiajie and Sun, Boyang and Giammarino, Luca Di and Blum, Hermann and Pollefeys, Marc},
  booktitle = 	 {Proceedings of The 9th Conference on Robot Learning},
  pages = 	 {1225--1245},
  year = 	 {2025},
  editor = 	 {Lim, Joseph and Song, Shuran and Park, Hae-Won},
  volume = 	 {305},
  series = 	 {Proceedings of Machine Learning Research},
  month = 	 {27--30 Sep},
  publisher =    {PMLR},
  url = 	 {https://proceedings.mlr.press/v305/li25b.html},
}

@inproceedings{zhang2020fisher,
  title={Beyond point clouds: Fisher information field for active visual localization},
  author={Zhang, Zichao and Scaramuzza, Davide},
  booktitle={IEEE International Conference on Robotics and Automation (ICRA)},
  pages={5986--5992},
  year={2019},
  organization={IEEE}
}

@article{yan2022mui,
  title={MUI-TARE: Multi-Agent Cooperative Exploration with Unknown Initial Position},
  author={Yan, Jingtian and Lin, Xingqiao and Ren, Zhongqiang and Zhao, Shiqi and Yu, Jieqiong and Cao, Chao and Yin, Peng and Zhang, Ji and Scherer, Sebastian},
  journal={arXiv preprint arXiv:2209.10775},
  year={2022}
}

@article{Kompis2021InformedScenes,
    title = {{Informed Sampling Exploration Path Planner for 3D Reconstruction of Large Scenes}},
    year = {2021},
    journal = {IEEE Robotics and Automation Letters},
    author = {Kompis, Yves and Bartolomei, Luca and Mascaro, Ruben and Teixeira, Lucas and Chli, Margarita},
    number = {4},
    month = {10},
    pages = {7894--7901},
    volume = {6},
    publisher = {Institute of Electrical and Electronics Engineers Inc.},
    doi = {10.1109/LRA.2021.3101856},
    issn = {23773766}
}

@article{padilla2026openfrontier,
  title={OpenFrontier: General Navigation with Visual-Language Grounded Frontiers},
  author={Padilla-Cerdio, Esteban and Sun, Boyang and Pollefeys, Marc and Blum, Hermann},
  journal={arXiv preprint arXiv:2603.05377},
  year={2026}
}

@article{batra2020objectnav,
  title={Objectnav revisited: On evaluation of embodied agents navigating to objects},
  author={Batra, Dhruv and Gokaslan, Aaron and Kembhavi, Aniruddha and Maksymets, Oleksandr and Mottaghi, Roozbeh and Savva, Manolis and Toshev, Alexander and Wijmans, Erik},
  journal={arXiv preprint arXiv:2006.13171},
  year={2020}
}

@article{krantz2020vlnce,
  title={Beyond the Nav-Graph: Vision-and-Language Navigation in Continuous Environments},
  author={Krantz, Jacob and Wijmans, Erik and Majumdar, Arjun and Batra, Dhruv and Lee, Stefan},
  journal={arXiv preprint arXiv:2004.02857},
  year={2020}
}

@article{khanna2024goatbench,
  title={GOAT-Bench: A Benchmark for Multi-Modal Lifelong Navigation},
  author={Khanna, Mukul and Ramrakhya, Ram and Chhablani, Gunjan and Yenamandra, Sriram and Gervet, Theophile and Chang, Matthew and Kira, Zsolt and Chaplot, Devendra Singh and Batra, Dhruv and Mottaghi, Roozbeh},
  journal={arXiv preprint arXiv:2404.06609},
  year={2024}
}

@article{yokoyama2024hm3dovon,
  title={{HM3D-OVON}: A Dataset and Benchmark for Open-Vocabulary Object Goal Navigation},
  author={Yokoyama, Naoki and Ramrakhya, Ram and Das, Abhishek and Batra, Dhruv and Ha, Sehoon},
  journal={arXiv preprint arXiv:2409.14296},
  year={2024}
}

@inproceedings{chang2023goat,
  title={{GOAT}: {GO} to Any Thing},
  author={Chang, Matthew and Gervet, Th{\'e}ophile and Khanna, Mukul and Yenamandra, Sriram and Shah, Dhruv and Min, So Yeon and Shah, Kavit and Paxton, Chris and Gupta, Saurabh and Batra, Dhruv and others},
  booktitle={Robotics: Science and Systems},
  year={2024}
}

@inproceedings{zhang20233d,
  title={3d-aware object goal navigation via simultaneous exploration and identification},
  author={Zhang, Jiazhao and Dai, Liu and Meng, Fanpeng and Fan, Qingnan and Chen, Xuelin and Xu, Kai and Wang, He},
  booktitle={Proceedings of the IEEE/CVF Conference on Computer Vision and Pattern Recognition},
  pages={6672--6682},
  year={2023}
}

@misc{beliefmapnav,
Author = {Zibo Zhou and Yue Hu and Lingkai Zhang and Zonglin Li and Siheng Chen},
Title = {BeliefMapNav: 3D Voxel-Based Belief Map for Zero-Shot Object Navigation},
Year = {2025},
Eprint = {arXiv:2506.06487},
}

@inproceedings{xie2025naviformer,
  title={{NaviFormer}: A Spatio-Temporal Context-Aware Transformer for Object Navigation},
  author={Xie, Wei and Jiang, Haobo and Zhu, Yun and Qian, Jianjun and Xie, Jin},
  booktitle={Proceedings of the AAAI Conference on Artificial Intelligence},
  volume={39},
  number={14},
  pages={14708--14716},
  year={2025}
}

@article{zeng2025janusvln,
  title={Janusvln: Decoupling semantics and spatiality with dual implicit memory for vision-language navigation},
  author={Zeng, Shuang and Qi, Dekang and Chang, Xinyuan and Xiong, Feng and Xie, Shichao and Wu, Xiaolong and Liang, Shiyi and Xu, Mu and Wei, Xing and Guo, Ning},
  journal={arXiv preprint arXiv:2509.22548},
  year={2025}
}

@article{wei2025streamvln,
  title={Streamvln: Streaming vision-and-language navigation via slowfast context modeling},
  author={Wei, Meng and Wan, Chenyang and Yu, Xiqian and Wang, Tai and Yang, Yuqiang and Mao, Xiaohan and Zhu, Chenming and Cai, Wenzhe and Wang, Hanqing and Chen, Yilun and others},
  journal={arXiv preprint arXiv:2507.05240},
  year={2025}
}

@inproceedings{wang2026vggtomega,
  title     = {{VGGT-$\Omega$}},
  author    = {Jianyuan Wang and Minghao Chen and Shangzhan Zhang and Nikita Karaev and Johannes Sch{\"o}nberger and Patrick Labatut and Piotr Bojanowski and David Novotny and Andrea Vedaldi and Christian Rupprecht},
  booktitle = {Proceedings of the IEEE/CVF Conference on Computer Vision and Pattern Recognition (CVPR)},
  year      = {2026}
}

@article{long2024instructnav,
  title={Instructnav: Zero-shot system for generic instruction navigation in unexplored environment},
  author={Long, Yuxing and Cai, Wenzhe and Wang, Hongcheng and Zhan, Guanqi and Dong, Hao},
  journal={arXiv preprint arXiv:2406.04882},
  year={2024}
}

@article{goetting2024end,
  title={End-to-end navigation with vision language models: Transforming spatial reasoning into question-answering},
  author={Goetting, Dylan and Singh, Himanshu Gaurav and Loquercio, Antonio},
  journal={arXiv preprint arXiv:2411.05755},
  year={2024}
}

@article{habibpour2025history,
  title={History-augmented vision-language models for frontier-based zero-shot object navigation},
  author={Habibpour, Mobin and Afghah, Fatemeh},
  journal={arXiv preprint arXiv:2506.16623},
  year={2025}
}

@article{yin2025unigoal, 
      title={UniGoal: Towards Universal Zero-shot Goal-oriented Navigation}, 
      author={Hang Yin and Xiuwei Xu and Linqing Zhao and Ziwei Wang and Jie Zhou and Jiwen Lu},
      journal={arXiv preprint arXiv:2503.10630},
      year={2025}
}

@misc{wang2025moge2accuratemonoculargeometry,
      title={MoGe-2: Accurate Monocular Geometry with Metric Scale and Sharp Details}, 
      author={Ruicheng Wang and Sicheng Xu and Yue Dong and Yu Deng and Jianfeng Xiang and Zelong Lv and Guangzhong Sun and Xin Tong and Jiaolong Yang},
      year={2025},
      eprint={2507.02546},
      archivePrefix={arXiv},
      primaryClass={cs.CV},
      url={https://arxiv.org/abs/2507.02546}, 
}

@misc{koo2025toward,
      title={Toward Ambulatory Vision: Learning Visually-Grounded Active View Selection}, 
      author={Juil Koo and Daehyeon Choi and Sangwoo Youn and Phillip Y. Lee and Minhyuk Sung},
      year={2025},
      eprint={2512.13250},
      archivePrefix={arXiv},
      primaryClass={cs.CV},
      url={https://arxiv.org/abs/2512.13250}, 
}

@misc{das2017embodiedquestionanswering,
      title={Embodied Question Answering}, 
      author={Abhishek Das and Samyak Datta and Georgia Gkioxari and Stefan Lee and Devi Parikh and Dhruv Batra},
      year={2017},
      eprint={1711.11543},
      archivePrefix={arXiv},
      primaryClass={cs.CV},
      url={https://arxiv.org/abs/1711.11543}, 
}

@inproceedings{OpenEQA2023,
    title         = {OpenEQA: Embodied Question Answering in the Era of Foundation Models}, 
    booktitle     = {Conference on Computer Vision and Pattern Recognition (CVPR)},
    author        = {Majumdar, Arjun and Ajay, Anurag and Zhang, Xiaohan and Putta, Pranav and Yenamandra, Sriram and Henaff, Mikael and Silwal, Sneha and Mcvay, Paul and Maksymets, Oleksandr and Arnaud, Sergio and Yadav, Karmesh and Li, Qiyang and Newman, Ben and Sharma, Mohit and Berges, Vincent and Zhang, Shiqi and Agrawal, Pulkit and Bisk, Yonatan and Batra, Dhruv and Kalakrishnan, Mrinal and Meier, Franziska and Paxton, Chris and Sax, Sasha and Rajeswaran, Aravind},
    year          = {2024},
}

@inproceedings{ren2024explore,
  title={Explore until Confident: Efficient Exploration for Embodied Question Answering},
  author={Ren, Allen Z. and Clark, Jaden and Dixit, Anushri and Itkina, Masha and Majumdar, Anirudha and Sadigh, Dorsa},
  booktitle={Robotics: Science and Systems},
  year={2024}
}

@inproceedings{EXPRESSBench,
title={Beyond the Destination: A Novel Benchmark for Exploration-Aware Embodied Question Answering},
author={Jiang, Kaixuan and Liu, Yang and Chen, Weixing and Luo, Jingzhou and Chen, Ziliang and Pan, Ling and Li, Guanbin and Lin, Liang},
year={2025},
booktitle={IEEE/CVF International Conference on Computer Vision (ICCV)}
}

@misc{sakamoto2026e3vs,
      title={E3VS-Bench: A Benchmark for Viewpoint-Dependent Active Perception in 3D Gaussian Splatting Scenes}, 
      author={Koya Sakamoto and Taiki Miyanishi and Daichi Azuma and Shuhei Kurita and Shu Morikuni and Naoya Chiba and Motoaki Kawanabe and Yusuke Iwasawa and Yutaka Matsuo},
      year={2026},
      eprint={2604.17969},
      archivePrefix={arXiv},
      primaryClass={cs.CV},
      url={https://arxiv.org/abs/2604.17969}, 
}

@INPROCEEDINGS{zhou2018continuity,
  author={Zhou, Yi and Barnes, Connelly and Lu, Jingwan and Yang, Jimei and Li, Hao},
  booktitle={2019 IEEE/CVF Conference on Computer Vision and Pattern Recognition (CVPR)}, 
  title={On the Continuity of Rotation Representations in Neural Networks}, 
  year={2019},
  volume={},
  number={},
  pages={5738-5746},
  doi={10.1109/CVPR.2019.00589}}

@article{Qwen3-VL,
      title={Qwen3-VL Technical Report}, 
      author={Shuai Bai and Yuxuan Cai and Ruizhe Chen and Keqin Chen and Xionghui Chen and Zesen Cheng and Lianghao Deng and Wei Ding and Chang Gao and Chunjiang Ge and Wenbin Ge and Zhifang Guo and Qidong Huang and Jie Huang and Fei Huang and Binyuan Hui and Shutong Jiang and Zhaohai Li and Mingsheng Li and Mei Li and Kaixin Li and Zicheng Lin and Junyang Lin and Xuejing Liu and Jiawei Liu and Chenglong Liu and Yang Liu and Dayiheng Liu and Shixuan Liu and Dunjie Lu and Ruilin Luo and Chenxu Lv and Rui Men and Lingchen Meng and Xuancheng Ren and Xingzhang Ren and Sibo Song and Yuchong Sun and Jun Tang and Jianhong Tu and Jianqiang Wan and Peng Wang and Pengfei Wang and Qiuyue Wang and Yuxuan Wang and Tianbao Xie and Yiheng Xu and Haiyang Xu and Jin Xu and Zhibo Yang and Mingkun Yang and Jianxin Yang and An Yang and Bowen Yu and Fei Zhang and Hang Zhang and Xi Zhang and Bo Zheng and Humen Zhong and Jingren Zhou and Fan Zhou and Jing Zhou and Yuanzhi Zhu and Ke Zhu},
	  journal={arXiv preprint arXiv:2511.21631},
      year={2025}
}

@article{li2025back,
  title={Back to basics: Let denoising generative models denoise},
  author={Li, Tianhong and He, Kaiming},
  journal={arXiv preprint arXiv:2511.13720},
  year={2025}
}

@inproceedings{grauman2024ego,
  title={Ego-exo4d: Understanding skilled human activity from first-and third-person perspectives},
  author={Grauman, Kristen and Westbury, Andrew and Torresani, Lorenzo and Kitani, Kris and Malik, Jitendra and Afouras, Triantafyllos and Ashutosh, Kumar and Baiyya, Vijay and Bansal, Siddhant and Boote, Bikram and others},
  booktitle={Proceedings of the IEEE/CVF Conference on Computer Vision and Pattern Recognition},
  pages={19383--19400},
  year={2024}
}

@article{zheng2026egoscale,
  title={Egoscale: Scaling dexterous manipulation with diverse egocentric human data},
  author={Zheng, Ruijie and Niu, Dantong and Xie, Yuqi and Wang, Jing and Xu, Mengda and Jiang, Yunfan and Casta{\~n}eda, Fernando and Hu, Fengyuan and Tan, You Liang and Fu, Letian and others},
  journal={arXiv preprint arXiv:2602.16710},
  year={2026}
}

@inproceedings{han2025roomtour3d,
  title={Roomtour3d: Geometry-aware video-instruction tuning for embodied navigation},
  author={Han, Mingfei and Ma, Liang and Zhumakhanova, Kamila and Radionova, Ekaterina and Zhang, Jingyi and Chang, Xiaojun and Liang, Xiaodan and Laptev, Ivan},
  booktitle={Proceedings of the IEEE/CVF Conference on Computer Vision and Pattern Recognition},
  pages={27586--27596},
  year={2025}
}

@article{kawaharazuka2025vision,
  title={Vision-language-action models for robotics: A review towards real-world applications},
  author={Kawaharazuka, Kento and Oh, Jihoon and Yamada, Jun and Posner, Ingmar and Zhu, Yuke},
  journal={IEEE Access},
  year={2025},
  publisher={IEEE}
}

@article{zhang2024vision,
  title={Vision-and-language navigation today and tomorrow: A survey in the era of foundation models},
  author={Zhang, Yue and Ma, Ziqiao and Li, Jialu and Qiao, Yanyuan and Wang, Zun and Chai, Joyce and Wu, Qi and Bansal, Mohit and Kordjamshidi, Parisa},
  journal={arXiv preprint arXiv:2407.07035},
  year={2024}
}

@inproceedings{gu2022vision,
  title={Vision-and-language navigation: A survey of tasks, methods, and future directions},
  author={Gu, Jing and Stefani, Eliana and Wu, Qi and Thomason, Jesse and Wang, Xin},
  booktitle={Proceedings of the 60th Annual Meeting of the Association for Computational Linguistics (Volume 1: Long Papers)},
  pages={7606--7623},
  year={2022}
}

@inproceedings{yeshwanth2023scannet++,
  title={Scannet++: A high-fidelity dataset of 3d indoor scenes},
  author={Yeshwanth, Chandan and Liu, Yueh-Cheng and Nie{\ss}ner, Matthias and Dai, Angela},
  booktitle={Proceedings of the IEEE/CVF International Conference on Computer Vision},
  pages={12--22},
  year={2023}
}

@inproceedings{chen2025vidbot,
    author    = {Chen, Hanzhi and Sun, Boyang and Zhang, Anran and Pollefeys, Marc and Leutenegger, Stefan},
    title     = {{VidBot}: Learning Generalizable {3D} Actions from In-the-Wild {2D} Human Videos for Zero-Shot Robotic Manipulation},
    booktitle = {Proceedings of the Computer Vision and Pattern Recognition Conference (CVPR)},
    year      = {2025},
}

@article{depthanything3,
  title={Depth Anything 3: Recovering the visual space from any views},
  author={Haotong Lin and Sili Chen and Jun Hao Liew and Donny Y. Chen and Zhenyu Li and Guang Shi and Jiashi Feng and Bingyi Kang},
  journal={arXiv preprint arXiv:2511.10647},
  year={2025}
}

@inproceedings{kareer2025egomimic,
  title={Egomimic: Scaling imitation learning via egocentric video},
  author={Kareer, Simar and Patel, Dhruv and Punamiya, Ryan and Mathur, Pranay and Cheng, Shuo and Wang, Chen and Hoffman, Judy and Xu, Danfei},
  booktitle={2025 IEEE International Conference on Robotics and Automation (ICRA)},
  pages={13226--13233},
  year={2025},
  organization={IEEE}
}

@inproceedings{ma2024nymeria,
  title={Nymeria: A massive collection of multimodal egocentric daily motion in the wild},
  author={Ma, Lingni and Ye, Yuting and Hong, Fangzhou and Guzov, Vladimir and Jiang, Yifeng and Postyeni, Rowan and Pesqueira, Luis and Gamino, Alexander and Baiyya, Vijay and Kim, Hyo Jin and others},
  booktitle={European Conference on Computer Vision},
  pages={445--465},
  year={2024},
  organization={Springer}
}

@article{damen2020epic,
  title={The epic-kitchens dataset: Collection, challenges and baselines},
  author={Damen, Dima and Doughty, Hazel and Farinella, Giovanni Maria and Fidler, Sanja and Furnari, Antonino and Kazakos, Evangelos and Moltisanti, Davide and Munro, Jonathan and Perrett, Toby and Price, Will and others},
  journal={IEEE Transactions on Pattern Analysis and Machine Intelligence},
  volume={43},
  number={11},
  pages={4125--4141},
  year={2020},
  publisher={IEEE}
}

@misc{zhang2024uninavid,
        title={Uni-NaVid: A Video-based Vision-Language-Action Model for Unifying Embodied Navigation Tasks}, 
        author={Jiazhao Zhang and Kunyu Wang and Shaoan Wang and Minghan Li and Haoran Liu and Songlin Wei and Zhongyuan Wang and Zhizheng Zhang and He Wang},
        year={2024},
        journal = {arXiv preprint arXiv:2412.06224}
      }

@inproceedings{cheng2024navila,
title = {NaVILA: Legged Robot Vision-Language-Action Model for Navigation},
    author = {Cheng, An-Chieh and Ji, Yandong and Yang, Zhaojing and Zou, Xueyan and Kautz, Jan and Biyik, Erdem and Yin,
    Hongxu and Liu, Sifei and Wang, Xiaolong},
    booktitle = {RSS},
    year = {2025},
}

@article{chu2026abot,
  title={Abot-n0: Technical report on the vla foundation model for versatile embodied navigation},
  author={Chu, Zedong and Xie, Shichao and Wu, Xiaolong and Shen, Yanfen and Luo, Minghua and Wang, Zhengbo and Liu, Fei and Leng, Xiaoxu and Hu, Junjun and Yin, Mingyang and others},
  journal={arXiv preprint arXiv:2602.11598},
  year={2026}
}

@article{wei2025ground,
  title={Ground slow, move fast: A dual-system foundation model for generalizable vision-and-language navigation},
  author={Wei, Meng and Wan, Chenyang and Peng, Jiaqi and Yu, Xiqian and Yang, Yuqiang and Feng, Delin and Cai, Wenzhe and Zhu, Chenming and Wang, Tai and Pang, Jiangmiao and others},
  journal={arXiv preprint arXiv:2512.08186},
  year={2025}
}

@article{kim24openvla,
    title={OpenVLA: An Open-Source Vision-Language-Action Model},
    author={{Moo Jin} Kim and Karl Pertsch and Siddharth Karamcheti and Ted Xiao and Ashwin Balakrishna and Suraj Nair and Rafael Rafailov and Ethan Foster and Grace Lam and Pannag Sanketi and Quan Vuong and Thomas Kollar and Benjamin Burchfiel and Russ Tedrake and Dorsa Sadigh and Sergey Levine and Percy Liang and Chelsea Finn},
    journal = {arXiv preprint arXiv:2406.09246},
    year={2024},
}

@article{black2024pi_0,
  title={$\pi_0$: A Vision-Language-Action Flow Model for General Robot Control},
  author={Black, Kevin and Brown, Noah and Driess, Danny and Esmail, Adnan and Equi, Michael and Finn, Chelsea and Fusai, Niccolo and Groom, Lachy and Hausman, Karol and Ichter, Brian and others},
  journal={arXiv preprint arXiv:2410.24164},
  year={2024}
}

@misc{intelligence2025pi05visionlanguageactionmodelopenworld,
      title={$\pi_{0.5}$: a Vision-Language-Action Model with Open-World Generalization},
      author={{Physical Intelligence} and Kevin Black and Noah Brown and James Darpinian and Karan Dhabalia and Danny Driess and Adnan Esmail and Michael Equi and Chelsea Finn and Niccolo Fusai and Manuel Y. Galliker and Dibya Ghosh and Lachy Groom and Karol Hausman and Brian Ichter and Szymon Jakubczak and Tim Jones and Liyiming Ke and Devin LeBlanc and Sergey Levine and Adrian Li-Bell and Mohith Mothukuri and Suraj Nair and Karl Pertsch and Allen Z. Ren and Lucy Xiaoyang Shi and Laura Smith and Jost Tobias Springenberg and Kyle Stachowicz and James Tanner and Quan Vuong and Homer Walke and Anna Walling and Haohuan Wang and Lili Yu and Ury Zhilinsky},
      year={2025},
      eprint={2504.16054},
      archivePrefix={arXiv},
      primaryClass={cs.LG},
      url={https://arxiv.org/abs/2504.16054},
}

@misc{liu2023liberobenchmarkingknowledgetransfer,
      title={LIBERO: Benchmarking Knowledge Transfer for Lifelong Robot Learning},
      author={Bo Liu and Yifeng Zhu and Chongkai Gao and Yihao Feng and Qiang Liu and Yuke Zhu and Peter Stone},
      year={2023},
      eprint={2306.03310},
      archivePrefix={arXiv},
      primaryClass={cs.AI},
      url={https://arxiv.org/abs/2306.03310},
}

@inproceedings{zitkovich2023rt,
  title={Rt-2: Vision-language-action models transfer web knowledge to robotic control},
  author={Zitkovich, Brianna and Yu, Tianhe and Xu, Sichun and Xu, Peng and Xiao, Ted and Xia, Fei and Wu, Jialin and Wohlhart, Paul and Welker, Stefan and Wahid, Ayzaan and others},
  booktitle={Conference on Robot Learning},
  pages={2165--2183},
  year={2023},
  organization={PMLR}
}

@inproceedings{xiong2025vision,
  title={Vision in Action: Learning Active Perception from Human Demonstrations},
  author={Xiong, Haoyu and Xu, Xiaomeng and Wu, Jimmy and Hou, Yifan and Bohg, Jeannette and Song, Shuran},
  booktitle={Conference on Robot Learning},
  pages={5450--5463},
  year={2025},
  organization={PMLR}
}

@inproceedings{kerrj2025eyerobot,
  title={Eye, Robot: Learning to Look to Act with a {BC-RL} Perception-Action Loop},
  author={Justin Kerr and Kush Hari and Ethan Weber and Chung Min Kim and Brent Yi and Tyler Bonnen and Ken Goldberg and Angjoo Kanazawa},
  booktitle={9th Annual Conference on Robot Learning},
  year={2025},
  url={http://arxiv.org/abs/2506.10968}
}

@inproceedings{yu2026egomi,
  title={{EgoMI}: Learning Active Vision and Whole-Body Manipulation from Egocentric Human Demonstrations},
  author={Yu, Justin and Shentu, Yide and Wu, Di and Abbeel, Pieter and Goldberg, Ken and Wu, Philipp},
  booktitle={2026 IEEE International Conference on Robotics and Automation (ICRA)},
  pages={6018--6025},
  year={2026},
  organization={IEEE}
}

@article{zou2026activeglasses,
  title={ActiveGlasses: Learning Manipulation with Active Vision from Ego-centric Human Demonstration},
  author={Zou, Yanwen and Shi, Chenyang and Yu, Wenye and Xue, Han and Lv, Jun and Pan, Ye and Wen, Chuan and Lu, Cewu},
  journal={arXiv preprint arXiv:2604.08534},
  year={2026}
}

@article{wang2025observer,
  title={Observer Actor: Active Vision Imitation Learning with Sparse View Gaussian Splatting},
  author={Wang, Yilong and Qian, Cheng and Fan, Ruomeng and Johns, Edward},
  journal={arXiv preprint arXiv:2511.18140},
  year={2025}
}

@article{liu2026activevla,
  title={ActiveVLA: Injecting Active Perception into Vision-Language-Action Models for Precise 3D Robotic Manipulation},
  author={Liu, Zhenyang and Gu, Yongchong and Wang, Yikai and Xue, Xiangyang and Fu, Yanwei},
  journal={arXiv preprint arXiv:2601.08325},
  year={2026}
}

@misc{huang2026perceive,
      title={I-Perceive: A Foundation Model for Active Perception with Language Instructions}, 
      author={Yongxi Huang and Zhuohang Wang and Wenjing Tang and Cewu Lu and Panpan Cai},
      year={2026},
      eprint={2603.00600},
      archivePrefix={arXiv},
      primaryClass={cs.RO},
      url={https://arxiv.org/abs/2603.00600}, 
}

@misc{InteriorGS2025,
  title        = {InteriorGS: A 3D Gaussian Splatting Dataset of Semantically Labeled Indoor Scenes},
  author       = {SpatialVerse Research Team, Manycore Tech Inc.},
  year         = {2025},
  howpublished = {\url{https://huggingface.co/datasets/spatialverse/InteriorGS}}
}

@InProceedings{VLN-BERT,
    author    = {Hong, Yicong and Wu, Qi and Qi, Yuankai and Rodriguez-Opazo, Cristian and Gould, Stephen},
    title     = {A Recurrent Vision-and-Language BERT for Navigation},
    booktitle = {Proceedings of the IEEE/CVF Conference on Computer Vision and Pattern Recognition (CVPR)},
    month     = {June},
    year      = {2021},
    pages     = {1643-1653}
}

@misc{gemini31pro,
    title        = {{Gemini 3.1 Pro} Model Card},
    author       = {{Google DeepMind}},
    year         = {2026},
    howpublished = {\url{https://deepmind.google/models/model-cards/gemini-3-1-pro/}}
}

@misc{openai2026gpt56,
    title        = {{GPT-5.6} System Card},
    author       = {{OpenAI}},
    year         = {2026},
    howpublished = {\url{https://deploymentsafety.openai.com/gpt-5-6}}
}

@misc{openai2025gpt5,
    title        = {{GPT-5} System Card},
    author       = {{OpenAI}},
    year         = {2025},
    howpublished = {\url{https://cdn.openai.com/gpt-5-system-card.pdf}}
}

@article{comanici2025gemini,
    title   = {Gemini 2.5: Pushing the Frontier with Advanced Reasoning, Multimodality, Long Context, and Next Generation Agentic Capabilities},
    author  = {Comanici, Gheorghe and Bieber, Eric and Schaekermann, Mike and others},
    journal = {arXiv preprint arXiv:2507.06261},
    year    = {2025}
}

@article{bai2025qwen25vl,
    title   = {{Qwen2.5-VL} Technical Report},
    author  = {Bai, Shuai and Chen, Keqin and Liu, Xuejing and others},
    journal = {arXiv preprint arXiv:2502.13923},
    year    = {2025}
}

@article{wu2025vilasr,
    title   = {Reinforcing Spatial Reasoning in Vision-Language Models with Interwoven Thinking and Visual Drawing},
    author  = {Wu, Junfei and Guan, Jian and Feng, Kaituo and Liu, Qiang and Wu, Shu and Wang, Liang and Wu, Wei and Tan, Tieniu},
    journal = {arXiv preprint arXiv:2506.09965},
    year    = {2025}
}

@inproceedings{ma2025spatialreasoner,
    title     = {{SpatialReasoner}: Towards Explicit and Generalizable {3D} Spatial Reasoning},
    author    = {Ma, Wufei and Chou, Yu-Cheng and Liu, Qihao and Wang, Xingrui and de Melo, Celso and Xie, Jianwen and Yuille, Alan},
    booktitle = {Advances in Neural Information Processing Systems (NeurIPS)},
    year      = {2025}
}

\clearpage
\appendix
\raggedbottom % appendix pages keep natural spacing instead of being stretched to full height

\begin{center}
{\Large\bfseries Supplementary Material for}\\[0.4em]
{\large\bfseries ``LIME: Learning Intent-aware Camera Motion from Egocentric Video''}
\end{center}
\vspace{0.5em}

\section{Technical Details}
\label{sec:supplementary}
\subsection{Dataset Curation}
\label{sec:supp-dataset-curation}

The RoomTour3D and Nymeria labeling process uses dataset-specific forks of the same structured pair-labeling prompt. Both forks share the output schema below; the Nymeria fork additionally filters egocentric hands, body, and held-object content, to prevent the labeling result from concentrating on them. 

\begin{tcolorbox}[
  title={Qwen3-VL Dataset Labeling Prompt},
  label={box:egocentric-pair-labeling-prompt},
  colback=gray!3,
  colframe=gray!60,
  boxrule=0.4pt,
  arc=1mm,
  left=1mm,
  right=1mm,
  top=1mm,
  bottom=1mm,
  breakable
]
\small
\textbf{System prompt.}

You are an expert annotator producing structured observation-gain labels for pairs of frames from an egocentric trajectory. Your output is the supervision signal for a vision-language model that learns to reason about what new information a camera move reveals.

You receive, in order:
\begin{enumerate}
    \item Current frame: the agent's view before the move.
    \item Goal frame: the agent's view after the move.
    \item Pose summary: the camera motion from the current frame to the goal frame, expressed in OpenCV camera convention. Trust this metric ground truth over your visual guess of the motion.
\end{enumerate}

Produce a single JSON object describing what new information the goal frame gives the agent that was not available in the current frame, plus a closed-vocabulary tag for the type of camera motion that produced the gain. Use the pose summary as the primary signal for choosing \texttt{motion\_type}; the visual content tells what was revealed, and the pose tells how.

Set \texttt{info\_gain\_present} to false when the move is static, disconnected, or when no new static-scene content is visible in the goal frame. When false, leave the sub-fields, captions, and intentions empty, null, or \texttt{[]}. When true, fill the following structured fields:
\begin{itemize}
    \item \texttt{newly\_visible}: items not visible in the current frame but clearly visible in the goal frame.
    \item \texttt{enhanced\_views}: items already visible in the current frame that are now seen better.
    \item \texttt{spatial\_anchor}: one sentence describing how the goal frame is positioned relative to the current frame in space.
\end{itemize}

\texttt{caption\_long} is a purely visual observation-gain description. Describe the visual information gained going from the current frame to the goal frame, grounded in \texttt{newly\_visible} and \texttt{enhanced\_views}. Keep it about what is seen, not how the camera moved, and lead directly with content. \texttt{caption\_short} is a short intent-style query and must be a faithful compression of \texttt{caption\_long}.

Produce a diverse set of short imperative intentions a user might say to ask the agent to make this move. Each intention is grounded in one of the structured fields: \texttt{newly\_visible}, \texttt{enhanced\_views}, \texttt{spatial\_anchor}, or \texttt{motion\_type}. The allowed intention kinds are \texttt{find}, \texttt{explore}, \texttt{inspect}, and \texttt{navigate}.

Output strict JSON with these fields:
\begin{itemize}
    \item \texttt{motion\_type}: one of the closed-set motion types.
    \item \texttt{info\_gain\_present}: boolean.
    \item \texttt{reason}: string or null.
    \item \texttt{newly\_visible}: list of entries with \texttt{item}, \texttt{kind}, and \texttt{where\_relative\_to\_current}.
    \item \texttt{enhanced\_views}: list of entries with \texttt{item} and \texttt{change}.
    \item \texttt{spatial\_anchor}: string or null.
    \item \texttt{caption\_long}: string or null.
    \item \texttt{caption\_short}: string or null.
    \item \texttt{intentions}: list of entries with \texttt{text}, \texttt{kind}, and \texttt{anchor}, where \texttt{kind} is \texttt{find}, \texttt{explore}, \texttt{inspect}, or \texttt{navigate}.
    \item \texttt{quality\_self\_score}: integer from 1 to 5.
\end{itemize}

\vspace{0.5em}
\textbf{User prompt.}

\texttt{[current frame -- view BEFORE the move]}\\
\texttt{[goal frame -- view AFTER the move]}\\
\texttt{[POSE SUMMARY]}\\
\texttt{<POSE\_SUMMARY>}

Produce the JSON object now.
\end{tcolorbox}

\begin{figure}[H]
    \centering
    \includegraphics[width=0.384\linewidth]{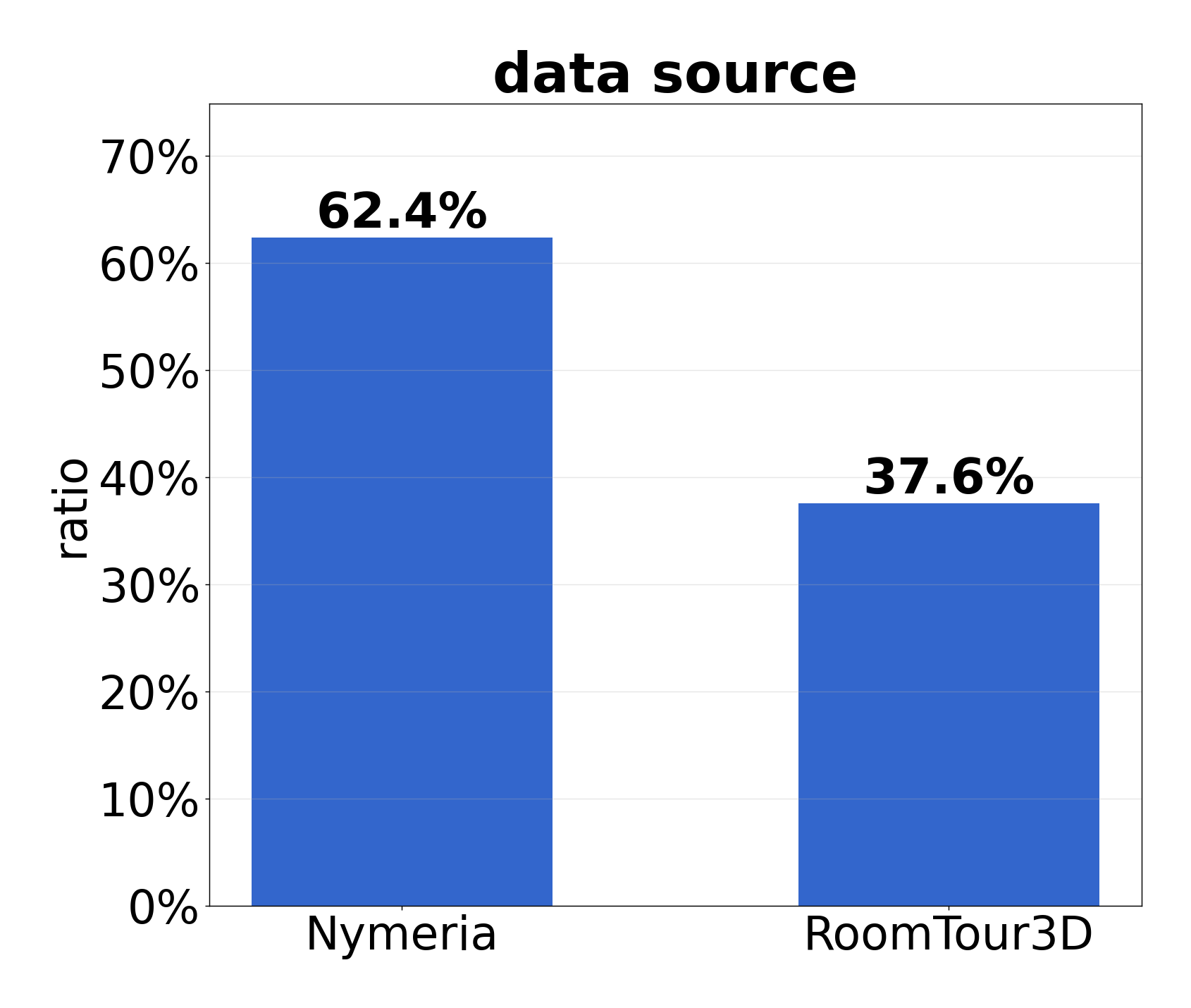}
    \hfill
    \includegraphics[width=0.48\linewidth]{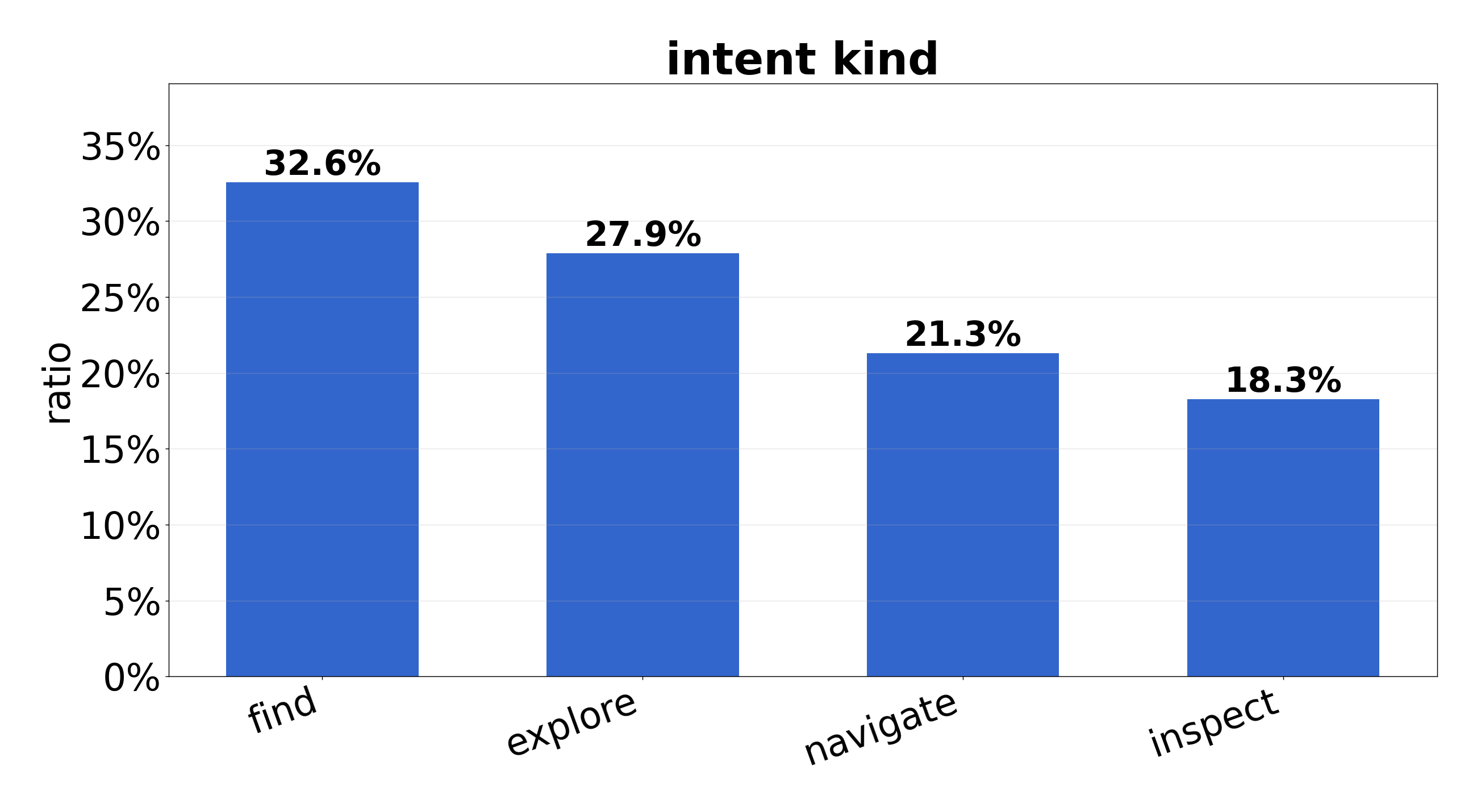}

    \vspace{0.35em}
    \includegraphics[width=0.48\linewidth]{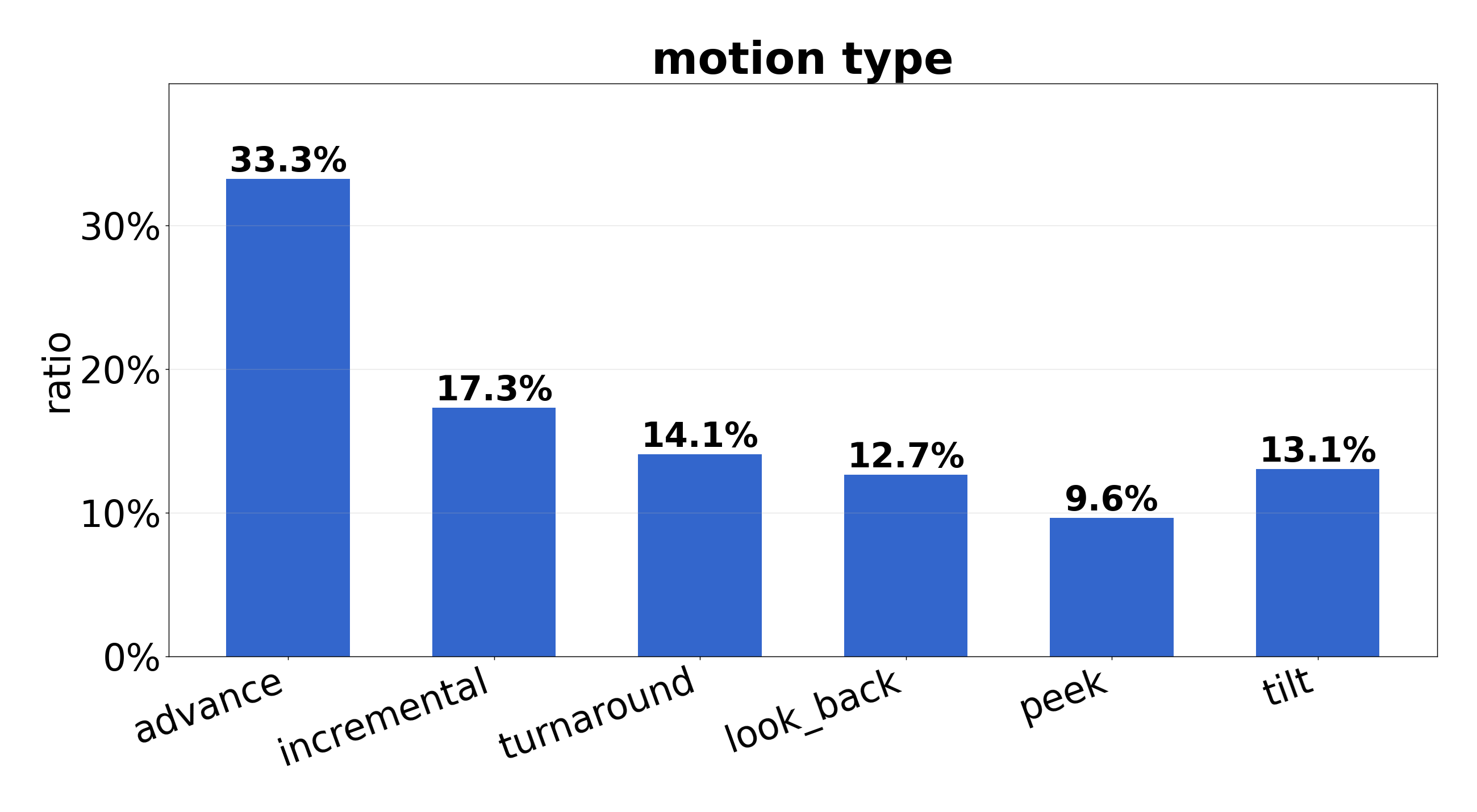}
    \hfill
    \includegraphics[width=0.48\linewidth]{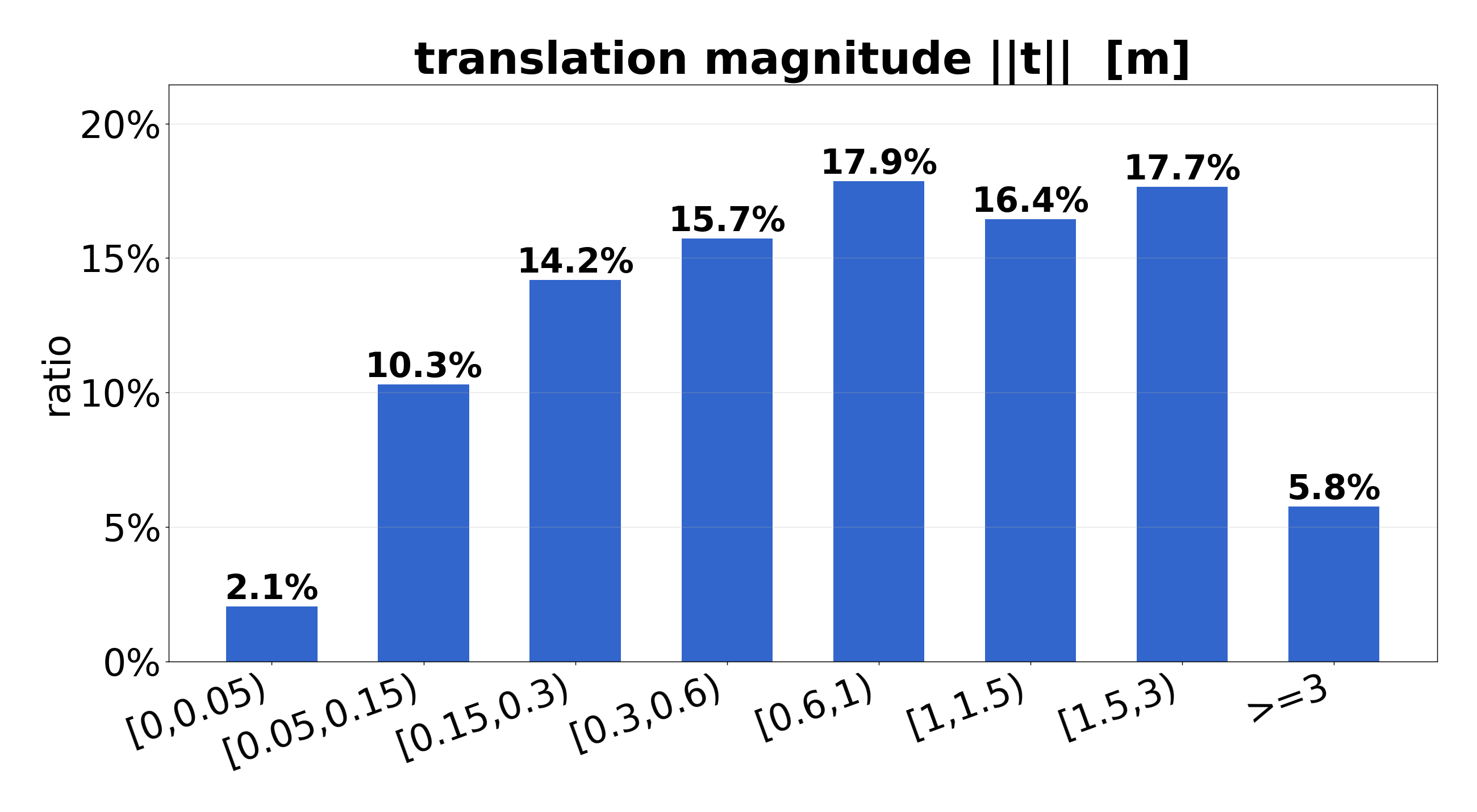}

    \vspace{0.35em}
    \includegraphics[width=0.62\linewidth]{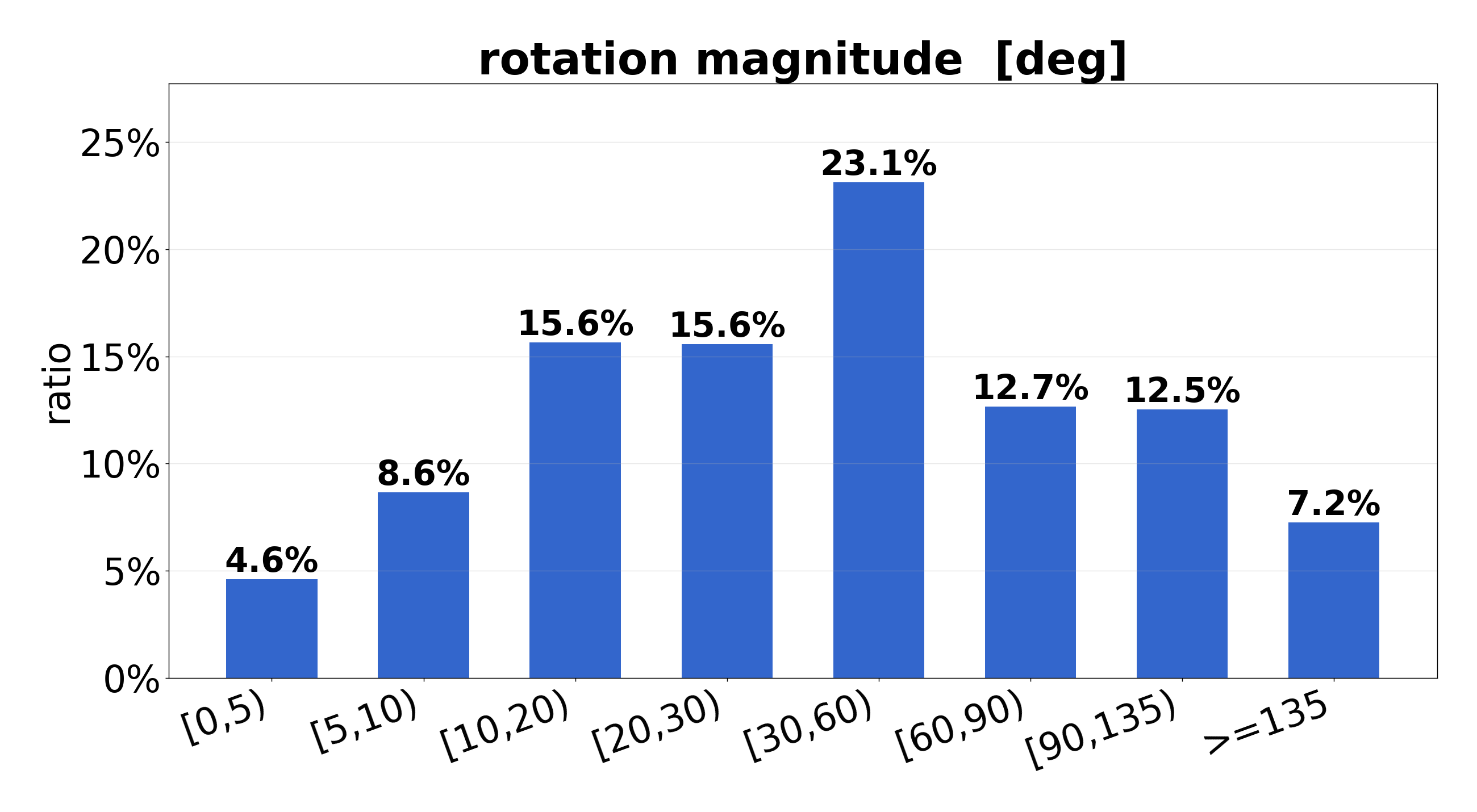}

    \caption{Dataset distributions after balanced subsampling. We balance the start-goal image pairs across data source, intent kind, motion type, translation magnitude, and rotation magnitude before expanding them into intent-conditioned training examples.}
    \label{fig:supp_dataset_balance}
\end{figure}

\begin{figure}[p]
    \centering
    \includegraphics[width=\linewidth,height=0.78\textheight,keepaspectratio]{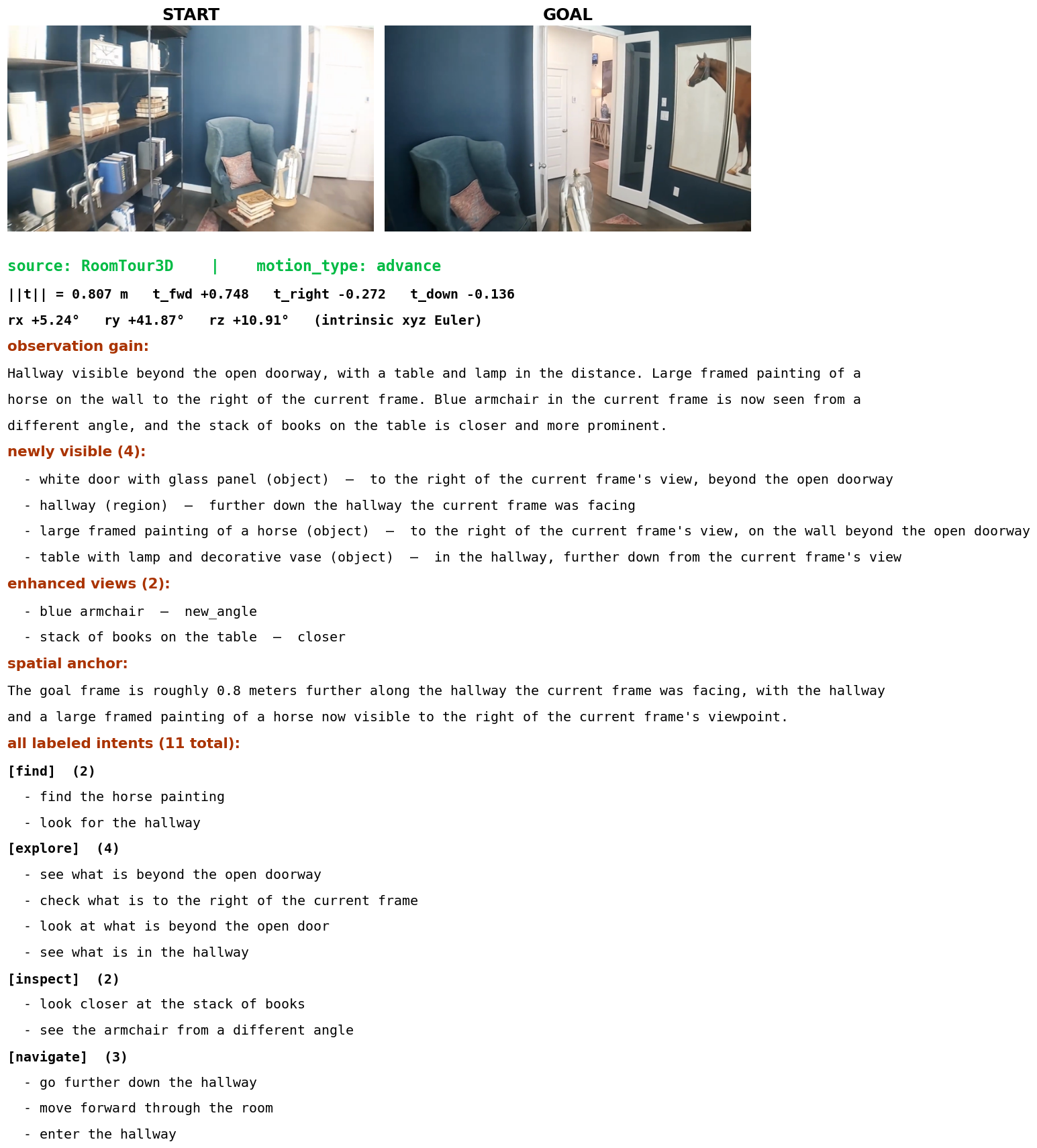}

    \caption{Example of a start--goal pair label from RoomTour3D, showing the paired frames, motion metadata, observation-gain description, structured visual-change fields, and generated intent set.}
    \label{fig:supp_dataset_label_example_rm3d}
\end{figure}

\begin{figure}[p]
    \centering
    \includegraphics[width=\linewidth,height=0.78\textheight,keepaspectratio]{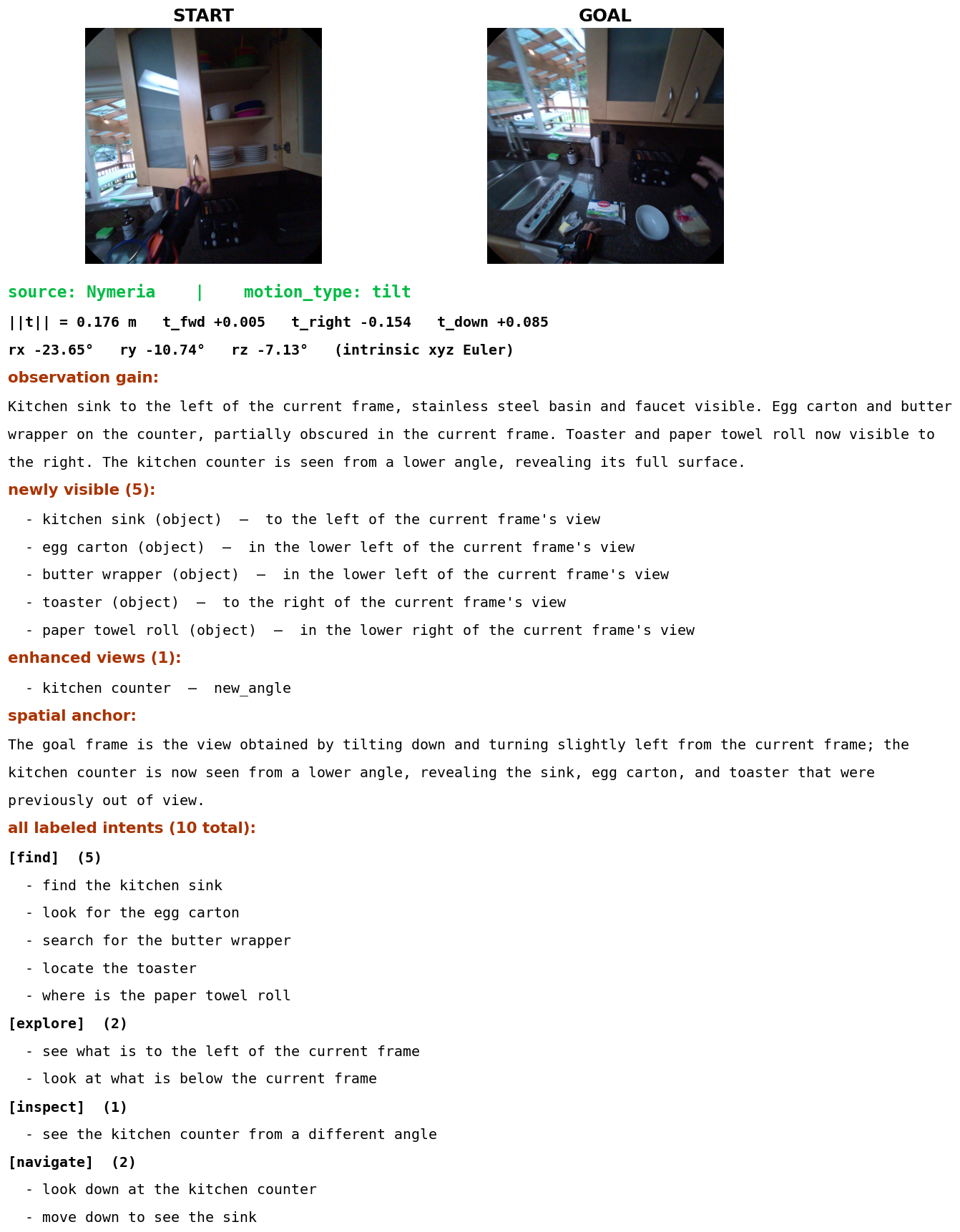}

    \caption{Example of a start--goal pair label from Nymeria, showing the paired frames, motion metadata, observation-gain description, structured visual-change fields, and generated intent set.}
    \label{fig:supp_dataset_label_example_nymeria}
\end{figure}

Starting from roughly $2$M candidate start--goal pairs extracted from RoomTour3D and Nymeria, we apply balanced subsampling before expanding pairs into intent-conditioned examples.
We subsample across data source, intent kind, motion type, translation magnitude, and rotation magnitude.
Figure~\ref{fig:supp_dataset_balance} summarizes the resulting distributions, showing that the final training pool retains coverage over semantic and geometric axes rather than collapsing to short forward motions or a single intent family. Figures~\ref{fig:supp_dataset_label_example_rm3d} and~\ref{fig:supp_dataset_label_example_nymeria} show representative valid image pairs with generated labels. During training, the dataloader samples one intent from each available intent category for a retained pair.

\subsection{Training Setup}
\label{sec:supp-training}

Table \ref{tab:supp_training_hyperparams} lists the key configurations and hyperparameters for the optimizer, input, flow-head, and inference used in training our main model.

\begin{table}[H]
  \centering
  \begin{tabular}{ll}
    \toprule
    \textbf{Component} & \textbf{Setting} \\
    \midrule
    Optimizer & AdamW, learning rate $1\mathrm{e}{-5}$ \\
    Schedule & Cosine decay, warmup ratio $0.03$ \\
    Regularization & Weight decay $0.1$, max gradient norm $1.0$ \\
    Precision & bf16 with DeepSpeed ZeRO-3 \\
    Sequence length & Maximum length $8192$ tokens \\
    Image resolution budget & $200704$ max pixels, $784$ min pixels \\
    Loss weight & $\lambda_{\mathrm{pose}}=1.0$ \\
    Pose target & 3D translation + first two rotation columns \\
    Flow head & 512 hidden dimension, 6 cross-attention blocks, 8 heads \\
    Time embedding & 256-D sinusoidal embedding followed by an MLP \\
    Flow parameterization & x-prediction with zero-initialized output projection \\
    Training augmentation & Hidden-state noise $0.01$; no intent-token masking \\
    Inference & 128 max gain tokens, 10 Euler steps, 5 pose samples \\
    \bottomrule
  \end{tabular}
  \caption{Training and inference hyperparameters for the main LIME model.}
  \label{tab:supp_training_hyperparams}
\end{table}

For the real-world robot experiments, we further adapt the LIME checkpoint with a lightweight LoRA fine-tuning stage on a small real-world dataset collected with Aria Gen~1 glasses.
The set contains around $1{,}700$ start--goal pairs, aligned to the robot camera setting and balanced across find, explore, and navigate-style intents.
We initialize from the main LIME checkpoint, continue fine-tuning the flow-matching head, and train LoRA adapters on the VLM backbone with rank $64$, alpha $128$, and dropout $0.0$.
This adaptation is used only for the real-world robot experiments.
The robot-adaptation run uses learning rate $5\mathrm{e}{-5}$, batch size $4$, gradient accumulation $1$, and $3$ epochs on $4$ GPUs.

\section{Benchmark Design and Evaluation Details}

\subsection{Benchmark Construction}
\label{sec:supp-benchmark-construction}

Sec.~\ref{sec:bench_design} describes the high-level construction of our benchmark. Here we provide additional details, including the intent-family distribution, start--reference camera-motion statistics, and the camera intrinsics and height ranges used for rendering.

\subsubsection{Intent-Family Distribution and Test Set Statistics}
\label{sec:supp-intent-family-stats}

The benchmark is built from $105$ InteriorGS scenes and $259$ curated start--reference pairs. Since a single start--reference pair may support multiple language intents, the final benchmark contains $425$ instruction-level examples. These examples are distributed across the three intent families: $152$ Target-approaching, $142$ Exploration, and $131$ Perspective-shift.

We summarize the start--reference motion distribution across the $425$ instruction-level examples in Table~\ref{tab:supp_benchmark_motion_stats}. Translation is measured as the Euclidean distance between the start and reference camera centers, and rotation is measured as the geodesic angle between the start and reference camera orientations. These statistics characterize the spatial scale of the annotated reference motions and show that most remain within a local viewpoint-change range, consistent with the benchmark's focus on local intent-conditioned camera motion.

\begin{table}[H]
  \centering
  \resizebox{0.85\textwidth}{!}{
  \begin{tabular}{lcccc}
    \toprule
    \textbf{Intent Family}
    & \textbf{\#Pairs}
    & \textbf{\#Examples}
    & \textbf{Trans. med./p90 (m)}
    & \textbf{Rot. med./p90 ($^\circ$)} \\
    \midrule
    Target-approaching & 90 & 152 & 4.10 / 5.03 & 46.4 / 74.3 \\
    Exploration & 78 & 142 & 3.17 / 4.41 & 88.5 / 143.5 \\
    Perspective-shift & 91 & 131 & 1.92 / 3.78 & 43.5 / 140.0 \\
    \midrule
    Overall & 259 & 425 & 3.18 / 4.74 & 56.2 / 123.5 \\
    \bottomrule
  \end{tabular}
  }
  \caption{Benchmark dataset and motion statistics. Examples are instruction-level samples derived from the start--reference pairs. Translation and rotation are computed between the annotated start and reference poses and aggregated over instruction-level examples.}
  \label{tab:supp_benchmark_motion_stats}
\end{table}

\subsubsection{Camera Agent Configuration}
\label{sec:supp-camera-intrinsics}

All benchmark images are rendered at $640 \times 360$ resolution. For each curated start--reference pair, we use a pinhole camera model with square pixels, centered principal point, and focal length $f_x=f_y$ uniformly sampled from $[260,350]$ pixels. This corresponds to a vertical field-of-view range of approximately $54.5^\circ$ to $69.4^\circ$. The sampled intrinsics are held fixed for the start image, reference image, and evaluation-trajectory frames of each benchmark example, and are shared across all evaluated methods.

\begin{table}[H]
  \centering
  \begin{tabular}{lc}
    \toprule
    \textbf{Quantity} & \textbf{Value / Range} \\
    \midrule
    Image resolution & $640 \times 360$ \\
    Principal point & $(320, 180)$ px \\
    Focal length & $f_x=f_y \sim \mathcal{U}(260,350)$ px \\
    Vertical field of view & $54.5^\circ$--$69.4^\circ$ \\
    Start-view height, 10--90 percentile & $1.40$--$1.81$ m \\
    Reference-view height, 10--90 percentile & $1.41$--$1.86$ m \\
    \bottomrule
  \end{tabular}
  \caption{Camera intrinsics, image resolution, and camera-height statistics for the benchmark. Intrinsics are sampled per curated start--reference entry and kept fixed throughout evaluation.}
  \label{tab:supp_camera_intrinsics_heights}
\end{table}

\subsubsection{Benchmark Examples}
\label{sec:supp-benchmark-examples}

\begin{figure}[!t]
  \centering
  \captionsetup[subfigure]{justification=centering}

  \begin{subfigure}{0.82\textwidth}
    \centering
    \includegraphics[width=0.48\linewidth]{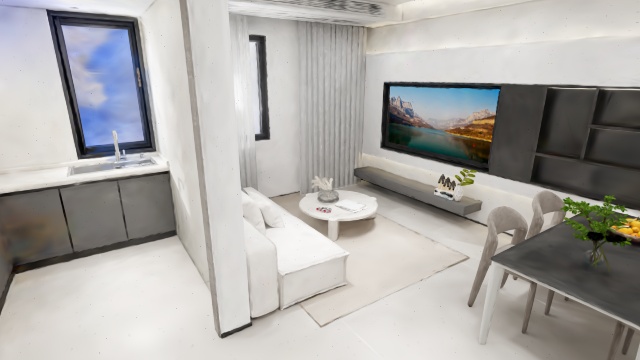}
    \hspace{0.02\linewidth}
    \includegraphics[width=0.48\linewidth]{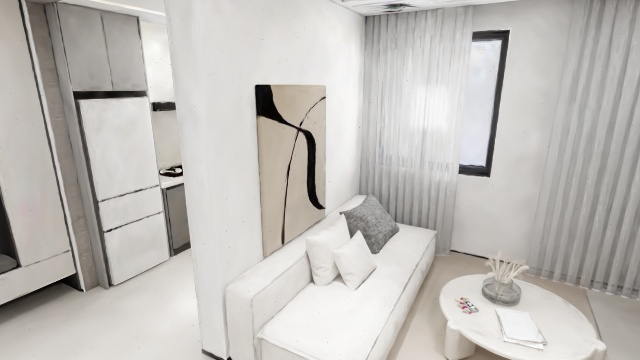}
    \caption{\textbf{Exploration:} Go to see the painting on the wall in the living room.}
  \end{subfigure}

  \vspace{0.75em}

  \begin{subfigure}{0.82\textwidth}
    \centering
    \includegraphics[width=0.48\linewidth]{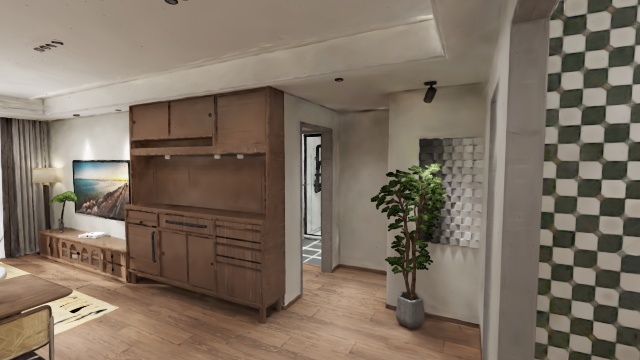}
    \hspace{0.02\linewidth}
    \includegraphics[width=0.48\linewidth]{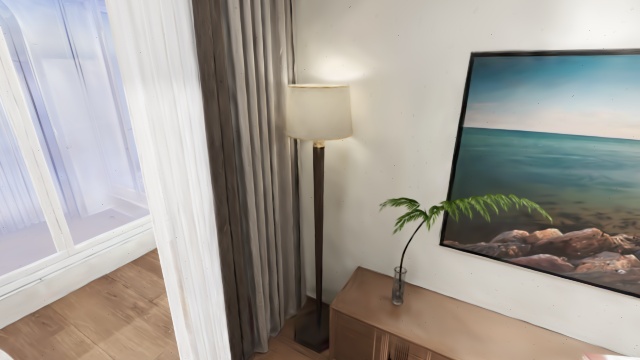}
    \caption{\textbf{Target-approaching:} Go to the floor lamp next to the TV.}
  \end{subfigure}

  \vspace{0.75em}

  \begin{subfigure}{0.82\textwidth}
    \centering
    \includegraphics[width=0.48\linewidth]{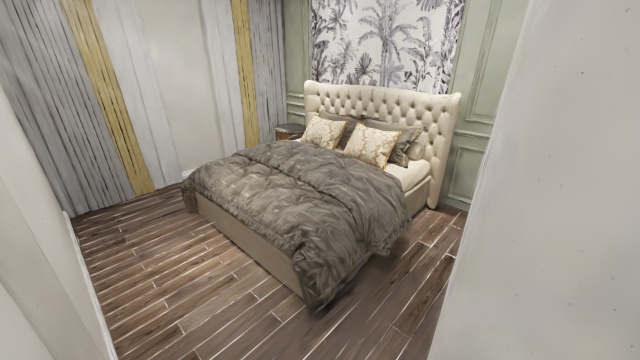}
    \hspace{0.02\linewidth}
    \includegraphics[width=0.48\linewidth]{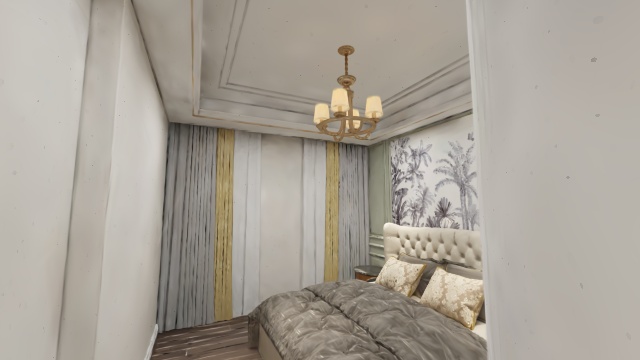}
    \caption{\textbf{Perspective-shift:} Look at the ceiling area above the bed in front of you.}
  \end{subfigure}

  \caption{Representative benchmark examples from the three intent families. In each row, the left image is the start view and the right image is the held-out goal reference view.}
  \label{fig:supp_benchmark_examples}
\end{figure}

Figure~\ref{fig:supp_benchmark_examples} shows representative benchmark examples from the three intent families. Each example consists of a start image, a language intent, and a held-out reference view that illustrates one possible intent-satisfying camera pose.

\subsection{Evaluation Protocol}

All methods are evaluated through the same budgeted multi-step protocol. Each method receives the current rendered observation and the language intent in its adapted input format, predicts a camera motion, action, or stop signal, and the resulting view is rendered in InteriorGS. Success is evaluated over the generated trajectory prefix under a shared motion budget, rather than by direct pose error to the annotated reference view. The held-out reference image $I_g$ is used only by the evaluator as visual evidence of one intent-satisfying view; it is not shown to the method.

The following subsections define the normalized motion budget, VLM-as-a-judge criteria for Exploration and Perspective-shift, two-stage Target-approaching evaluation, collision-aware evaluation, human validation of Gemini judgments, robustness to the choice of VLM judge, baseline adaptation details, and evaluation hardware.

\subsubsection{Sequential-Query Evaluation Trajectory and Step Budget}

The benchmark supports multi-step execution by repeatedly applying each method to the latest rendered observation. Each evaluation trajectory starts from the benchmark start view, denoted frame $0$. At frame $k$, the model receives the current rendered RGB image and the language intent, then predicts either a relative camera motion, a discrete/parameterized action, or a stop signal depending on the method. The predicted motion is applied to the current camera pose, and the next observation is rendered from the resulting pose using the sample's camera intrinsics. Thus, for $k>0$, frame $k$ is the view obtained after applying the $k$-th model prediction.

Because different methods use different action parameterizations, we measure evaluation-trajectory length using a normalized motion budget rather than a fixed number of model calls. For a transition from pose $p_{k-1}$ to pose $p_k$, let $\Delta t_k$ be the Euclidean distance between the two camera centers in meters, and let $\Delta r_k$ be the geodesic rotation angle between the two camera orientations in degrees. We define the transition cost as
\[
  c_k = \max\left(\frac{\Delta t_k}{0.1}, \frac{\Delta r_k}{10}\right).
\]
The cumulative cost at frame $k$ is
\[
  C_k = \sum_{i=1}^{k} c_i.
\]
Under our default success metric, a frame is eligible to count as successful only if $C_k \leq 60$. This corresponds to a budget of up to $6$ m of pure translation or $600^\circ$ of pure rotation, while also constraining mixed translation--rotation trajectories. The budget is intentionally local: it allows multi-step correction and limited exploration around the start view, while preventing the evaluation from becoming long-horizon navigation or allowing success through unconstrained random walk.

If a method emits a stop action, the evaluation trajectory terminates at the current frame. Methods without an explicit stop action are evaluated until no further frame can be produced within the shared budget. If a predicted transition would exceed the budget, we record the over-budget prediction for debugging but do not treat the resulting frame as a valid success candidate. Therefore, success is determined only over frames whose cumulative cost is within the budget.

We report example-level success rate (SR), computed over benchmark examples rather than generated frames. A benchmark example is counted as successful if at least one eligible frame in the evaluation trajectory satisfies the category-specific success criterion described below. Otherwise, it is counted as a failure. We do not evaluate by direct pose error to the annotated reference pose, because many different viewpoints can satisfy the same language intent. The held-out reference view is used only as evaluation evidence for what an intent-satisfying observation can look like, not as a unique target pose that the model must reproduce.

\subsubsection{VLM-as-a-Judge Success Criteria for Exploration and Perspective-Shift}
\label{sec:supp-judge-criteria}

For examples in the Exploration and Perspective-shift intent families, success cannot be reliably measured by distance to the annotated reference pose. The same intent may be satisfied by multiple nearby or even substantially different viewpoints, as long as the resulting image reveals the requested visual evidence. We therefore evaluate these categories using a VLM-as-a-judge protocol, following recent viewpoint-dependent active perception evaluations such as E3VS-Bench~\cite{sakamoto2026e3vs}.

For each eligible candidate frame in an evaluation trajectory, the judge is given the language intent, the start image $I_s$, the held-out reference image $I_g$, and the candidate image rendered from the model's predicted pose. The reference image is used as evidence for one valid way to satisfy the intent, but it is not treated as a pixel-level target or as the only acceptable view. A candidate frame is judged successful if it provides sufficient visual evidence to satisfy the instruction, even when its viewpoint, scale, or composition differs from $I_g$.

For Exploration, success requires the candidate view to reveal the requested object, region, or visual evidence that is absent or insufficiently recognizable from the start view; moving in a plausible exploratory direction is not sufficient unless the requested evidence becomes visible. For Perspective-shift, success requires the candidate view to improve observation of the specified object, region, or spatial relation, for example by revealing occluded content, changing the viewing side, inspecting above/below/around an object, or adjusting distance to obtain a more informative view. In both cases, the candidate frame need not match the held-out reference view exactly, but it must provide enough visual evidence to satisfy the intent.

We use Gemini-3.1-Pro-Preview~\cite{gemini31pro} as the automatic judge and ask it for a binary success decision for each candidate frame. Candidate frames are evaluated under the shared step budget described above, and the first eligible frame judged successful is recorded as the first successful frame. If no eligible frame is judged successful, the evaluation trajectory is counted as a failure for that benchmark example. The exact judge prompt for these two intent families is provided below.

\begin{tcolorbox}[
  title={Exploration and Perspective-shift Gemini judge prompt},
  label={box:exploration-perspective-judge-prompt},
  colback=gray!3,
  colframe=gray!60,
  boxrule=0.4pt,
  arc=1mm,
  left=1mm,
  right=1mm,
  top=1mm,
  bottom=1mm,
  breakable
]
\small
\textbf{System prompt.}

You are judging one candidate view from a robot/agent navigation trajectory in a 3D indoor scene.
You will be shown exactly three RGB images in this order:
\begin{enumerate}
    \item START context: the stored starting view that the human labeller wrote the instruction against. Use it only as context to interpret the instruction.
    \item END reference frame: a human-labelled reference view for the intended end state.
    \item CANDIDATE frame: the agent trajectory view being judged.
\end{enumerate}

The instruction in the user message is the source of truth. Decide whether the CANDIDATE frame satisfies that instruction. Candidate frame 0 is the agent's initial trajectory view; depending on the evaluation mode it may or may not look identical to the START context. An agent can be successful at its initial view, so frame 0 is a valid candidate. The END reference frame is only a visual reference for the intended goal state; it is not the only acceptable answer and not a pixel-matching goal. The candidate does not need to match the exact viewpoint, distance, crop, or angle of the END reference. It can succeed from a different view if it clearly satisfies the instruction.

Use the START frame only as context for the initial situation. Do not mark the candidate successful merely because it is different from START. Do not evaluate the physical plausibility or quality of intermediate path frames; this judge only decides whether this one candidate view satisfies the instruction, with the END reference as non-exclusive visual context.

Output strict JSON with two fields:
\begin{itemize}
    \item \texttt{success}: boolean, true iff the candidate frame clearly satisfies the instruction.
    \item \texttt{reasoning}: string, at most 60 words, giving a brief justification. Mention how the candidate relates to the instruction and, when useful, to the END reference.
\end{itemize}

\vspace{0.5em}
\textbf{User prompt.}

\texttt{Instruction: "<INSTRUCTION>"}\\
\texttt{Candidate frame: frame <CANDIDATE\_FRAME\_INDEX> of <N\_FRAMES\_TOTAL> total frames.}

Images are provided in order: START context frame, END reference frame, then CANDIDATE frame. The candidate may be frame 0, meaning the agent's initial trajectory view, which may or may not look identical to the START context. The END reference is a helpful example of the intended goal state, but it is not a pixel-matching goal. The candidate does not need to match the exact viewpoint, distance, crop, or angle of the END reference. Judge whether the candidate satisfies the instruction.

Respond with strict JSON: \texttt{\{"success": <bool>, "reasoning": <str>\}}.
\end{tcolorbox}

\subsubsection{Geometric and Visual Success Criteria for Target-Approaching}
\label{sec:supp-target-criteria}

Target-approaching examples ask the model to move toward a specified target object, object group, or fixture. Unlike Exploration and Perspective-shift examples, these examples include an explicit spatial requirement: the model should not merely obtain any view in which the target object is visible, but should move close enough to the intended target object for the view to support inspection. We therefore use a two-stage evaluation protocol that combines geometric proximity with visual verification.

In the first stage, we check whether each candidate camera pose is sufficiently close to the annotated target. Each Target-approaching example is associated with a target name and a 3D target bounding box in the InteriorGS scene. For a candidate frame, we compute the distance from the camera center to the target box surface. The frame passes stage 1 if this distance is below an adaptive threshold determined by the physical size of the target object. Let $s$ denote the maximum side length of the target bounding box. We assign a threshold of $0.8$ m to targets at the 10th percentile of $s$, and a threshold of $1.2$ m to targets at the 90th percentile of $s$. For targets with intermediate sizes, the distance threshold is linearly interpolated between these two values; targets outside the percentile range use the corresponding clipped endpoint threshold. We visually inspected representative targets near these two percentile anchors to ensure that the thresholds correspond to physically meaningful close-up distances for both small tabletop objects and larger furniture-scale targets.

The geometric stage is necessary because Target-approaching success depends on spatial proximity, not only visual presence. A candidate view may contain the target while still being far away, especially for large or salient objects, and a VLM-as-a-judge evaluator is not reliable at estimating metric distance from a single rendered image. Conversely, geometric proximity alone is also insufficient: a camera can be close to the target box while the target is occluded, outside the field of view, or visually ambiguous. The second stage therefore verifies visual recognizability.

In the second stage, frames that pass the geometric proximity test are evaluated by a Gemini-based VLM-as-a-judge visual-verification step. The judge is given the original instruction, the target object name, a contextual target-object phrase, the END goal reference image, and the candidate image. The exact visual-verification prompt used for this stage is provided below. A frame is counted as a Target-approaching success only if it passes both stages: it must be geometrically close to the annotated target and the target object must be visually identifiable in the rendered image. The first eligible frame satisfying both conditions is recorded as the first successful frame. If no frame within the step budget satisfies both stages, the trajectory is counted as a Target-approaching failure.

\begin{figure}[H]
  \centering
  \begin{tikzpicture}
    \begin{axis}[
      width=0.92\textwidth,
      height=0.48\textwidth,
      ymin=0,
      ymax=90,
      ylabel={Target stage-1 SR (\%)},
      xlabel={Adaptive threshold pair (small / large target, m)},
      symbolic x coords={0.8/1.2,1.0/1.5,1.2/1.8,1.4/2.1,1.6/2.4},
      xtick=data,
      tick align=outside,
      axis line style={black!70},
      tick style={black!70},
      x tick label style={font=\small},
      y tick label style={font=\small},
      label style={font=\small},
      ymajorgrids=true,
      xmajorgrids=false,
      grid style={draw=gray!25},
      legend style={
        at={(0.5,1.04)},
        anchor=south,
        legend columns=5,
        font=\small,
        draw=none,
        /tikz/every even column/.append style={column sep=0.5em}
      },
      mark size=2.4pt,
      every axis plot/.append style={line width=1.15pt},
    ]

    \addplot+[color=janus, mark=o, mark options={fill=white}] coordinates {
      (0.8/1.2,5.9) (1.0/1.5,7.9) (1.2/1.8,15.1) (1.4/2.1,19.1) (1.6/2.4,29.6)
    };
    \addlegendentry{JanusVLN}

    \addplot+[color=uninavid, mark=square*, mark options={fill=white}] coordinates {
      (0.8/1.2,0.0) (1.0/1.5,0.4) (1.2/1.8,3.3) (1.4/2.1,7.2) (1.6/2.4,11.8)
    };
    \addlegendentry{Uni-NaVid}

    \addplot+[color=vgavs, mark=triangle*, mark options={fill=white}] coordinates {
      (0.8/1.2,19.7) (1.0/1.5,32.9) (1.2/1.8,53.9) (1.4/2.1,66.4) (1.6/2.4,80.3)
    };
    \addlegendentry{VG-AVS}

    \addplot+[color=vlmnav, mark=diamond*, mark options={fill=white}] coordinates {
      (0.8/1.2,0.0) (1.0/1.5,0.0) (1.2/1.8,0.7) (1.4/2.1,0.7) (1.6/2.4,0.7)
    };
    \addlegendentry{VLMnav}

    \addplot+[
      color=ours,
      mark=*,
      mark options={fill=ours},
      line width=2.0pt
    ] coordinates {
      (0.8/1.2,64.0) (1.0/1.5,71.3) (1.2/1.8,77.0) (1.4/2.1,83.1) (1.6/2.4,86.6)
    };
    \addlegendentry{\textbf{Ours}}

    \end{axis}
  \end{tikzpicture}
  \caption{Target stage-1 proximity SR under increasingly relaxed adaptive distance thresholds. Each x-axis tick denotes the small-object / large-object threshold pair in meters, assigned to the 10th and 90th percentiles of target AABB max-side size, with intermediate thresholds linearly interpolated. Curves report mean SR over three runs under the shared motion budget.}
  \label{fig:target_stage1_threshold_sweep}
\end{figure}
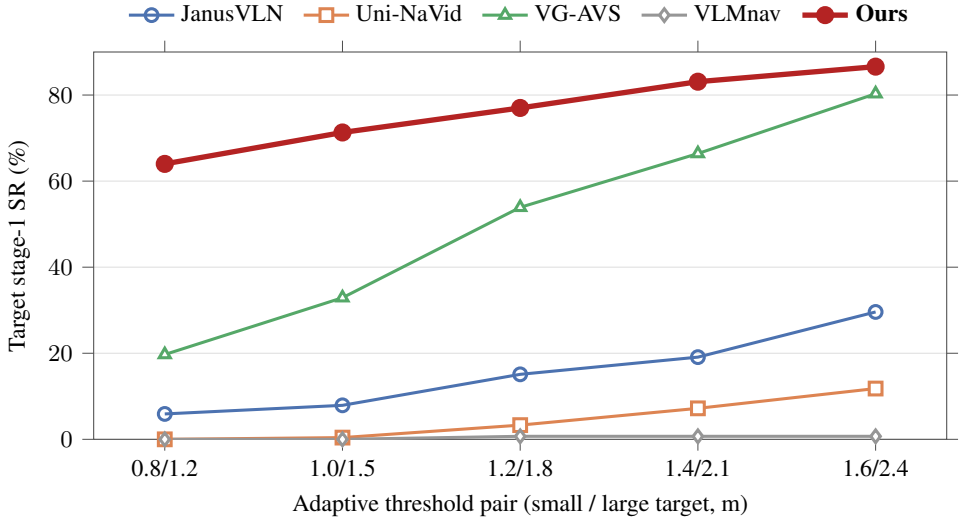

This two-stage protocol avoids two complementary failure modes. It prevents image-only false positives in which the target is visible but not actually approached, and it prevents geometry-only false positives in which the camera is near the target annotation but the rendered image does not provide recognizable visual evidence of the target. We use the default adaptive threshold pair $(0.8\text{ m}, 1.2\text{ m})$ for the results in Table~\ref{tab:main_results}; Fig.~\ref{fig:target_stage1_threshold_sweep} analyzes how target stage-1 SR changes as this threshold pair is relaxed. The main ranking between different methods is stable across thresholds.

Relaxing the threshold increases stage-1 SR for all methods, as expected, but the relative ordering remains stable across the sweep: our method achieves the highest stage-1 SR at every threshold pair, while VLMnav remains near zero and Uni-NaVid improves only modestly. VG-AVS is the most threshold-sensitive baseline, rising sharply as the allowed distance increases, which suggests that it often moves in the general direction of the target but does not approach it as closely under the default close-up criterion. This sensitivity analysis supports the default $(0.8\text{ m}, 1.2\text{ m})$ setting as a strict but physically meaningful target-approach criterion rather than an arbitrary operating point.

\begin{tcolorbox}[
  title={Stage-2 Target-approaching visual-verification judge prompt},
  label={box:target-stage2-prompt},
  colback=gray!3,
  colframe=gray!60,
  boxrule=0.4pt,
  arc=1mm,
  left=1mm,
  right=1mm,
  top=1mm,
  bottom=1mm,
  breakable
]
\small
\textbf{System prompt.}

You are judging whether one robot/agent candidate view clearly shows a specific target object in a 3D indoor scene. You will be shown exactly two RGB images in this order:
\begin{enumerate}
    \item END goal reference image: a human-labelled reference view that helps identify the intended target object and context.
    \item Candidate prediction frame: the trajectory frame being judged.
\end{enumerate}

The END goal reference is not a pixel-matching goal. Use it only as visual context for the target object's identity, appearance, and scene context. The candidate can succeed from a different viewpoint, crop, distance, or angle if it clearly shows the intended target object.

Decide whether the candidate prediction frame shows the target object clearly enough for a human to recognize or inspect it. Mild occlusion, different viewpoint, or imperfect centering can still be successful. Reject the candidate if the target is absent, too tiny, too blurry/dark, heavily occluded, or cut off so much that the target cannot be identified.

Output strict JSON with exactly these fields:
\begin{itemize}
    \item \texttt{success}: boolean
    \item \texttt{visibility}: \texttt{clear}, \texttt{mostly\_clear}, \texttt{partially\_visible}, \texttt{not\_visible}, or \texttt{uncertain}
    \item \texttt{occlusion}: \texttt{none}, \texttt{mild}, \texttt{moderate}, \texttt{heavy}, or \texttt{uncertain}
    \item \texttt{reasoning}: string, at most 60 words
\end{itemize}

\vspace{0.5em}
\textbf{User prompt.}

\texttt{Original instruction: "<INSTRUCTION>"}\\
\texttt{Target object name: "<TARGET\_OBJECT\_NAME>"}\\
\texttt{Target-object phrase with context: "<PROMPT\_TARGET\_PHRASE>"}

Images are provided in order: END goal reference image, then candidate prediction frame. The END goal reference image is a non-exclusive reference for identifying the target object and context, not a pixel-matching goal. Judge only whether the candidate prediction frame clearly shows the target object enough for recognition or inspection.

The candidate frame has already passed a geometric proximity check to the target annotation. Do not judge navigation path quality, physical plausibility, or whether the camera is close enough. Your only job is to judge whether the target is visually visible and identifiable in the candidate image.

Use the target-object phrase/context to disambiguate the intended object when there are multiple similar objects. Do not require every contextual support object to be fully visible if the target object identity is otherwise clear.

If the target-object phrase is plural or refers to a group, such as ``the books'' or ``the oranges'', success does not require every individual item to be visible. Mark success when enough of the intended group is clearly visible to recognize or inspect the target group.

Respond with strict JSON: \texttt{\{"success": <bool>, "visibility": "clear|mostly\_clear|partially\_visible|not\_visible|uncertain", "occlusion": "none|mild|moderate|heavy|uncertain", "reasoning": <str>\}}.
\end{tcolorbox}

\subsubsection{Collision-Aware Success Rate (CA-SR)}

The standard SR metric evaluates whether a trajectory eventually reaches an intent-satisfying view, but it does not penalize trajectories whose camera path passes through scene geometry before reaching that view. CA-SR uses the same example-success criterion as SR under the shared motion budget, but requires the trajectory prefix up to the first successful frame to remain collision-free with respect to the InteriorGS scene point cloud.

We model the agent as a point located at the camera center. For each scene, we use the InteriorGS 3D Gaussian point cloud as the geometric proxy for occupied scene structure and build a nearest-neighbor index over its 3D point positions. For an originally successful trajectory with first successful frame $f^\star$, we check the camera centers for frames $0,1,\ldots,f^\star$. Frame $0$ corresponds to the start camera pose, and frame $k>0$ corresponds to the camera pose after the $k$-th model prediction. Let $q_k$ be the camera center at frame $k$, and let
\[
  d_k = \min_{p \in \mathcal{P}} \|q_k - p\|_2
\]
be its nearest-neighbor distance to the scene point cloud $\mathcal{P}$. A frame is considered collision-free if $d_k \geq \tau_{\mathrm{clear}}$, where we use $\tau_{\mathrm{clear}} = 0.15$ m for the reported results. A successful trajectory remains collision-aware successful only if all frames in the prefix satisfy this clearance constraint:
\[
  \min_{0 \leq k \leq f^\star} d_k \geq \tau_{\mathrm{clear}}.
\]

Original failures remain failures under CA-SR. Original successes are converted to collision-aware failures if any checked camera center violates the clearance threshold before or at the first successful frame. The denominator of CA-SR is unchanged from SR: all benchmark examples in the corresponding intent family are counted.

This audit is intentionally conservative but lightweight. It checks only the evaluated discrete frames, not continuous line segments between consecutive poses, and it treats the camera center as a point agent rather than modeling the full camera body. We also only audit the prefix through the first successful frame, since later frames are irrelevant once the trajectory has already satisfied the benchmark example. Thus, CA-SR should be interpreted as a stricter version of SR that penalizes visually successful trajectories whose successful prefix intersects the reconstructed scene geometry.

\subsubsection{Human-Judge Validation}
\label{sec:supp-human-judge}

Because our benchmark uses automatic visual judging for semantic success, we validate the Gemini-based evaluator against human annotations on a balanced subset of trajectories. This audit is intended to measure whether the automatic judge agrees with human perception of task success, rather than to replace the full automatic evaluation. We use the same budgeted success criterion as in Table~\ref{tab:main_results}.

We sample $30$ examples from each intent family: Target-approaching, Exploration, and Perspective-shift. This gives $90$ benchmark examples in total. For each of them, we include the run-1 evaluation trajectory from each of the five methods in Table~\ref{tab:main_results}, resulting in $450$ trajectory-level human judgments. Samples are drawn from examples for which all five methods have run-1 evaluation trajectories, so every selected benchmark example can be compared across methods. Human annotators view the start image, the held-out reference image, the instruction, and the generated trajectory frames, then mark the first successful frame if any frame within the evaluation trajectory satisfies the intent; otherwise the trajectory is marked as failure. We compare these human labels against the corresponding Gemini-based SR labels. Agreement is the fraction of trajectory-level binary success/failure labels that match between the human annotator and Gemini; it does not require the first successful frame index to be identical.

\begin{table}[H]
  \centering
  \resizebox{0.9\textwidth}{!}{
  \begin{tabular}{lccccc}
    \toprule
    \textbf{Intent Family} & \textbf{$n$} & \textbf{Human SR} & \textbf{Gemini SR} & \textbf{$\Delta$ (G--H)} & \textbf{Agreement} \\
    \midrule
    Target-approaching & 150 & 11.3 & 13.3 & +2.0 & 98.0 \\
    Exploration & 150 & 24.7 & 34.0 & +9.3 & 90.7 \\
    Perspective-shift & 150 & 28.7 & 36.7 & +8.0 & 85.3 \\
    \midrule
    Overall & 450 & 21.6 & 28.0 & +6.4 & 91.3 \\
    \bottomrule
  \end{tabular}
  }
  \caption{Human validation of the Gemini-based SR evaluator on a balanced subset of benchmark trajectories under the shared motion budget. We sample $30$ examples per intent family and evaluate run-1 trajectories from five methods, giving $150$ judgments per intent family and $450$ judgments in total. SR values and agreement are reported in percent. $\Delta$ denotes Gemini SR minus human SR.}
  \label{tab:human_gemini_sr_validation}
\end{table}

The automatic judge shows strong agreement with human annotations, reaching $91.3\%$ agreement overall. Agreement is highest for Target-approaching examples, where the two-stage geometric and visual protocol makes the success criterion relatively explicit. Exploration and Perspective-shift examples have lower agreement because multiple views can partially satisfy an intent, making the boundary between partial and sufficient visual evidence less crisp. Gemini SR is consistently higher than human SR, indicating that the automatic judge is somewhat more permissive than human annotators. Nevertheless, agreement remains high across all intent families, supporting the use of Gemini-based judging for the full benchmark while retaining this human audit as a calibration check.

\subsubsection{Robustness to the Choice of VLM Judge}
\label{sec:supp-judge-robustness}

To test whether our conclusions depend on the choice of judge, we re-judge the run-1 evaluation trajectories of the five methods in Table~\ref{tab:main_results} on all $425$ benchmark examples with GPT-5.6-Sol~\cite{openai2026gpt56}, Qwen3-VL-8B, and Qwen3-VL-32B~\cite{Qwen3-VL}, keeping the prompts, the Target-approaching geometric stage, and the collision-audit procedure unchanged. As shown in Table~\ref{tab:judge_robustness}, judges differ in absolute leniency, with the open-weight judges being far more permissive, but LIME ranks first in overall SR and CA-SR under every judge.

\begin{table}[H]
  \centering
  \resizebox{\textwidth}{!}{
  \begin{tabular}{lcccccccc}
    \toprule
    \textbf{Method}
    & \multicolumn{2}{c}{\textbf{Gemini-3.1-Pro-Preview}}
    & \multicolumn{2}{c}{\textbf{GPT-5.6-Sol}}
    & \multicolumn{2}{c}{\textbf{Qwen3-VL-8B}}
    & \multicolumn{2}{c}{\textbf{Qwen3-VL-32B}} \\
    \cmidrule(lr){2-3}
    \cmidrule(lr){4-5}
    \cmidrule(lr){6-7}
    \cmidrule(lr){8-9}
    & \textbf{SR} & \textbf{CA-SR}
    & \textbf{SR} & \textbf{CA-SR}
    & \textbf{SR} & \textbf{CA-SR}
    & \textbf{SR} & \textbf{CA-SR} \\
    \midrule
    JanusVLN & $21.9$ & $19.5$ & $26.6$ & $24.0$ & $56.7$ & $52.7$ & $54.1$ & $49.6$ \\
    Uni-NaVid & $11.8$ & $11.3$ & $15.1$ & $14.4$ & $49.6$ & $47.3$ & $46.1$ & $43.1$ \\
    VG-AVS & $24.5$ & $16.5$ & $26.1$ & $17.9$ & $52.0$ & $37.9$ & $50.4$ & $37.2$ \\
    VLMnav & $4.0$ & $3.8$ & $6.1$ & $5.6$ & $29.9$ & $27.1$ & $36.0$ & $34.4$ \\
    \textbf{Ours} & $\mathbf{47.8}$ & $\mathbf{34.6}$ & $\mathbf{52.0}$ & $\mathbf{39.3}$ & $\mathbf{72.2}$ & $\mathbf{57.4}$ & $\mathbf{72.7}$ & $\mathbf{56.5}$ \\
    \bottomrule
  \end{tabular}
  }
  \caption{Overall SR and CA-SR (\%) under different VLM judges, computed on the same run-1 evaluation trajectories under the shared motion budget. Only the VLM success decisions are replaced; the Target-approaching geometric stage and the collision-audit procedure are unchanged.}
  \label{tab:judge_robustness}
\end{table}

\subsubsection{Baseline Adaptation Details}
\label{sec:supp-baselines}

We compare LIME against four representative open-source baselines: JanusVLN and Uni-NaVid for language-conditioned navigation, VG-AVS for active view selection in visual question answering, and VLMnav for zero-shot VLM-based navigation. These methods were not originally designed for relative $SE(3)$ target-pose prediction in InteriorGS, so we adapt their input and action interfaces while keeping the shared motion budget and success metrics fixed.

The baselines differ mainly in their language interface, motion output, and native observation/action convention. JanusVLN and Uni-NaVid receive navigation-style instructions and output discrete planar actions such as moving forward, turning left/right, or stopping. VG-AVS receives a question-style input and outputs a planar active-view action parameterized by heading rotation, forward distance, and final view rotation. VLMnav is a zero-shot navigation pipeline that first queries whether to stop and then selects a polar navigation action from depth-derived navigability candidates. For a fair comparison, we instantiate the VLMnav pipeline with Qwen3-VL-4B as its VLM backend, matching the scale of LIME's VLM backbone. In contrast, LIME directly predicts a relative $SE(3)$ target camera pose from the current RGB observation and language intent. JanusVLN and Uni-NaVid are naturally tied to gravity-aligned planar navigation conventions; VG-AVS follows a similar planar active-view convention in its AVS-HM3D evaluation setup; and VLMnav renders observations from a fixed downward-pitch camera viewpoint in its native setup.

For the comparison in Table~\ref{tab:main_results}, we choose an adapter that preserves the benchmark observation while respecting each baseline's native action space. Specifically, JanusVLN, Uni-NaVid, and VG-AVS observe the benchmark's original sample-start view, but their predicted planar actions are executed in a gravity-planar action frame derived from the start pose. This avoids changing the visual input seen by the method, while still applying actions in the planar convention expected by these baselines. VG-AVS is given an EQA-style question derived from the same underlying intent. VLMnav is evaluated with the sample-start view rather than its native fixed downward-pitch camera convention in Table~\ref{tab:main_results}, but its pipeline still uses rendered depth to construct navigability masks and candidate polar actions. LIME does not require an action adapter because its output is already a relative $SE(3)$ camera motion. For LIME, each evaluation step uses $10$ Euler integration steps and draws $5$ flow-matching pose samples. We execute the mean of the sampled poses as the relative camera motion for that step. This is the simplest way to reduce the samples to a single motion; more advanced sample-selection strategies can be applied instead, as in our object-scanning experiment (Appendix~\ref{sec:supp-multistep}).

We also evaluate alternative adapter choices to test whether the main conclusion depends on this interface choice. For JanusVLN, Uni-NaVid, and VG-AVS, the gravity-planar variant renders the initial observation from the gravity-planar view, which is closer to their native embodied-agent setup but changes the benchmark start image. For VLMnav, the VLMnav-pitch variant renders observations using its native fixed downward-pitch camera convention instead of the benchmark sample-start view. Table~\ref{tab:baseline_mode_comparison} reports these variants.

The adapter comparison shows that baseline performance is sensitive to the observation/action convention, but the overall conclusion is stable. Preserving the sample-start view is generally stronger for JanusVLN and VG-AVS, while Uni-NaVid benefits somewhat from the fully gravity-planar variant. VLMnav improves when using its native fixed downward-pitch camera convention, indicating that its pipeline is particularly tied to its original camera convention. However, all adapter variants remain substantially below LIME, suggesting that the main result is not an artifact of a single unfavorable baseline adapter.

We also execute LIME's predictions in the same planar action frame as the sample-start baselines; this planar variant still outperforms all baselines overall, while its gap to the full model, mostly in Target-approaching, reflects the benefit of native $SE(3)$ prediction.

\begin{table}[t]
  \centering
  \resizebox{\textwidth}{!}{
  \begin{tabular}{lcccccccc}
    \toprule
    \textbf{Method}
    & \multicolumn{2}{c}{\textbf{Target-approaching}}
    & \multicolumn{2}{c}{\textbf{Exploration}}
    & \multicolumn{2}{c}{\textbf{Perspective-shift}}
    & \multicolumn{2}{c}{\textbf{All}} \\
    \cmidrule(lr){2-3}
    \cmidrule(lr){4-5}
    \cmidrule(lr){6-7}
    \cmidrule(lr){8-9}
    & \textbf{SR} & \textbf{CA-SR}
    & \textbf{SR} & \textbf{CA-SR}
    & \textbf{SR} & \textbf{CA-SR}
    & \textbf{SR} & \textbf{CA-SR} \\
    \midrule
    JanusVLN (sample-start) & $2.0$ & $1.3$ & $33.8$ & $27.5$ & $32.1$ & $32.1$ & $21.9$ & $19.5$ \\
    JanusVLN (gravity-planar) & $0.7$ & $0.7$ & $21.8$ & $14.8$ & $29.8$ & $29.8$ & $16.7$ & $14.4$ \\
    Uni-NaVid (sample-start) & $0.0$ & $0.0$ & $9.9$ & $8.5$ & $27.5$ & $27.5$ & $11.8$ & $11.3$ \\
    Uni-NaVid (gravity-planar) & $1.3$ & $1.3$ & $12.0$ & $8.5$ & $32.1$ & $32.1$ & $14.4$ & $13.2$ \\
    VG-AVS (sample-start) & $8.6$ & $6.6$ & $30.3$ & $12.0$ & $36.6$ & $32.8$ & $24.5$ & $16.5$ \\
    VG-AVS (gravity-planar) & $8.6$ & $6.6$ & $23.2$ & $9.9$ & $32.8$ & $30.5$ & $20.9$ & $15.1$ \\
    VLMnav (sample-start) & $0.0$ & $0.0$ & $4.9$ & $4.2$ & $7.6$ & $7.6$ & $4.0$ & $3.8$ \\
    VLMnav (VLMnav-pitch) & $0.0$ & $0.0$ & $10.6$ & $8.5$ & $15.3$ & $13.7$ & $8.2$ & $7.1$ \\
    Ours (planar actions) & $14.5$ & $8.6$ & $\mathbf{53.5}$ & $30.3$ & $38.9$ & $32.8$ & $35.1$ & $23.3$ \\
    \textbf{Ours} & $\mathbf{48.0}$ & $\mathbf{34.9}$ & $50.7$ & $\mathbf{31.0}$ & $\mathbf{44.3}$ & $\mathbf{38.2}$ & $\mathbf{47.8}$ & $\mathbf{34.6}$ \\
    \bottomrule
  \end{tabular}
  }
  \caption{Run-1 SR and CA-SR for baseline model variants and our method under the shared motion budget. Sample-start variants preserve the benchmark sample's original start view for observation, while gravity-planar variants render the initial observation from a gravity-aligned planarized start pose. VLMnav-pitch renders observations using VLMnav's native fixed downward-pitch camera convention instead of the benchmark sample-start view. Ours (planar actions) keeps only the horizontal translation and yaw of LIME's predictions, with camera height and initial tilt fixed.}
  \label{tab:baseline_mode_comparison}
\end{table}

\subsubsection{Evaluation Hardware}

Evaluation trajectories were generated on NVIDIA GPUs. JanusVLN was evaluated on an NVIDIA A100 80GB GPU because its inference pipeline exceeded the memory available on 24GB GPUs. All other methods, including LIME, Uni-NaVid, VG-AVS, and VLMnav, were evaluated on NVIDIA RTX 4090 24GB GPUs. This hardware difference was only used to satisfy model memory requirements; all methods followed the same InteriorGS rendering setup, budgeted multi-step protocol, success criteria, and judging protocol described above.

\section{Additional Experiments}

\subsection{Ablations and Model Variants}

We study the main design choices of LIME under the same benchmark setting used for the results in Table~\ref{tab:main_results}, using the same motion budget, Gemini-based success metric, and collision-aware audit. Table~\ref{tab:ablation_results_with_ca_per_category} reports SR and CA-SR for each intent family and overall, in two groups. The \emph{ablations} vary a component of the default model: training without observation-gain supervision (w/o Gain), and executing the mean of one or ten flow-matching samples instead of the default five (FM samples=1, FM samples=10). The \emph{variants} extend the default model: back-propagating the pose loss into the VLM backbone instead of detaching the hidden states fed to the flow head (FM attach), augmenting the flow-matching condition with a monocular depth image~\cite{wang2025moge2accuratemonoculargeometry} (Depth-aug FM), and using a larger Qwen3-VL-8B backbone (8B backbone).

\begin{table}[H]
  \centering
  \resizebox{\textwidth}{!}{
  \begin{tabular}{lcccccccc}
    \toprule
    \textbf{Method}
    & \multicolumn{2}{c}{\textbf{Target-approaching}}
    & \multicolumn{2}{c}{\textbf{Exploration}}
    & \multicolumn{2}{c}{\textbf{Perspective-shift}}
    & \multicolumn{2}{c}{\textbf{All}} \\
    \cmidrule(lr){2-3}
    \cmidrule(lr){4-5}
    \cmidrule(lr){6-7}
    \cmidrule(lr){8-9}
    & \textbf{SR} & \textbf{CA-SR}
    & \textbf{SR} & \textbf{CA-SR}
    & \textbf{SR} & \textbf{CA-SR}
    & \textbf{SR} & \textbf{CA-SR} \\
    \midrule
    \multicolumn{9}{l}{\textit{Ablations}} \\
    w/o Gain & $9.2$ & $5.9$ & $33.1$ & $18.3$ & $29.8$ & $26.7$ & $23.5$ & $16.5$ \\
    FM samples=1 & $40.8$ & $28.9$ & $42.3$ & $24.6$ & $46.6$ & $40.5$ & $43.1$ & $31.1$ \\
    FM samples=10 & $46.7$ & $33.6$ & $53.5$ & $\mathbf{35.9}$ & $50.4$ & $40.5$ & $50.1$ & $\mathbf{36.5}$ \\
    \midrule
    \multicolumn{9}{l}{\textit{Default}} \\
    Ours main & $48.0$ & $34.9$ & $50.7$ & $31.0$ & $44.3$ & $38.2$ & $47.8$ & $34.6$ \\
    \midrule
    \multicolumn{9}{l}{\textit{Variants}} \\
    FM attach & $\mathbf{55.3}$ & $\mathbf{38.2}$ & $50.0$ & $30.3$ & $50.4$ & $40.5$ & $\mathbf{52.0}$ & $36.2$ \\
    Depth-aug FM & $39.5$ & $30.3$ & $42.3$ & $29.6$ & $\mathbf{54.2}$ & $39.7$ & $44.9$ & $32.9$ \\
    8B backbone & $44.7$ & $27.0$ & $\mathbf{54.2}$ & $33.8$ & $45.8$ & $\mathbf{41.2}$ & $48.2$ & $33.6$ \\
    \bottomrule
  \end{tabular}
  }
  \caption{Ablations and model variants on the proposed benchmark. All rows use run 1 and report SR and CA-SR in percent under the shared motion budget. Unless specified otherwise, the default number of flow-matching samples is 5.}
  \label{tab:ablation_results_with_ca_per_category}
\end{table}

The ablations confirm the role of observation-gain supervision and show that averaging more flow-matching samples helps, with overall SR rising from $43.1$ with one sample to $50.1$ with ten. The variants indicate headroom: back-propagating the pose loss into the backbone raises overall SR to $52.0$ at about $30\%$ longer training, while the larger 8B backbone changes results unevenly across intent families. We keep the lighter configuration as the default. All rows are single-run results, so small differences should be read as preliminary.

\subsection{Sequential-Query Inference Efficiency}

We evaluate whether LIME reaches successful views with fewer sequential model decisions. Because LIME is trained from start--goal image pairs, it can predict a local 3D target pose that makes larger progress in a single step rather than relying on short primitive actions. Inference steps are counted up to the first successful frame under the shared success criterion, where frame $0$ is the start image and frame $k$ corresponds to the $k$-th decision step. Runtime is averaged over all recorded trajectory steps, and also over successful trajectories up to the first successful frame. For VLMnav, one decision step includes both a stopping query and an action-selection query. Table~\ref{tab:inference_efficiency} reports each method's overall SR, the mean number of inference steps to success, and runtime per inference. Since inference-step averages are conditioned on successful trajectories, the SR column indicates how broad each method's success set is. LIME requires fewer inference steps than the VLN-style baselines while achieving substantially higher SR; VG-AVS uses fewer steps, but each inference is substantially slower.

\begin{table}[H]
  \centering
  \resizebox{\textwidth}{!}{
  \begin{tabular}{lccccccc}
    \toprule
    \textbf{Method}
    & \textbf{SR (All)}
    & \makecell{\textbf{\#Inf.}\\\textbf{Target-approaching}}
    & \makecell{\textbf{\#Inf.}\\\textbf{Exploration}}
    & \makecell{\textbf{\#Inf.}\\\textbf{Perspective-shift}}
    & \makecell{\textbf{\#Inf.}\\\textbf{All}}
    & \textbf{Time/Inf. (s)}
    & \textbf{Succ. Time/Inf. (s)} \\
    \midrule
    JanusVLN (7B) & $21.9$ & $22.67$ & $15.71$ & $10.83$ & $13.78$ & $0.72$ & $0.64$ \\
    Uni-NaVid (7B) & $12.6$ & -- & $12.60$ & $7.70$ & $9.43$ & $0.24$ & $0.23$ \\
    VG-AVS (7B) & $24.3$ & $5.31$ & $3.80$ & $2.31$ & $3.30$ & $9.61$ & $9.62$ \\
    VLMnav (4B) & $4.1$ & -- & $1.89$ & $2.97$ & $2.58$ & $3.80$ & $4.22$ \\
    \textbf{Ours (4B)} & $\mathbf{47.7}$ & $7.66$ & $3.59$ & $3.89$ & $5.08$ & $2.83$ & $2.79$ \\
    \bottomrule
  \end{tabular}
  }
  \caption{Multi-step inference efficiency and runtime on the primary benchmark trajectories. SR (All) is the overall mean SR from Table~\ref{tab:main_results}, reported in percent under the shared motion budget. Inference-step counts are averaged over successful trajectories up to the first successful frame for each intent family. Time/Inf. averages all recorded trajectory steps, while Succ. Time/Inf. averages successful trajectories up to the first successful frame.}
  \label{tab:inference_efficiency}
\end{table}

\subsection{Additional Intent-conditioned View Predictions}

\begin{figure*}[p]
  \centering
  \begin{subfigure}{\textwidth}
    \raggedright
    \includegraphics[width=0.235\textwidth]{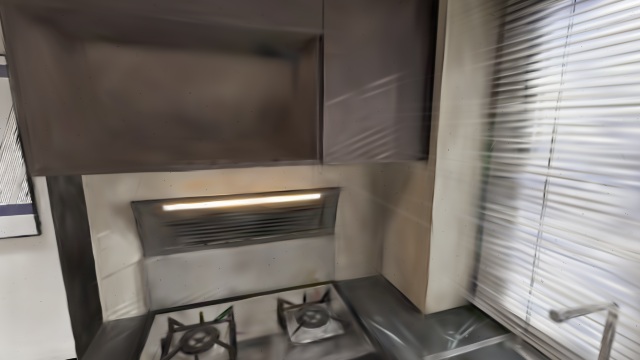}\hspace{0.35em}\includegraphics[width=0.235\textwidth]{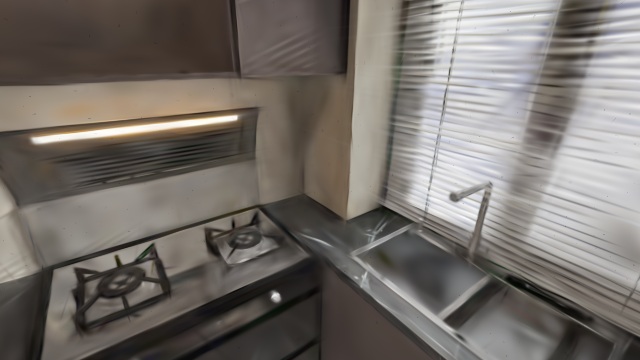}
    \caption{Look at the sink under the faucet.}
  \end{subfigure}
  \vspace{0.55em}

  \begin{subfigure}{\textwidth}
    \raggedright
    \includegraphics[width=0.235\textwidth]{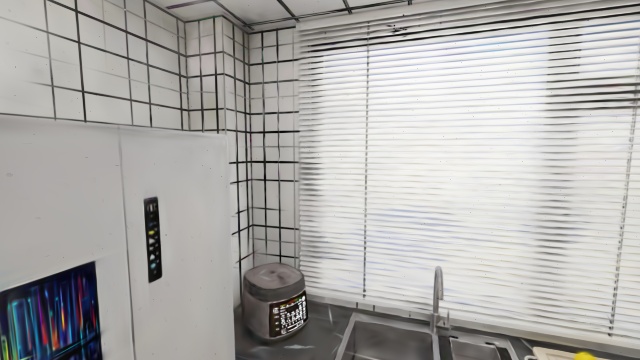}\hspace{0.35em}\includegraphics[width=0.235\textwidth]{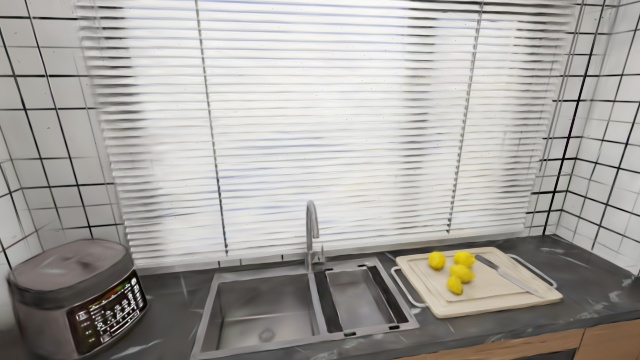}
    \caption{Look at the lemons on the chopping board.}
  \end{subfigure}
  \vspace{0.55em}

  \begin{subfigure}{\textwidth}
    \raggedright
    \includegraphics[width=0.235\textwidth]{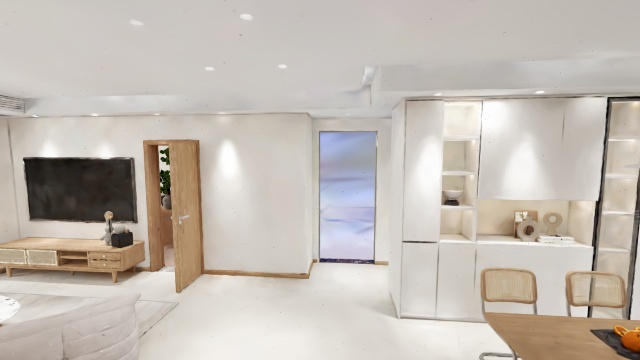}\hspace{0.35em}\includegraphics[width=0.235\textwidth]{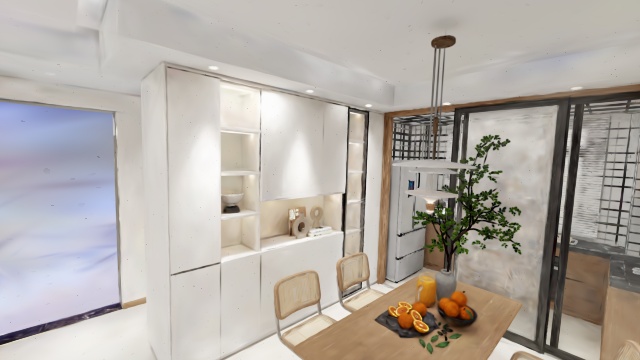}
    \caption{Find the orange juice on the dining table.}
  \end{subfigure}
  \vspace{0.55em}

  \begin{subfigure}{\textwidth}
    \raggedright
    \includegraphics[width=0.235\textwidth]{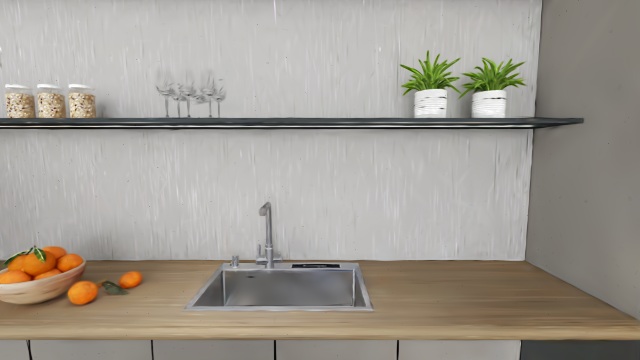}\hspace{0.35em}\includegraphics[width=0.235\textwidth]{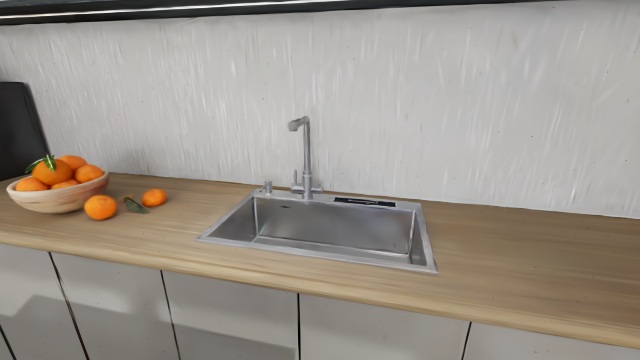}\hspace{0.35em}\includegraphics[width=0.235\textwidth]{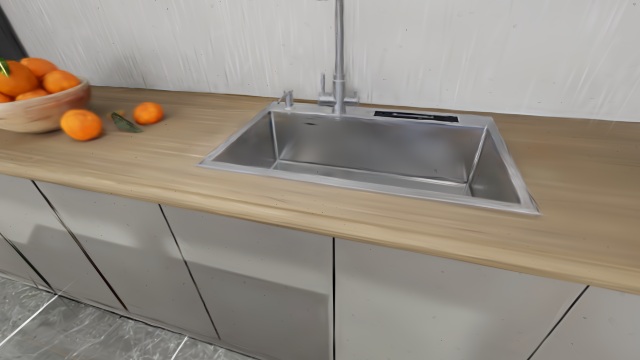}
    \caption{Check what is under the sink.}
  \end{subfigure}
  \vspace{0.55em}

  \begin{subfigure}{\textwidth}
    \raggedright
    \includegraphics[width=0.235\textwidth]{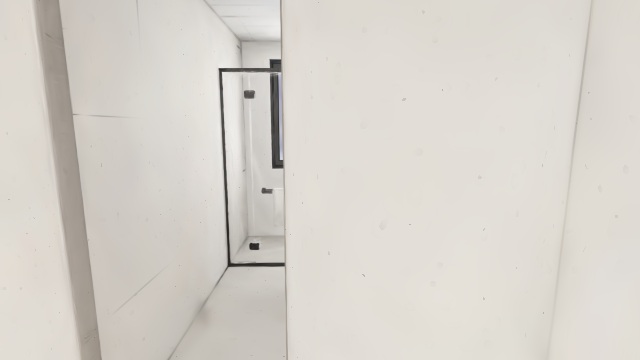}\hspace{0.35em}\includegraphics[width=0.235\textwidth]{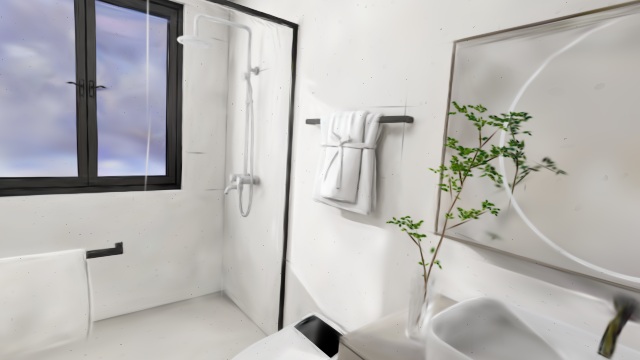}
    \caption{Check what is behind the wall.}
  \end{subfigure}
  \vspace{0.55em}

  \begin{subfigure}{\textwidth}
    \raggedright
    \includegraphics[width=0.235\textwidth]{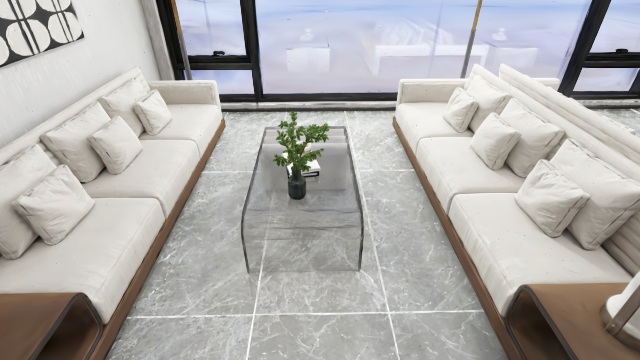}\hspace{0.35em}\includegraphics[width=0.235\textwidth]{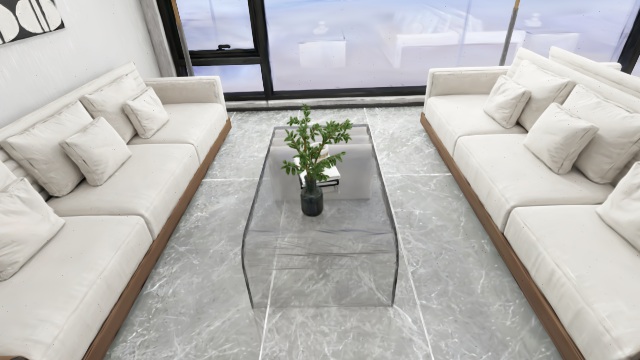}\hspace{0.35em}\includegraphics[width=0.235\textwidth]{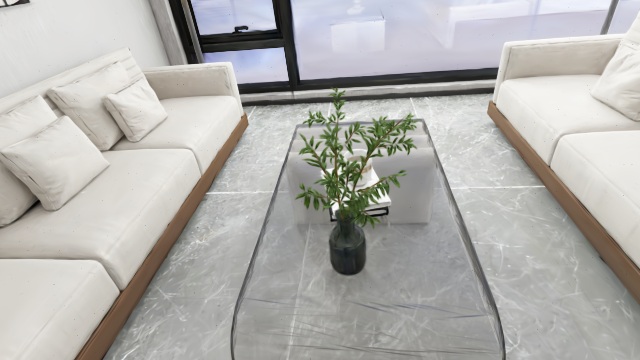}\hspace{0.35em}\includegraphics[width=0.235\textwidth]{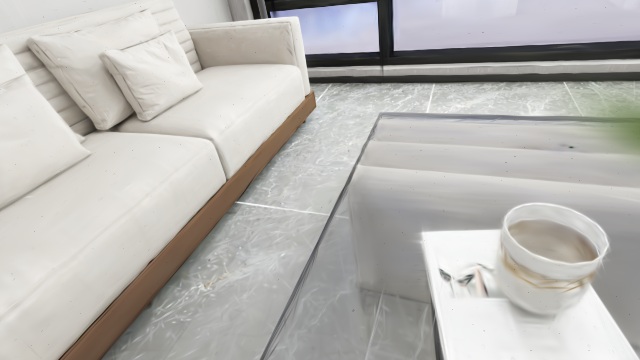}
    \caption{Get a clear view of the entire cup on the coffee table in front of you.}
  \end{subfigure}
  \vspace{0.55em}

  \begin{subfigure}{\textwidth}
    \raggedright
    \includegraphics[width=0.235\textwidth]{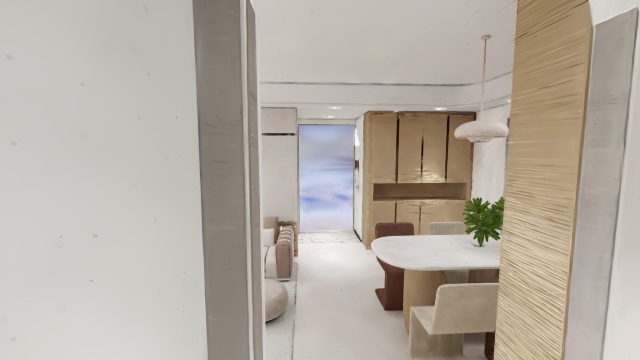}\hspace{0.35em}\includegraphics[width=0.235\textwidth]{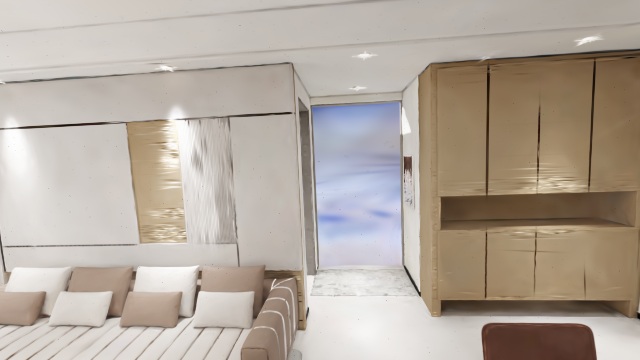}\hspace{0.35em}\includegraphics[width=0.235\textwidth]{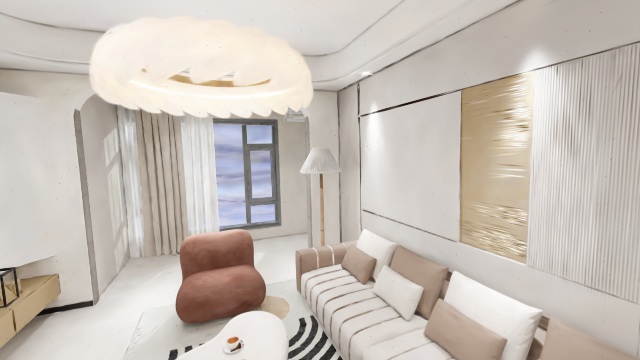}
    \caption{Go to the multi-seat sofa in the living room.}
  \end{subfigure}
  \vspace{0.55em}

  \caption{Additional qualitative LIME prediction examples. For each row, images are ordered from the start frame to the first successful frame.}
  \label{fig:qualitative_picked_examples}
\end{figure*}

Figure~\ref{fig:qualitative_picked_examples} shows additional LIME prediction examples in InteriorGS scenes. Each row starts from the initial observation and proceeds through LIME-predicted views up to the first successful view. These examples focus on local viewpoint changes that still require intent reasoning and $SE(3)$ pose prediction, with the goal reached in one or a few steps.

\subsection{LIME as Active Perception Module for Manipulation}
\label{sec:supp-libero}

Sec.~\ref{sec:exp_results} summarizes our LIBERO-Goal~\cite{liu2023liberobenchmarkingknowledgetransfer} evaluation; here we provide the full protocol, per-task results, and qualitative examples. LIBERO-Goal is a 10-task language-conditioned manipulation suite that varies task goals while keeping objects and layouts controlled.
For each task instruction, we compare two pipelines from the same initial robot state: directly running a VLA policy $\pi_{0.5}$~\cite{intelligence2025pi05visionlanguageactionmodelopenworld}, and first using LIME to move the wrist camera before running the same policy.
LIME receives a simple target-seeking instruction, ``Look for the \texttt{<object>},'' using the task-relevant object or fixture.
After LIME predicts a relative camera pose, the robot arm moves the wrist camera to the predicted view, and $\pi_{0.5}$ is then executed with the standard LIBERO-Goal manipulation budget. Instead of using the default LIBERO initial wrist-camera views, we sample constrained reachable starts where the task-relevant object or fixture is only partially visible. This initial observation is generally more challenging for a VLA policy to finish the tasks.
Both pipelines are evaluated from the same sampled starts, yielding paired direct-vs-LIME rollouts that isolate whether one active camera-motion step improves the visual precondition for manipulation.

\raggedbottom

\begin{table}[H]
  \centering
  \caption{LIBERO-Goal manipulation results from sampled initial states. Each task uses five sampled starts, with one rollout per start. Direct SR runs $\pi_{0.5}$ from the sampled start pose, while LIME+$\pi_{0.5}$ SR first applies one LIME-predicted wrist-camera motion from the same state. SR is reported as a percentage.}
  \label{tab:libero_goal_manipulation}
  \resizebox{\linewidth}{!}{
  \begin{tabular}{clcc}
    \toprule
    \textbf{Task ID} & \textbf{Task instruction} & \textbf{Direct $\pi_{0.5}$ SR} & \textbf{LIME+$\pi_{0.5}$ SR} \\
    \midrule
    0 & Open the middle drawer of the cabinet & $60\%$ & $100\%$ \\
    1 & Put the bowl on the stove & $60\%$ & $100\%$ \\
    2 & Put the wine bottle on top of the cabinet & $0\%$ & $60\%$ \\
    3 & Open the top drawer and put the bowl inside & $0\%$ & $20\%$ \\
    4 & Put the bowl on top of the cabinet & $60\%$ & $100\%$ \\
    5 & Push the plate to the front of the stove & $20\%$ & $80\%$ \\
    6 & Put the cream cheese in the bowl & $0\%$ & $60\%$ \\
    7 & Turn on the stove & $0\%$ & $60\%$ \\
    8 & Put the bowl on the plate & $60\%$ & $100\%$ \\
    9 & Put the wine bottle on the rack & $0\%$ & $60\%$ \\
    \midrule
    \textbf{Overall} & \textbf{All sampled rollouts} & $\mathbf{26\%}$ & $\mathbf{74\%}$ \\
    \bottomrule
  \end{tabular}
  }
\end{table}

\begin{figure}[p]
  \centering
  \begin{tabular}{@{}c@{\hspace{0.025\linewidth}}c@{\hspace{0.08\linewidth}}c@{\hspace{0.025\linewidth}}c@{}}
    \multicolumn{2}{@{}c@{}}{\footnotesize\textbf{Initial state}} &
    \multicolumn{2}{@{}c@{}}{\footnotesize\textbf{After LIME}} \\
    \scriptsize External & \scriptsize Wrist &
    \scriptsize External & \scriptsize Wrist \\[0.25em]

    \includegraphics[width=0.205\linewidth]{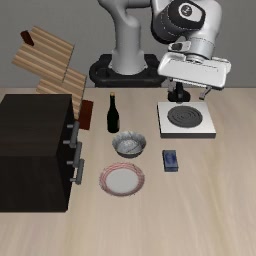} &
    \includegraphics[width=0.205\linewidth]{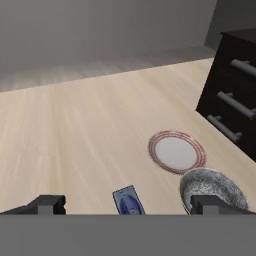} &
    \includegraphics[width=0.205\linewidth]{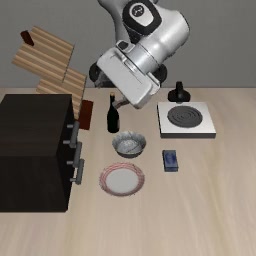} &
    \includegraphics[width=0.205\linewidth]{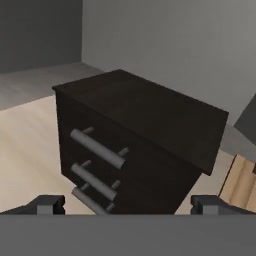} \\
    \multicolumn{4}{c}{\footnotesize \textbf{Task 0:} Open the middle drawer of the cabinet} \\[0.35em]

    \includegraphics[width=0.205\linewidth]{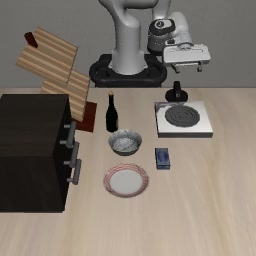} &
    \includegraphics[width=0.205\linewidth]{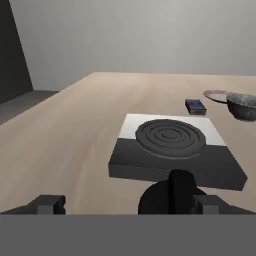} &
    \includegraphics[width=0.205\linewidth]{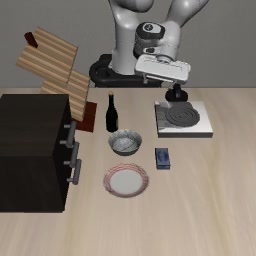} &
    \includegraphics[width=0.205\linewidth]{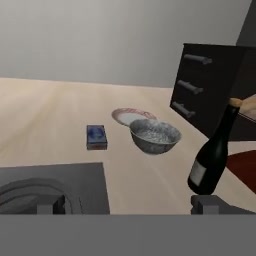} \\
    \multicolumn{4}{c}{\footnotesize \textbf{Task 1:} Put the bowl on the stove} \\[0.35em]

    \includegraphics[width=0.205\linewidth]{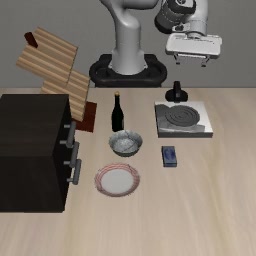} &
    \includegraphics[width=0.205\linewidth]{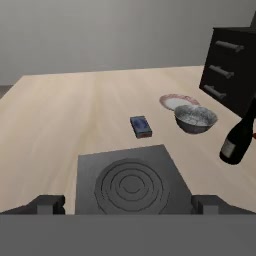} &
    \includegraphics[width=0.205\linewidth]{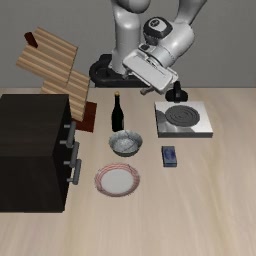} &
    \includegraphics[width=0.205\linewidth]{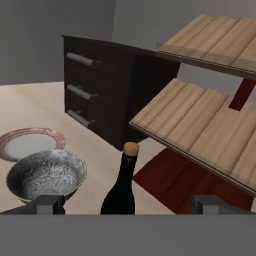} \\
    \multicolumn{4}{c}{\footnotesize \textbf{Task 2:} Put the wine bottle on top of the cabinet} \\[0.35em]

    \includegraphics[width=0.205\linewidth]{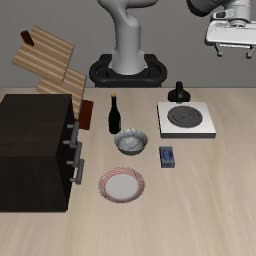} &
    \includegraphics[width=0.205\linewidth]{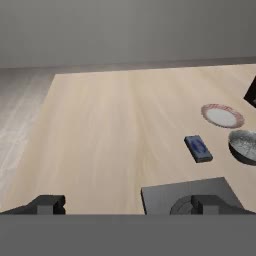} &
    \includegraphics[width=0.205\linewidth]{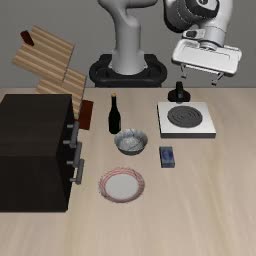} &
    \includegraphics[width=0.205\linewidth]{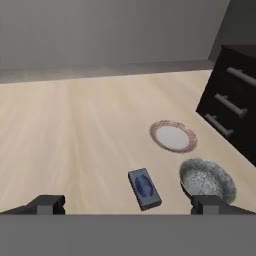} \\
    \multicolumn{4}{c}{\footnotesize \textbf{Task 4:} Put the bowl on top of the cabinet} \\[0.35em]

    \includegraphics[width=0.205\linewidth]{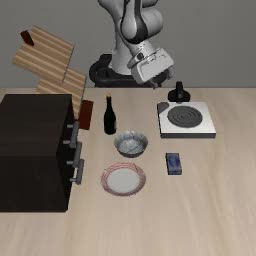} &
    \includegraphics[width=0.205\linewidth]{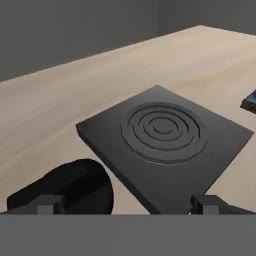} &
    \includegraphics[width=0.205\linewidth]{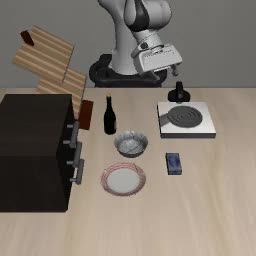} &
    \includegraphics[width=0.205\linewidth]{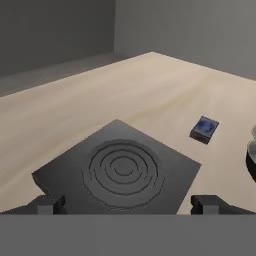} \\
    \multicolumn{4}{c}{\footnotesize \textbf{Task 6:} Put the cream cheese in the bowl}
  \end{tabular}
  \caption{Qualitative LIBERO-Goal examples from sampled initial states. Columns are grouped by the sampled start view and the post-LIME view; each group shows the external and wrist cameras. The post-LIME views are captured before executing the downstream $\pi_{0.5}$ manipulation policy.}
  \label{fig:libero_goal_active_perception}
\end{figure}

Table~\ref{tab:libero_goal_manipulation} shows that the LIME-assisted pipeline substantially improves the downstream manipulation success rate in this targeted setting, \textbf{increasing overall success from $\mathbf{26\%}$ to $\mathbf{74\%}$}.
The result supports the intended use of LIME as an active visual preconditioning module: it acquires a more informative wrist-camera view before the same $\pi_{0.5}$ policy performs the manipulation.
The per-task results also show that the benefit is not uniform across manipulation tasks; for example, the two-stage drawer-and-placement task remains difficult even with the additional LIME viewpoint.
Figure~\ref{fig:libero_goal_active_perception} shows representative sampled starts and the corresponding post-LIME wrist-camera views, illustrating how the active-perception step changes the visual evidence available before executing the manipulation policy.

\subsection{LIME for Embodied Question Answering (EQA) Task}
\label{sec:supp-eqa}

LIME's intent-aware camera-motion capability can also be used for embodied question answering (EQA). We evaluate LIME on AVS-ProcTHOR \cite{koo2025toward}, an active-view-selection benchmark for visual question answering. Instead of judging whether a predicted camera trajectory satisfies an explicit intent, AVS-ProcTHOR measures whether an agent can select a view that improves question answering. This setting tests whether the camera-motion behavior learned by LIME transfers to a different simulator, action interface, and downstream objective.

We apply three adaptations to deploy LIME on AVS-ProcTHOR. First, following the baselines, we use Gemini-2.5-Flash~\cite{comanici2025gemini} as the verifier model that answers the question from the rendered view at the predicted pose. Second, because LIME takes an intent-style instruction rather than a question-answer input, we convert each AVS-ProcTHOR question into a view-selection instruction using its parsed target and supporting object. The conversion uses simple ``look at'' templates: Existence questions are converted to support-focused instructions such as ``Look at the sink,'' Counting questions to plural coverage instructions such as ``Look at all the apples on the side table,'' and State questions to target-and-support instructions such as ``Look at the book on the dining table.'' This preserves the visual evidence needed by the original VQA question while matching LIME's intent-conditioned camera-motion interface. Third, AVS-ProcTHOR expects a planar active-view action rather than a relative pose in 3D. We convert LIME's predicted relative pose into the VG-AVS action format consisting of a heading rotation, forward displacement, and final view rotation. The resulting action is then executed through the original VG-AVS ProcTHOR rendering path using the AI2-THOR agent camera, matching the official AVS-ProcTHOR embodied-agent protocol.

Table~\ref{tab:avs_procthor_results} reports the evaluation results. With LIME as the action model, the EQA agent outperforms the listed backbone, spatial-VLM, and EQA-framework baselines, while the proprietary models and the specialized VG-AVS framework remain stronger. This comparison should be read in context: LIME is trained from real-world egocentric video supervision and receives no training or fine-tuning on the ProcTHOR scene distribution, the AVS-ProcTHOR question distribution, or the VG-AVS planar action format. In contrast, the VG-AVS SFT/RL models are trained directly on ProcTHOR active-VQA data with the same planar action interface used at evaluation time, so remaining below these specialized variants is expected.

\begin{table}[H]
\centering
\caption{Results on the AVS-ProcTHOR benchmark. Accuracy is reported in percent using Gemini-2.5-Flash as the answer verifier, following the AVS-ProcTHOR evaluation setup. LIME predicts a relative $SE(3)$ camera motion, which we convert to the VG-AVS planar action format before rendering through the official ProcTHOR agent-camera path. All results except LIME are reported from VG-AVS~\cite{koo2025toward}. Average is the unweighted mean of the three question types, following VG-AVS.}
\label{tab:avs_procthor_results}
\resizebox{\linewidth}{!}{
\begin{tabular}{llcccc}
\toprule
\multicolumn{2}{c}{\textbf{Action Model}} &
\multicolumn{4}{c}{\textbf{AVS-ProcTHOR}} \\
\cmidrule(lr){3-6}
 & & \textbf{Existence} & \textbf{Counting} & \textbf{State} & \textbf{Average} \\
\midrule

\multirow{2}{*}{\textbf{No Action}}
& Query view  & 49.22 & 16.36 & 61.57 & 42.38 \\
& Target view & 93.02 & 69.14 & 92.58 & 84.91 \\
\midrule

\multirow{1}{*}{\textbf{Backbone Model}}
& Qwen2.5-VL-7B~\cite{bai2025qwen25vl} & 64.34 & 29.74 & 56.55 & 50.21 \\
% & Qwen3-VL-4B        & 65.31 & 25.28 & 66.38 & 51.39 \\
\midrule

\multirow{2}{*}{\textbf{Spatial VLMs}}
& ViLaSR~\cite{wu2025vilasr} & 57.95 & 25.46 & 52.84 & 45.42 \\
& SpatialReasoner~\cite{ma2025spatialreasoner} & 54.65 & 22.68 & 52.62 & 43.32 \\
\midrule

\textbf{EQA Framework}
& Fine-EQA~\cite{EXPRESSBench} & 63.57 & 31.97 & 64.41 & 53.32 \\
\midrule

\multirow{2}{*}{\textbf{Proprietary Models}}
& GPT-5~\cite{openai2025gpt5} & 81.01 & 55.58 & 79.69 & 72.09 \\
& Gemini-2.5-Pro~\cite{comanici2025gemini} & 82.95 & 52.79 & 81.00 & 72.25 \\
\midrule

\multirow{3}{*}{\makecell[l]{\textbf{VG-AVS Framework}}}
& SFT                & 91.28 & 57.06 & 83.84 & 77.39 \\
& RL                 & 86.82 & 65.24 & 83.41 & 78.49 \\
& SFT+RL             & 91.47 & 69.52 & 90.17 & 83.72 \\
\midrule

\makecell[l]{\textbf{Ours}}
& LIME               & 65.50 & 36.62 & 72.71 & 58.28 \\

\bottomrule
\end{tabular}
}
\end{table}

\begin{figure}[p]
    \centering
    \setlength{\abovecaptionskip}{0.3em}
    \setlength{\belowcaptionskip}{0pt}

    \begin{tabular}{@{}c@{}}
    \includegraphics[width=0.545\linewidth]{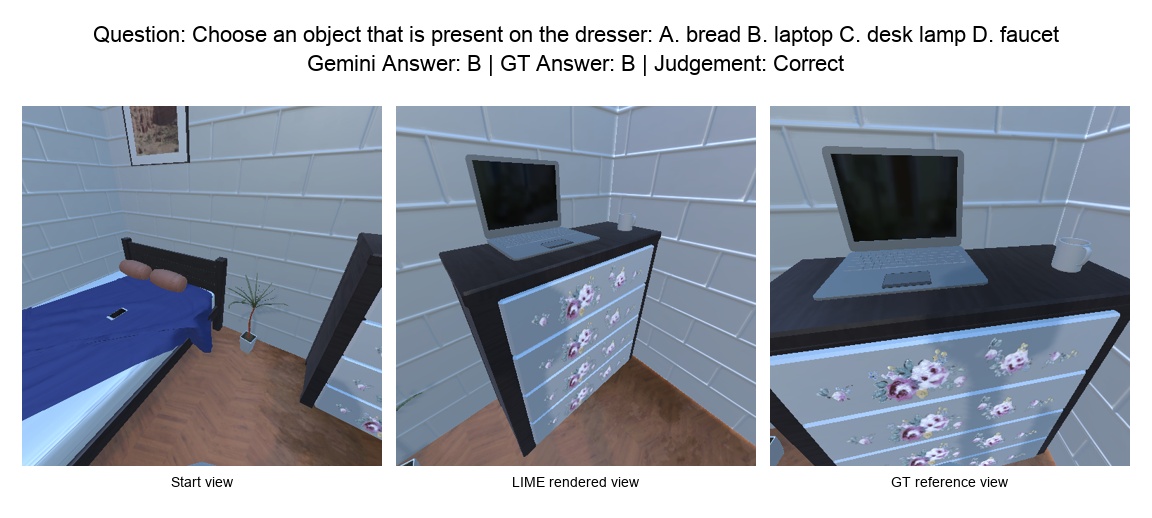} \\[0.1em]
    \includegraphics[width=0.545\linewidth]{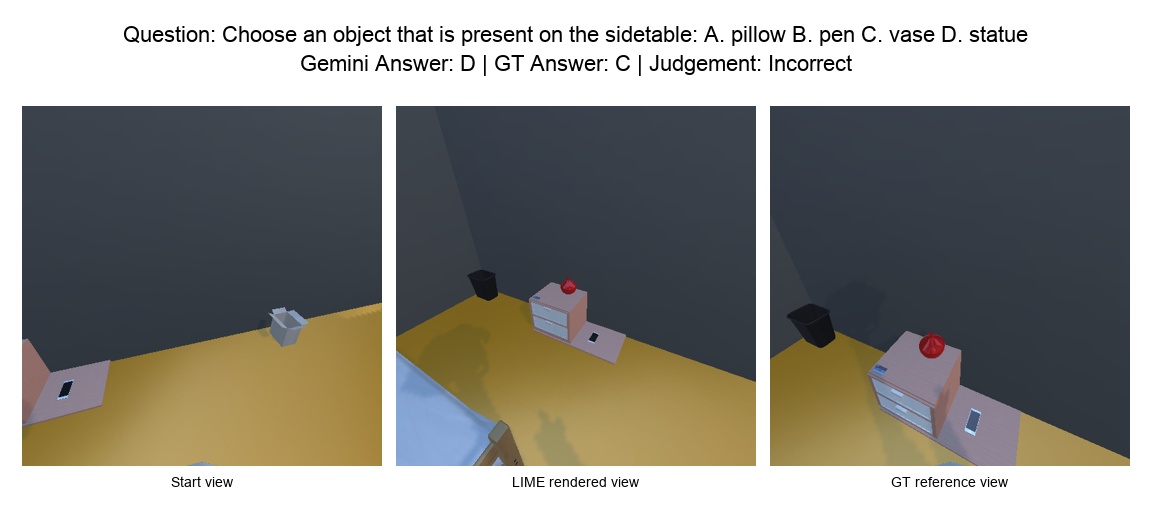} \\[0.1em]
    \includegraphics[width=0.545\linewidth]{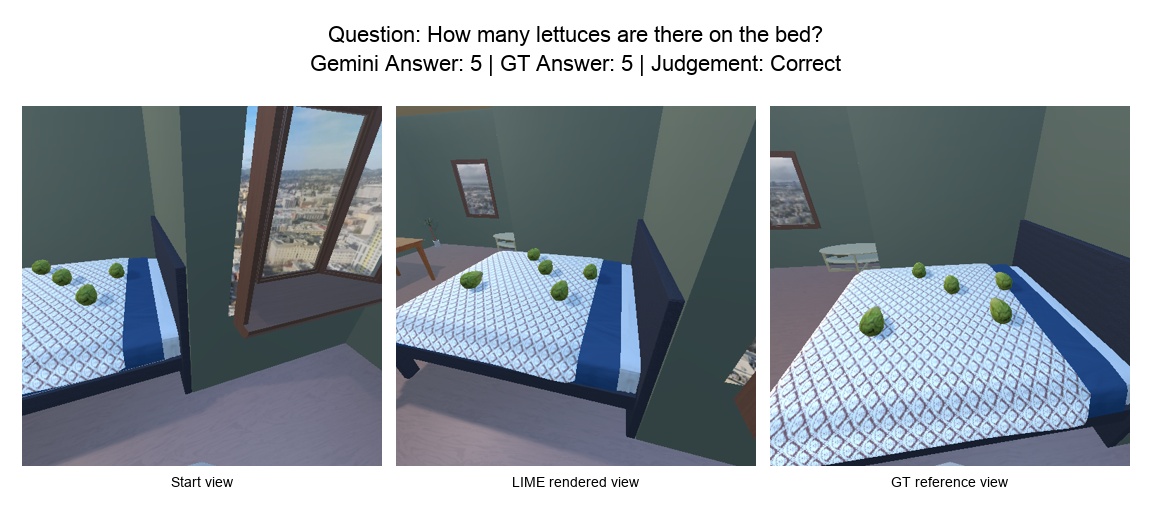} \\[0.1em]
    \includegraphics[width=0.545\linewidth]{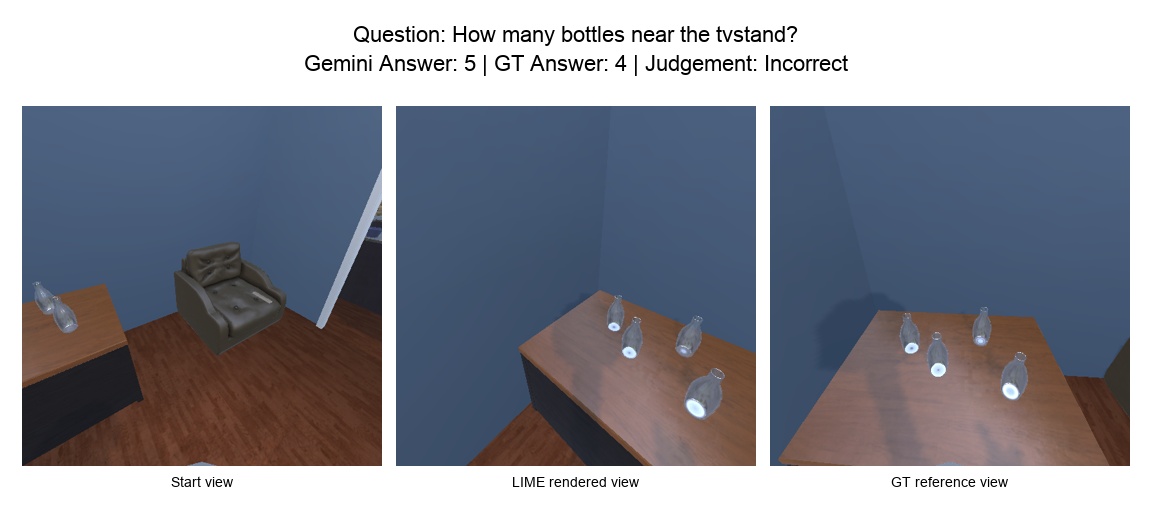} \\[0.1em]
    \includegraphics[width=0.545\linewidth]{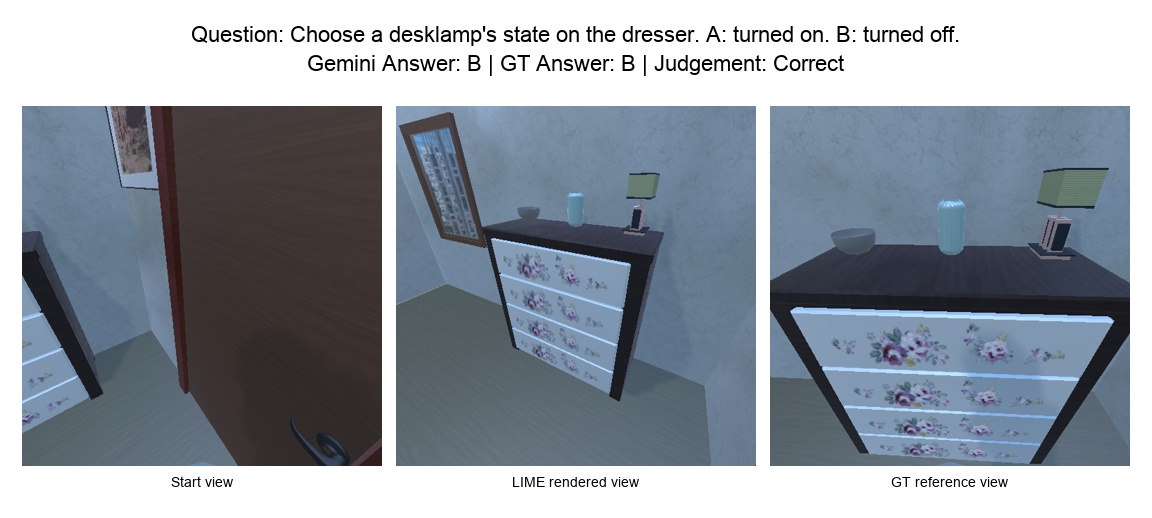} \\[0.1em]
    \includegraphics[width=0.545\linewidth]{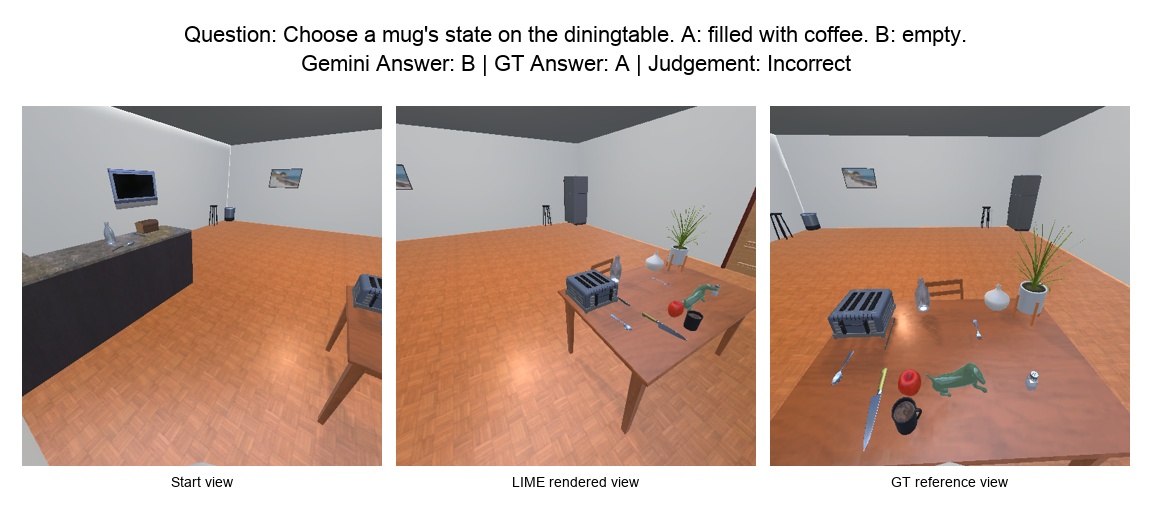}
    \end{tabular}

    \caption{Qualitative AVS-ProcTHOR examples showing the start view, LIME rendered view, and held-out reference view. Each example occupies one row; rows are ordered as two Existence examples, two Counting examples, and two State examples.}
    \label{fig:supp_lime_qualitative_examples}
\end{figure}

Figure~\ref{fig:supp_lime_qualitative_examples} shows qualitative examples from the three AVS-ProcTHOR question types. Each example is shown on a separate row, with two rows each for Existence, Counting, and State. LIME's predicted view is qualitatively meaningful and exposes the relevant object or region. Within each question type, the two examples include one case where Gemini-2.5-Flash answers correctly from the rendered view and one case where it does not. This illustrates that failures can arise not only from an unhelpful predicted pose, but also from the downstream answer verifier failing to extract the correct answer from an otherwise informative view.

\clearpage

\clearpage

\subsection{LIME for Multi-step Robot Tasks}
\label{sec:supp-multistep}

\begin{figure}[H]
    \centering
    \includegraphics[width=\linewidth]{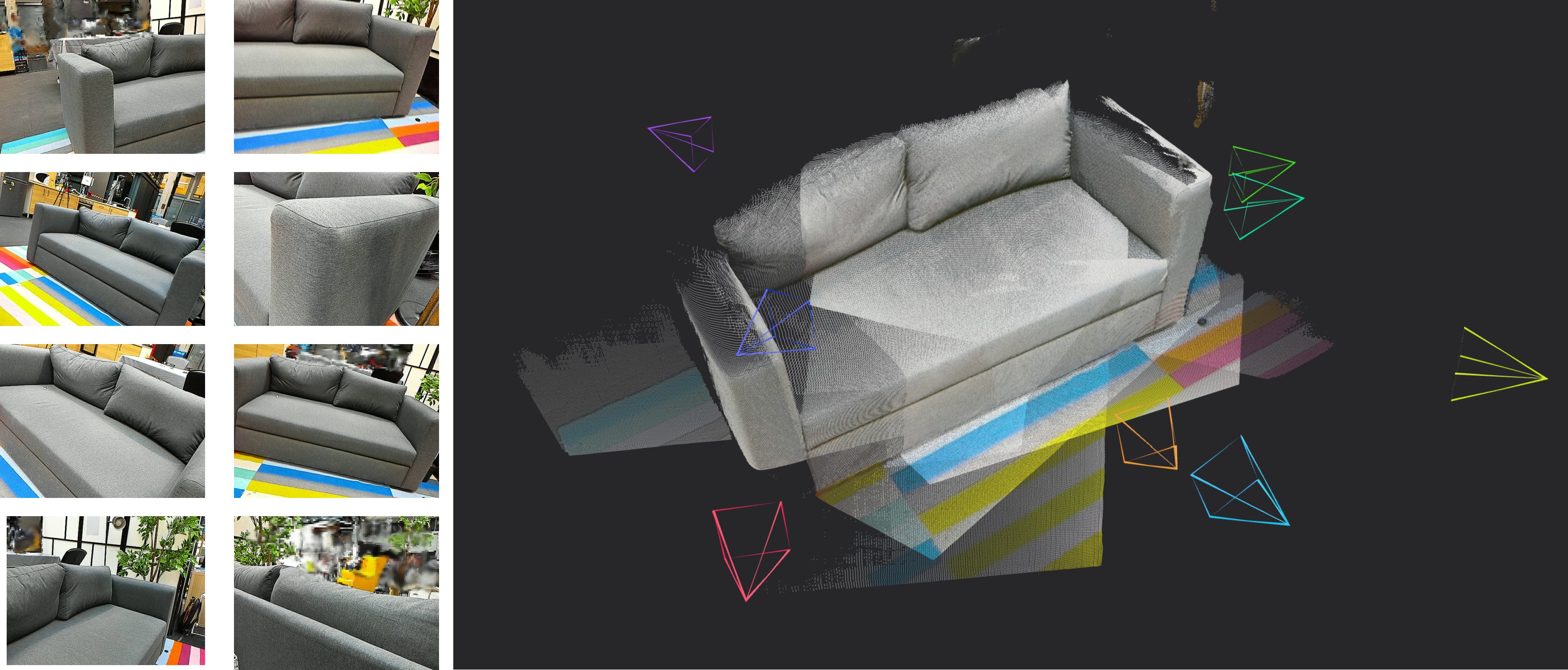}

    \caption{Language-instructed object scanning example. Repeated prompts and novelty-biased sampling produce a multi-view trajectory around the target. The reconstruction result on the right is generated with VGGT-Omega \cite{wang2026vggtomega}. }
    \label{fig:supp_robot_scan}
\end{figure}

We further test LIME in two preliminary robot settings that reuse the same sequential camera-motion interface.
First, LIME can be used for language-instructed object scanning by repeatedly prompting the robot with a fixed intent such as ``look at the object from a different angle.''
At each step, the robot observes the current wrist-camera image, samples multiple candidate camera motions from the flow-matching head, and executes the sample with the largest deviation from previously visited views to encourage novel coverage.
This produces a multi-view scanning trajectory around the target object without explicitly training a separate scanning policy.
Figure~\ref{fig:supp_robot_scan} shows an example scanning evaluation trajectory.

\begin{figure}[H]
    \centering
    \includegraphics[width=\linewidth]{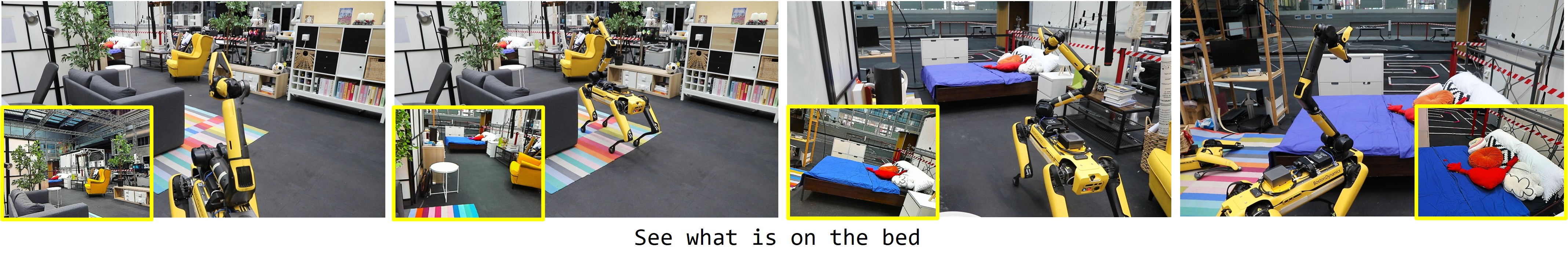}

    \caption{Mid-distance navigation example. Repeated target-directed prompting drives progress toward the goal over three local camera-motion iterations.}
    \label{fig:supp_robot_navigation}
\end{figure}

Second, we evaluate mid-distance navigation by repeatedly prompting LIME with the same target-directed command over multiple sequential camera-motion steps.
Although LIME predicts only local relative camera motions, the repeated execution shows that the robot can continue making progress toward a target across several viewpoints.
Figure~\ref{fig:supp_robot_navigation} shows an example mid-distance navigation evaluation trajectory.

Videos of the full processes for these two tasks, together with additional real-world results, are available on the project page.

\end{document}